\documentclass[journal]{IEEEtran}
\usepackage{wrapfig}

\usepackage{amsmath,amsfonts,bm}

\def\eqref#1{equation~\ref{#1}}

\def\1{\bm{1}}

\DeclareMathAlphabet{\mathsfit}{\encodingdefault}{\sfdefault}{m}{sl}
\SetMathAlphabet{\mathsfit}{bold}{\encodingdefault}{\sfdefault}{bx}{n}

\newcommand{\R}{\mathbb{R}}

\newcommand{\CB}{\mathrm{CodeBLEU}}
\newcommand{\cl}{\operatorname{clip}}
\newcommand{\cnt}{\operatorname{count}}
\usepackage{graphicx}
\usepackage{hyperref}
\usepackage{url}
\usepackage{booktabs}
\usepackage[ruled,vlined,linesnumbered]{algorithm2e}
\usepackage[dvipsnames]{xcolor}
\usepackage{listings}
\usepackage[inline]{enumitem}
\usepackage{amssymb}
\usepackage{amsmath,dsfont}
\usepackage[caption=false,font=footnotesize]{subfig}
\usepackage{float}
\usepackage{natbib}
\usepackage{cleveref}

\usepackage{framed}
\usepackage[most]{tcolorbox}
\usepackage{multirow}

\usepackage{marvosym}

\tcbset {
  enhanced jigsaw,
  base/.style={
    arc=0mm, 
    bottomtitle=0.5mm,
    boxrule=0mm,
    colbacktitle=black!10!white, 
    coltitle=black, 
    fonttitle=\bfseries, 
    left=2.5mm,
    leftrule=1mm,
    right=3.5mm,
    title={#1},
    toptitle=0.75mm, 
  }
}

\newtcolorbox{promptbox}[1]{%
  colback=blue!3!white,
  colframe=blue!50!black,
  boxrule=0.6pt,
  arc=2mm,
  fonttitle=\bfseries,
  title={#1},
  left=1.2ex, right=1.2ex, top=1ex, bottom=1ex
}

\newtcolorbox{promptbox*}[1]{
  width=\textwidth,
  colback=blue!3!white,
  colframe=blue!50!black,
  boxrule=0.6pt,
  arc=2mm,
  fonttitle=\bfseries,
  title={#1},
  left=1.2ex, right=1.2ex, top=1ex, bottom=1ex
}

\newtcolorbox{subsectionbox}[1]{%
  colback=#1!10,
  colframe=#1!50!black,
  boxrule=0.4pt,
  arc=1mm,
  left=0.8ex, right=0.8ex, top=0.8ex, bottom=0.8ex
}

\newtcolorbox{subsectionbox*}[1]{
  width=\textwidth,
  colback=#1!10,
  colframe=#1!50!black,
  boxrule=0.4pt,
  arc=1mm,
  left=0.8ex, right=0.8ex, top=0.8ex, bottom=0.8ex
}

\definecolor{mybeige}{RGB}{210,190,180}
\definecolor{mylightgray}{HTML}{E5E5E5}
\definecolor{titlegray}{gray}{0.3}
\newtcolorbox{taskbox}[1]{%
  colback=blue!3!white,
  colframe=mybeige,
  boxrule=0.6pt,
  arc=2mm,
  fonttitle=\bfseries,
  title={#1},
  coltitle=titlegray,
  left=1.2ex, right=1.2ex, top=1ex, bottom=1ex
}

\definecolor{keywordcolor}{rgb}{0.0,0.0,0.6}
\definecolor{commentcolor}{rgb}{0.0,0.5,0.0}
\definecolor{stringcolor}{rgb}{0.58,0,0.82}

\lstdefinestyle{pythonstyle}{%
    language=Python,
    basicstyle=\ttfamily\scriptsize,
    keywordstyle=\color{keywordcolor}\bfseries,
    commentstyle=\color{commentcolor}\itshape,
    stringstyle=\color{stringcolor},
    showstringspaces=false,
    frame=single,
    numbers=left,
    numberstyle=\tiny\color{gray},
    stepnumber=1,
    tabsize=4,
}

\begin{document}

\title{From Heuristic Selection to Automated Algorithm Design:\\ LLMs Benefit from Strong Priors}

\author{
\IEEEauthorblockN{Qi Huang, Furong Ye\textsuperscript{(\Letter)}, Ananta Shahane, Thomas B\"ack, Niki van Stein}\\
\IEEEauthorblockA{
Leiden Institute of Advanced Computer Science (LIACS)\\
Leiden University\\
2333 CC Leiden, The Netherlands\\
\{q.huang, f.ye, a.a.shahane, t.h.w.baeck, n.van.stein\}@liacs.leidenuniv.nl\\
Corresponding Author (\Letter): \texttt{f.ye@liacs.leidenuniv.nl} (Furong Ye)
}
}

\maketitle

\begin{abstract}
Large Language Models (LLMs) have already been widely adopted for automated algorithm design, demonstrating strong abilities in generating and evolving algorithms across various fields.
Existing work has largely focused on examining their effectiveness in solving specific problems, with search strategies primarily guided by adaptive prompt designs.
\textcolor{black}{In this paper, through investigating the token-wise attribution of the prompts to LLM-generated algorithmic code, we show that the embedded code examples and their associated strategy instructions have the strongest influence on the generated algorithms.}
\textcolor{black}{Building upon this insight, we propose leveraging existing benchmark algorithms as strong priors to guide LLM-driven optimization. We evaluate this approach with three LLMs on two black-box optimization benchmarks: the pseudo-Boolean optimization suite (pbo) and the black-box optimization suite (bbob). Despite relying on a simple (1+1) elitist search, it obtains the best average rank in five of six benchmark--LLM settings against six state-of-the-art LLM-driven methods, including more sophisticated search strategies.}
Our findings highlight the value of integrating benchmarking studies to enhance both efficiency and robustness of the LLM-driven black-box optimization methods.
\end{abstract}


\section{Introduction}
\textcolor{black}{Decades of development in the field of evolutionary computation have resulted in a vast amount of heuristic optimizers, which have been claimed to be effective. In contrast to researchers who confidently root for their own solvers, real-world practitioners, when handling an optimization problem, often encounter the challenges of \begin{enumerate*}
\item how to choose the best optimizer, and \item how to construct a dedicated optimizer \end{enumerate*}. Recent dedicated research has been proposed to automate the two tasks.} 
\textcolor{black}{
More specifically, on one hand, automated algorithm selection (AAS) addresses the first challenge of choosing, for each individual problem instance, an optimizer that is expected to perform best, based on prior performance data and measurable instance characteristics such as exploratory landscape analysis (ELA) features~\citep{kerschke2019automated}. On the other hand, automated algorithm design (AAD) aims to construct or configure metaheuristic algorithms automatically by exploring a space of algorithmic components and parameterizations, guided by performance feedback on representative problem instances~\citep{stutzle2018automated}.~\citet{schede2022survey} described classical AAD as focusing on algorithm configuration, relying on tuning parameter values and selecting operators. They further argue that AAS can be considered as a special case of instance-specific AAD where the decision variable is the categorical choice of heuristics.} 

Recently, the emergence of Large Language Models (LLMs) has elevated AAD to a new level, enabling the direct selection and evolution of algorithms by optimizing code.
After the pioneering work of FunSearch~\citep{romera2024mathematical} demonstrating the potential of LLMs by solving the cap set~\citep{grochow2019new} and bin packing~\citep{coffman1984approximation} problems, LLM-driven optimization methods have been applied to solve problems across specific domains and to evolve existing algorithmic frameworks. They have achieved significant success in diverse applications such as scheduling and routing, satisfiability, multi-objective optimization, black-box optimization, etc.\textcolor{black}{~\citep{liuevolution,sun2025automatically,yao2025multi,huang2025autonomous,van2024llamea}}, as well as in advancing modularized algorithmic frameworks including Bayesian Optimization and Local Search for Pseudo Boolean Optimization~\citep{liularge,llameahpo,li2025autopbo,li2025llameabolargelanguagemodel}. Moreover, the LLM-driven AAD methods evolved from relying on massive scale sampling of LLMs (on the order of $10^6$) to employing evolutionary approaches that require only hundreds of samples. 

Search strategies are recognized as an essential component of LLM-driven optimization methods~\citep{zhang2024understanding}, which are embedded with variation operators, such as generating diverse algorithms, that are accomplished through specific prompt design. However, these prompts have been largely designed based on intuition in existing work, with the assumption that LLMs would respond reliably to linguistic instructions. This fact raises concerns about the underlying behavior of LLMs and hinders the development of more efficient, robust, and reliable LLM-driven approaches. To address this, we apply AttnLRP~\citep{achtibat2024attnlrp}, an attention-aware feature attribution method, for the first time to investigate the token-wise contribution of prompts in code generation and LLM-driven optimization studies, aiming to uncover the mechanisms driving LLMs' behavior and to design more effective and robust LLM-driven optimization approaches.

Our study focuses on black-box optimization (BBO, defined in Appendix~\ref{appendix:bench}), which does not provide an internal structure of the objective function. In practice, the algorithms can access only the problem metadata (e.g., variable domain, dimensionality) and the fitness values of explored solutions during the optimization process. 
However, existing LLM-driven approaches~\citep{van2024llamea,liuevolution} commonly embed prior knowledge, such as problem names, into their prompts. While this is a reasonable choice to effectively guide LLMs toward producing useful outcomes, it remains crucial to carefully consider practical constraints when developing tools for BBO.
Meanwhile, traditional AAD for BBO relies on algorithm configuration and algorithm selection methods, which are based on self-guided search or feature-based learning techniques to construct competitive algorithms within predefined frameworks for specific tasks~\citep{schede2022survey,kerschke2019automated}. Extensive benchmark studies have complemented these efforts, offering valuable guidelines for building and evaluating these techniques~\citep{bartz2020benchmarking,bennet2021nevergrad}. \textcolor{black}{In contrast to domains focusing on practical problems such as Satisfiability (SAT) and travelling salesman problem, where commonly accepted algorithms and benchmark rankings are available, selecting an appropriate algorithm for a given BBO remains challenging. Different BBO algorithms often obtain fundamentally different algorithmic structures, making it impractical to apply LLMs to optimize particular algorithmic modules as in prior work~\citep{liuevolution,sun2025automatically}. Instead, we leverage a set of benchmark algorithms to guide LLMs toward generating improved algorithms.} 

In this paper, we study LLM-driven optimization approaches by strictly adhering to the \emph{black-box} settings:
\textcolor{black}{the prompt never reveals the type and the identity of the benchmark suite or function.}
Furthermore, inspired by our investigation on token-wise contributions of prompts, we demonstrate that leveraging prior benchmark algorithms can effectively guide LLMs towards \textcolor{black}{stronger} and more robust performance. With extensive experiments on the pseudo-boolean optimization (pbo) suite~\citep{doerr2020benchmarking} and the continuous black-box optimization (bbob) suite~\citep{hansen2021coco}, we demonstrate the advantages of integrating established benchmark practice with LLM-driven approaches. Our findings highlight that this integration will benefit not only BBO but also the broader field of LLM-driven optimization.

Overall, this work contributes to:
\begin{itemize}
    \item \textcolor{black}{A systematic analysis of the token-wise contribution of prompt design in the LLM-driven optimization frameworks, demonstrating that the embedded example code and its associated strategy instruction have the strongest influence on the algorithmic code produced by LLMs.}
    \item An explicit demonstration that the behavior of LLMs can be effectively guided through providing specific example code, restricting the algorithmic search regions of LLMs.
    \item \textcolor{black}{A competitive benchmark-guided approach that leverages existing benchmark algorithms as strong priors to empower LLM-driven optimization methods on two well-established BBO benchmarks, \textit{pbo} and \textit{bbob}. Its performance against more complex search strategies shows that providing strong prior code to the LLM can matter more than an elaborate search strategy, offering new insights into designing efficient and robust LLM-driven optimization methodologies for future work.}
\end{itemize}

\section{Related Work}

\begin{figure}
    \centering
    \includegraphics[width=0.98\linewidth]{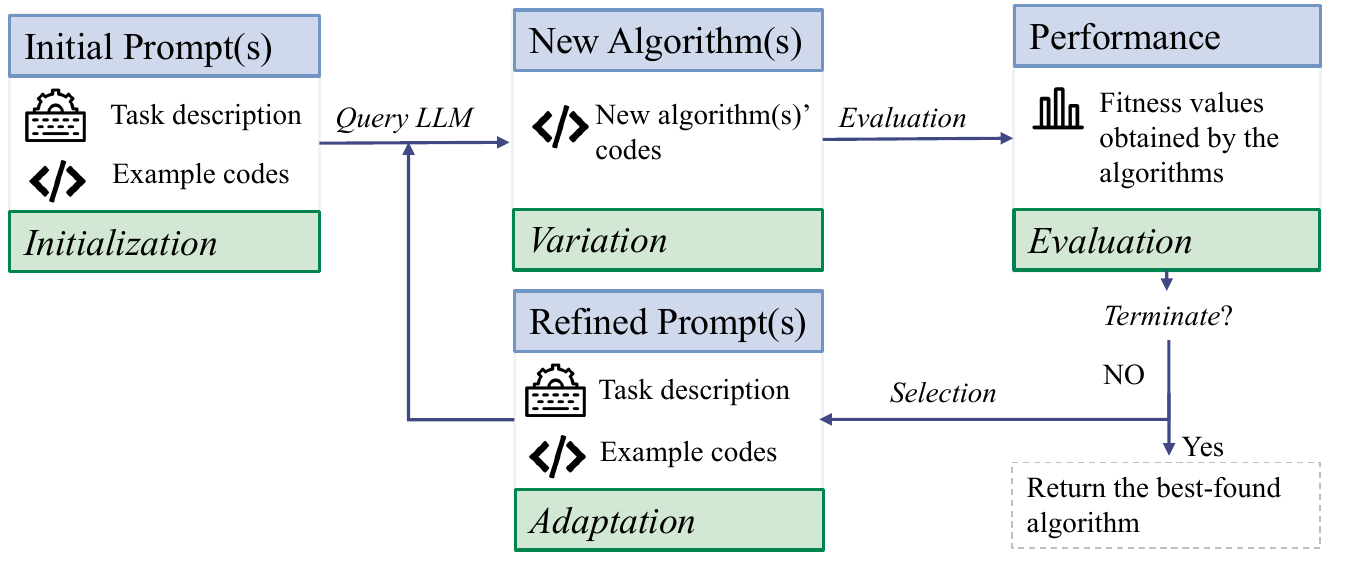}
    \caption{The workflow of LLM-driven optimization approaches}
    \label{fig:workflow}
\end{figure}

\subsection{LLM-driven Optimization}
\textcolor{black}{The idea of using LLMs to automatically evolve algorithms was first introduced by Algorithm Evolution using Large Language Model (AEL)~\citep{liu2023algorithm}, which employs an evolutionary framework to generate optimization algorithms without model training and with minimal manual algorithm design. This paradigm gained wider popularity in 2024 with FunSearch~\citep{romera2024mathematical}, which demonstrated the use of LLMs to solve the cap set and bin packing problems, after which LLM-driven optimization techniques were widely applied across various fields. Evolution of Heuristics (EoH)~\citep{liuevolution} extends AEL by co-evolving algorithms together with the natural-language \emph{thoughts} used to generate the code within a predefined algorithmic skeleton, and addresses problems such as the traveling salesman problem and flow shop scheduling problem. More recently, AlphaEvolve~\citep{novikov2025alphaevolve} extended this paradigm from evolving a single function to evolving entire codebases, delivering state-of-the-art results on matrix multiplication, open mathematical problems, and Google's production infrastructure. ShinkaEvolve~\citep{lange2026shinkaevolve} targets the sample inefficiency of such frameworks through adaptive parent sampling, novelty-based rejection sampling, and bandit-based LLM selection, matching AlphaEvolve's circle-packing result with orders of magnitude fewer evaluations.}
In contrast, LLAMEA~\citep{van2024llamea} applies LLMs to directly generate metaheuristics, achieving competitive performance on the continuous black-box optimization. Similar techniques of these works have also been applied for multi-objective optimization~\citep{yao2025multi} and Bayesian optimization~\citep{liularge,li2025llameabolargelanguagemodel}. 

In SAT, the AutoSAT framework~\citep{sun2025automatically} pioneered optimizing SAT solvers with LLMs, followed by NVIDIA Research, which recently achieved significant improvement on the state-of-the-art performance~\citep{yu2025autonomous}. Meanwhile, more recent efforts have also targeted the domain of constrained PBO~\citep{li2025autopbo}. 

Unlike the pioneering FunSearch, which requires a significant number of LLM queries to evolve algorithmic code, recent work adopts an evolutionary procedure loop, as illustrated in Figure.~\ref{fig:workflow}. This loop starts with an initial prompt design and iteratively queries LLMs for new algorithm variants with online refined prompts. In the context of evolutionary computation~\citep{back2023evolutionary}, this process follows the stages of \textit{initialization, variation, evaluation,} and \textit{adaptation}. Notably, the \emph{variation} and \emph{adaptation} are driven by LLMs and heavily influenced by prompt design. A recent study highlights the essential role of search strategies~\citep{zhang2024understanding}, for example, the number of algorithms generated at each generation and the selection criteria directly affecting \emph{adaptation}. \textcolor{black}{However, recent evidence also shows that simple random-sampling baselines can be competitive with complex code-evolution methods on several optimization tasks~\citep{gideoni2025random}, suggesting the importance of problem formulation and the prompt-defined search space alongside the search strategy itself.} Nevertheless, prompt design remains the most distinct aspect across applications of LLM-driven optimization, determining how effectively LLMs guide the search process.

\subsection{Attributing input prompts of Decoder-only LLMs}
State-of-the-art LLMs typically allow up to hundreds of thousands of input tokens and operate in an excessively black-box manner. The long context window not only enables rich task descriptions from users, but also raises two questions: \textit{Do we need all these input tokens? Do all input tokens contribute equally and positively?} To address these concerns, token-wise (feature) attribution methods are proposed to assist human understanding of the contributions of input tokens by providing comparable token-to-token relevance scores between an input prompt and its corresponding LLM's output. This type of explanation falls into local explanations and can be divided into four categories regarding their fundamental strategies~\citep{zhao2024explainability,schneider2024explainable}: Perturbation-based methods~\citep {li2016visualizing,wu2020perturbed}; Surrogate-based methods~\citep{ribeiro2016should,kokalj2021bert}; Gradient-based methods~\citep{sundararajan2017axiomatic,enguehard2023sequential}, and decomposition-based~\citep{montavon2019layer}. The cutting-edge methods tailored for LLMs, including AttnLRP~\citep{achtibat2024attnlrp}, Progressive Inference~\citep{kariyappa2024progressive}, and JoPA~\citep{chang2025jopa}, are essentially building upon multiple strategies. To deal with complex input-output pairs in LLM-driven optimization, we use AttnLRP as the explainer in this paper. It is fundamentally grounded in layer-wise relevance propagation~\citep{montavon2019layer} and empowered by gradient information, producing sparse and faithful explanations with fast inference.

\subsection{Benchmarking Black-box Optimization}
While researchers work on automated algorithm design to achieve competitive performance with minimal development cost, benchmarks are also extensively studied in BBO to better understand algorithms' behavior across different types of problems. One of the main goals is to provide comprehensive and fair comparisons of algorithms across diverse problem sets from multiple perspectives (e.g., considering multiple performance measures)~\citep{bartz2020benchmarking}. The resulting benchmark data serve as valuable resources for the learning process in traditional automated approaches, such as algorithm configuration and algorithm selection~\citep{schede2022survey,kerschke2019automated}.
Several platforms have become commonly accepted in the BBO community. In practice, bbob and its variants provide well-established problem suites for continuous BBO, supported by the COCO platform~\citep{hansen2021coco}. More recently, a problem suite for pseudo-boolean BBO (pbo) has been proposed, accompanied by the IOHprofiler platform~\cite{doerr2020benchmarking}. Another platform, Nevergrad, integrates these problems and provides extensive algorithms as well as an automated algorithm selector (NGopt) based on problem meta information~\citep{bennet2021nevergrad}.  Furthermore, both problem suites, pbo and bbob, are embedded in a novel BBO benchmark framework Bencher~\citep{papenmeier2025bencher}.

\section{Benchmarks}
\label{sec:bench}
All experiments in this paper are evaluated on two benchmark suites that are widely used in the BBO community to ensure fair comparisons, pbo~\citep{doerr2020benchmarking} and bbob~\citep{hansen2021coco}. The pbo suite contains 23 pseudo-boolean optimization problems, while the bbob suite contains 24 continuous optimization problems in their standard settings. Both suites are well-established and have made significant contributions to algorithm development in their respective domains. In addition, these benchmark suites have attracted attention beyond BBO and are also widely applied in other domains such as general benchmark~\citep{papenmeier2025bencher}, learning~\citep{mameta}, and algorithm configuration~\citep{ye2022automated,li2024pretrained,song2025reinforced}. More details of the benchmark problems are provided in Appendix~\ref{appendix:bench}.

Performance measures are a crucial aspect in benchmarking and analyzing the behavior of algorithms. Unlike traditional indicators, considering the best-found solution within a fixed time or the time required to reach a specific target, anytime performance measures have recently been widely accepted in the BBO community~\citep{hansen2022anytime,wang2022iohanalyzer}. In this work, we evaluate all BBO algorithms using the (approximated) area under the ECDF curve (AUC). Given a set of time points $T$ and a set of target values $\Phi$, the ECDF value of an algorithm at time $t \in T$ is defined as the fraction of target values in $\Phi$ that are worse than the best-found fitness obtained by $A$ up to time $t$. The AUC value of $A$ is then computed as the aggregation of ECDF values across all $t \in T$. In this paper, we normalize it by the size of $T$. The specific settings of $T$ and $\Phi$ used in our experiments are reported in Section~\ref{sec:bag experiment} and Appendix~\ref{appendix:bench}. Throughout this work, we evaluate algorithm performance on a problem by averaging the AUC over five problem instances (see~\citet{hansen2010comparing,doerr2020benchmarking}) to ensure robustness.

\section{\textcolor{black}{Analytical Tools}}
\subsection{\textcolor{black}{Revealing the impact of prompt elements with AttnLRP}}
\label{sec:attnlrp}
We apply AttnLRP~\citep{achtibat2024attnlrp}, an attention-aware feature attribution method that adapts the classic Layer-wise relevance propagation (LRP)~\citep{montavon2019layer}, to identify the contribution of input prompt tokens to the generated algorithms. LPR assumes that a function $f_j$ with input $N$ features $\bm{x} = x_i, i \in \{1,\ldots,N\}$ can be decomposed into individual contributions of each feature $R_{i \leftarrow j}$, which is called \emph{relevances}. $R_{i \leftarrow j}$ quantifies how much of the output $j$ is attributed to input feature (token) $i$, and LRP calculates the relevance of feature $i$ by summing across all outputs, $R_i = \sum_j R_{i \leftarrow j}$.
By treating a neural network as a layered directed acyclic graph, LRP denotes a neuron $j$ in layer $l+1$ as a function node $f_j^{l+1}$.  It enforces a \emph{conservation property}, such that scores are redistributed backwards layer by layer while preserving their total values: 
\begin{equation}
\label{eq:lrp relevance}
R^{l} = \sum_i R_i^l =  \sum_{i,j} R^{(l,l+1)}_{i \leftarrow j} =  \sum_j R_j^l= R^{l+1}
\end{equation}

\textcolor{black}{
Starting from the next-token logit, AttnLRP propagates relevance scores
backward through the stacked transformer layers, applying specialized rules to each submodule (e.g., linear layers, attention layers, and normalization layers). Further details can be found in the original paper ~\citep{achtibat2024attnlrp}.
}
\subsection{\textcolor{black}{Exploring temporal dependency of code with CodeBLEU}}
\textcolor{black}{We employ CodeBLEU~\citep{ren2020codebleu} to analyze the long-term influence of an \textit{ancestor} code of benchmark algorithm to future generated heuristics. 
CodeBLEU is a metric for measuring the similarity (relevance) between two pieces of code. It was originally proposed to evaluate the logical and structural quality of generated code by comparing it with a reference implementation. Given a reference code snippet ($x$) and a generated code snippet ($y$), CodeBLEU computes a weighted sum of four sub-metrics derived from $y$ and $x$.} 
\textcolor{black}{\begin{equation}
\label{eq:CodeBLEU}
\begin{aligned}
\CB(y,x)
&= \lambda_1\,B(y,x)+\lambda_2\,B_w(y,x) \\
   &\quad +\lambda_3\,A(y,x)+\lambda_4\,D(y,x).
\end{aligned}
\end{equation} with $\lambda_i\ge 0,\ \text{and} \sum_{i=1}^{4}\lambda_i=1$.}
\textcolor{black}{These include: 1) BLEU $B(y,x)$, which measures the percentage of n-grams\footnote{A contiguous sequence of n code tokens taken from a larger code sequence.} overlapped between $x$ and $y$~\citep{papineni2002bleu};
2) weighted n-gram matching of BLEU $B_w(y,x)$, which emphasizes the influence of key tokens such as data types and keywords.
\textcolor{black}{Let $g^{(n)}_i(y)$ be the $i$-th n-gram of $y$, and $\omega^{(n)}_i$ its weight,
\[
B_w(y,x)=\mathrm{BP}(y,x)\exp\!\Big(\sum_{n=1}^{N}\eta_n\log p^{(w)}_n(y,x)\Big), \text{with}
\]
\[
p^{(w)}_n(y,x)=
\frac{\sum_i \omega^{(n)}_i\,\cl\!\big(g^{(n)}_i(y);x\big)}
{\sum_i \omega^{(n)}_i\,\cnt_y\!\big(g^{(n)}_i(y)\big)}.
\] Here $\cnt_z(h)$ is the counts of an n-gram $h$ in sequence $z$, $\cl(\cdot;x)$ caps counts by their maximum occurrence in $x$, and $\eta_n$ are the n-gram weights. $\mathrm{BP}\in (0,1]$ is a brevity penalty that exponentially downweights the score when $\lvert y \rvert < \lvert x \rvert$ and equals 1 otherwise;}
3) abstract syntax tree (AST) matching $A(y,x)$ captures structural consistency across code lines,
\textcolor{black}{
\[
A(y,x)=\frac{\sum_{s\in\mathcal{S}(y)} \min\{\cnt_{\mathcal{S}(y)}(s),\cnt_{\mathcal{S}(x)}(s)\}}
{\sum_{s\in\mathcal{S}(x)} \cnt_{\mathcal{S}(x)}(s)}
\]
with $\mathcal{S}(c)$ denotes the multiset of all AST subtrees of code $c$ after dropping identifier leaves, and $\cnt_{\cdot}(\cdot)$ is defined as before;} and 4) data-flow matching $D(y,x)$, which evaluates semantic similarity based on variable dependencies,
\textcolor{black}{
\[
D(y,x)=\frac{\sum_{f\in\mathcal{F}(y)} \min\{\cnt_{\mathcal{F}(y)}(f),\cnt_{\mathcal{F}(x)}(f)\}}
{\sum_{f\in\mathcal{F}(x)} \cnt_{\mathcal{F}(x)}(f)}.
\] where $\mathcal{F}(c)$ is a multiset of data-flow items extracted from code $c$, after uniformly renaming its variables to $var_0, var_1, ...$, in appearing order. In summary, CodeBLEU is a asymmetric score in its arguments that can capture syntactic and semantic similarity between two code pieces.}}

\section{Token-wise Analysis of Prompt Design in LLM-driven \textcolor{black}{AAD for} BBO}
\label{sec:token importance analysis}
In the evolutionary procedure of searching for improved algorithms within LLM-driven optimization approaches, querying LLMs for new algorithms plays an essential role in the \emph{variation} step, as illustrated in Figure~\ref{fig:workflow}. Designing more effective search strategies for LLM-driven approaches, therefore, requires a deeper understanding of how LLM produces new algorithms. Such insights are not only valuable for explaining the evolutionary procedure but also for enabling precise control over the search process, eventually leading to better and robust results. For example, restart strategies and diversity control are two common techniques in evolutionary computation, highlighting this need for proper technique design. Restart strategies help escape from local optima by reinitializing the search state, while diversity control ensures the exploration of different solutions to prevent premature convergence. Current attempts at using these advanced techniques rely on prompt engineering of \emph{linguistic task descriptions}, such as \emph{``creating a novel algorithm''} or \emph{``exploring new heuristics''}~\citep{liuevolution,van2024llamea}.
However, a systematic understanding of how prompts influence code generation in LLM-driven optimization remains limited.

To address this gap, we investigate in this section the research question \emph{Q1: How does the prompt design affect the generated algorithmic code?}, by utilizing AttnLRP to analyze the token-wise contribution of the prompt to the algorithmic code generated by LLMs. 

\textcolor{black}{Decoder-only LLMs generate each token based on the input and previously generated tokens. For each output token \(j \in T_{\mathrm{out}}\), we therefore compute relevance over this full context, but retain only the relevance attributed to input tokens \(i \in T_{\mathrm{in}}\), excluding special and chat-template tokens.}
\textcolor{black}{We obtain the token relevance (importance) through utilizing AttnLRP (see section~\ref{sec:attnlrp}), for each output token $j$, AttnLRP assigns an unbounded signed relevance score $R^{0}_{i,j}$ to the input token $i$, capturing its} \textcolor{black}{influence on} \textcolor{black}{the generation of $j$.}
To fairly compare the \textcolor{black}{relevance} of input tokens with respect to \textcolor{black}{the generated code tokens $\mathrm{T_{code}} \subseteq \mathrm{T_{out}}$, i.e., the tokens inside the code block of the response, we normalize the relevance magnitudes per output token and average them over $\mathrm{T_{code}}$}:
\begin{equation}
\label{eq: aggregate relevance}
\textcolor{black}{R_{i}=\frac{1}{|\mathrm{T_{code}}|}\sum_{j \in \mathrm{T_{code}}} \widehat{R}^{0}_{i,j},\ \text{with}\ \widehat{R}^{0}_{i,j} = \frac{\lvert R^{0}_{i,j}\rvert}{\sum_{k \in \mathrm{T_{in}}} \lvert R^{0}_{k,j}\rvert}}
\end{equation}
\textcolor{black}{Using magnitudes (absolute values) captures how strongly a token influences the output, whether it raises or lowers the logit, and the per-token normalization gives every generated code token equal weight. Since $\sum_{i \in \mathrm{T_{in}}} R_i = 1$, the} final scores are comparable across all input tokens.

\begin{figure}
\centering
\includegraphics[width=0.48\textwidth]{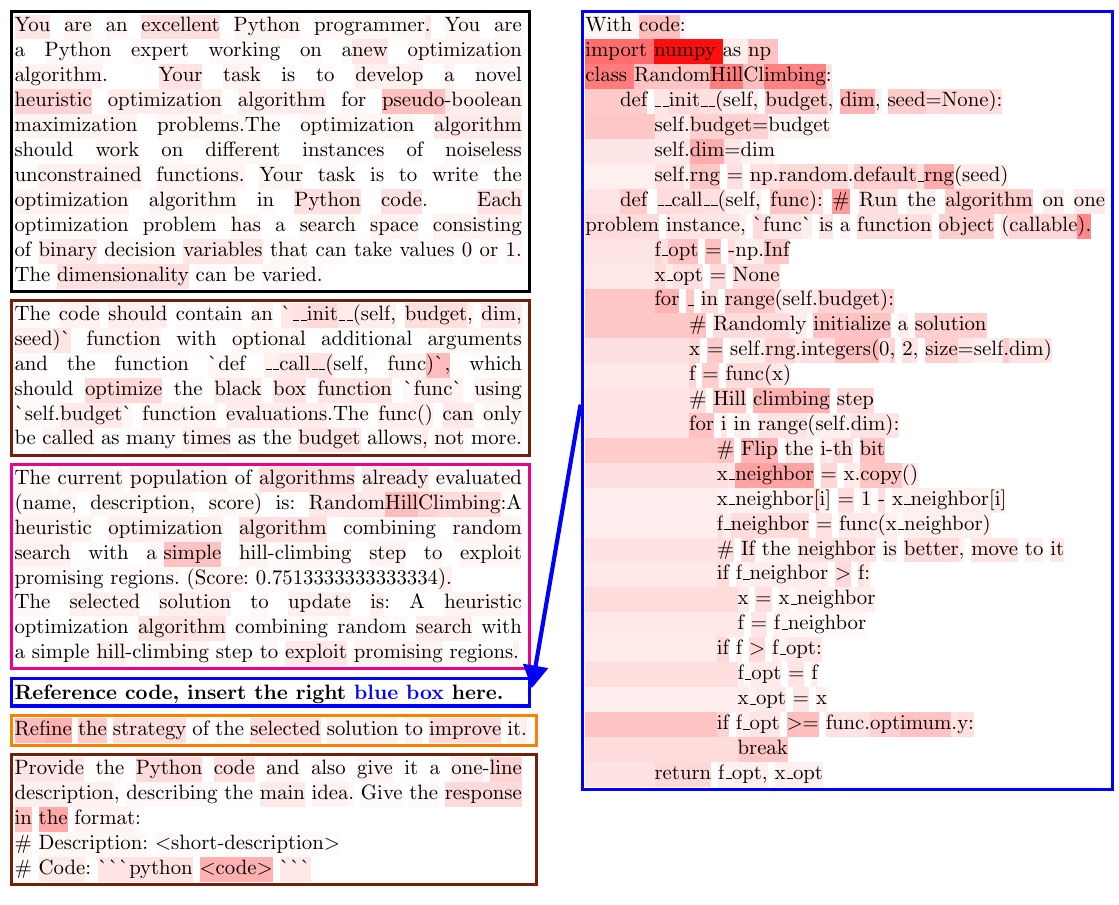}
\caption{The heatmap of the token-wise relevance of a given prompt to its corresponding newly generated algorithmic code. The result is obtained on an instruction-tuned 27b Gemma 3 LLM using the AttnLRP explainer. Darker shading indicates higher aggregated relevance scores, thus more important.}
\label{fig:heatmap}
\end{figure}


\begin{table*}
\caption{\textcolor{black}{Mean and standard deviation of relevance scores ($\times10^{-3}$) for different parts of prompts. The reported mean and standard deviation are aggregated over the conversations corresponding to five problems of the pbo suite for each LLM model. Relevance scores are normalized such that the sum of all token-level scores within a conversation equals 1. Consequently, the reported means are averages over tokens and do not sum to 1 across prompt parts. For each model, the most relevant prompt component is shown in \textbf{bold}, and the second most relevant is \underline{underlined}.}}
\label{tab:relevance}
\resizebox{\textwidth}{!}{%
\begin{tabular}{lccccc|cc}
\toprule
Model          & Task Description  & Strategy          & Expected Output   & Note              & Parent Heuristic(s) & Code Description  & Linguistic Description \\ \midrule
Qwen3 8b       & \underline{1.171 $\pm$ 0.374} & 1.090 $\pm$ 0.452 & 0.929 $\pm$ 0.389 & 0.146 $\pm$ 0.036 & 1.138 $\pm$ 0.636   & \textbf{1.232 $\pm$ 0.672} & 0.539 $\pm$ 0.145      \\
Qwen3 14b      & 0.688 $\pm$ 0.096 & \textbf{1.068 $\pm$ 0.300} & 0.893 $\pm$ 0.125 & 0.144 $\pm$ 0.034 & 0.863 $\pm$ 0.203   & \underline{0.901 $\pm$ 0.231} & 0.594 $\pm$ 0.080      \\
Qwen3 32b      & 1.139 $\pm$ 0.250 & 1.103 $\pm$ 0.327 & 1.013 $\pm$ 0.319 & 0.210 $\pm$ 0.043 & \underline{1.184 $\pm$ 0.499}   & \textbf{1.273 $\pm$ 0.517} & 0.703 $\pm$ 0.166      \\
\midrule
\midrule
Gemma3 4b  & 1.142 $\pm$ 0.757 & 0.790 $\pm$ 0.312 & 0.672 $\pm$ 0.402 & 0.135 $\pm$ 0.050 & \underline{1.547 $\pm$ 0.588}   & \textbf{1.684 $\pm$ 0.555} & 0.519 $\pm$ 0.385      \\
Gemma3 12b & 0.500 $\pm$ 0.285 & \textbf{1.501 $\pm$ 0.266} & 0.481 $\pm$ 0.137 & 0.129 $\pm$ 0.059 & 0.819 $\pm$ 0.215   & \underline{0.875 $\pm$ 0.228} & 0.355 $\pm$ 0.056      \\
Gemma3 27b & 0.507 $\pm$ 0.130 & \textbf{2.531 $\pm$ 0.539} & 0.611 $\pm$ 0.120 & 0.292 $\pm$ 0.093 & 0.954 $\pm$ 0.181   & \underline{1.019 $\pm$ 0.215} & 0.554 $\pm$ 0.104      \\ 
\bottomrule
\end{tabular}%
}
\end{table*}

\paragraph{Experiments}
We compute token-wise contributions for pairs of prompts and outputs in an elitist evolutionary procedure, where LLMs generate a new algorithm code at each iteration. If a generated algorithm outperforms the current best, we adapt the prompt by using its code.
Specifically, we leverage AttnLRP to explain the outputs of \textcolor{black}{six open-source LLMs, including three instruction-tuned Gemma 3 models (4B, 12B, and 27B) and three Qwen 3 models (8B, 14B, and 32B)}, for automatically designing algorithms to solve \textcolor{black}{five classic PBO problems (F1, F2, F19, F20, and F21 in pbo}, see details in Appendix~\ref{appendix:bench}). Figure~\ref{fig:heatmap} presents a heatmap of the prompt relevance to the corresponding generated algorithmic code, clearly showing that the code-related content obtains a strong influence. Additional heatmaps of prompt-code pairs in our public repository also demonstrate consistent behavior.

To quantify the contributions of different prompt elements, we partition prompts into five components: \textit{Task Description, Strategy, Expected Output, Note, and Parent Heuristics}, following the setting of prior work~\citep{liuevolution}.
\textcolor{black}{Using the token scores $R_i$ from Equation~\ref{eq: aggregate relevance}}, the (average) \textcolor{black}{relevance} of a tokenized prompt component $\mathrm{C} \textcolor{black}{\subseteq} \mathrm{T_{in}}$ is simply the average: \textcolor{black}{$R_\mathrm{C} = \frac{1}{\lvert \mathrm{C} \rvert}\sum_{i\in \mathrm{C}} R_i$}.
Average \emph{relevance} is preferred to mitigate the size difference between various prompt components.
Table~\ref{tab:relevance} reports this average tokens relevance of each component across all prompt-code pairs \textcolor{black}{for five problems in the PBO set, as generated by six LLMs of two model families.}
For \emph{Parent Heuristic(s)}, which consists of both code and linguistic description, we further separate it into two parts.
\emph{Strategy} is defined as a short instruction indicating whether to \emph{refine} the provided code or \emph{create} a novel algorithm.
\emph{Note} specifically contains the fitness value (e.g., score) obtained by the parent code.
Additional details are explained in Appendix~\ref{appendix:prompt}. 

The results demonstrate that the \emph{code description} and the associated \emph{strategy} obtain the strongest influence on the produced algorithmic code, consistently across \textcolor{black}{six} tested LLMs. Meanwhile, the linguistic descriptions of tasks and the current algorithms contribute less to the code-related content. In particular, the fitness value of the parent code, which is important for the \emph{selection} of evolutionary procedures, does not exhibit a significant impact on the behavior of LLMs.

Overall, our extensive experiments demonstrate that \emph{among all prompt components, the provided code example and its associated strategy instruction obtain the most significant impact on the output algorithms generated by LLMs}

\section{Guiding LLMs toward Specific Search Region}
\label{sec:refine experiments}
Motivated by the significant influence of code-related content, we investigate here the research question: \emph{Q2: Can we control LLMs to explore specific regions of algorithms?}

To address this question, we compare algorithms produced when LLMs are prompted with different example code. Each run begins with a distinct example code, and we iteratively query LLMs to refine the current best code following the elitist strategy (see the prompt in Appendix~\ref{appendix:prompt}). Our hypothesis is that, by providing specific example code, using LLMs to refine algorithmic code will behave as a neighborhood search, constraining exploration to a local region of the search space. \textcolor{black}{It is worth mentioning that our focus lies in improving the LLM-driven optimization process itself while controlling the evolutionary search towards better algorithms more efficiently. Specifically, we construct prompts that integrate example code to guide LLMs towards generating promising, diverse algorithm candidates. This differs from the \emph{in-context learning}~\citep{dong2024survey}, which focuses on adapting LLMs' behavior with a single prompt rather than exploring the search space of algorithms through optimization techniques.}

We implement and test a refinement strategy (Refine:A$i$) \textcolor{black}{on the (1+1)-LLaMEA~\citep{van2024llamea}} that iteratively queries LLMs to only \emph{refine} the current best algorithmic code, starting with a given example code A$i$. \textcolor{black}{It is worth noting that this strategy is closely related to the LLM-driven Heuristic Neighborhood Search (LHNS) method~\citep{xie2025llm}, which improves a given heuristic via iterative removal and reconstruction of its code segments. Building on this connection, we further examine the impact of providing strong prior code to LHNS by initializing it with a diverse set of \textit{example} metaheuristics.} Experiments are tested on pbo problems, and we test different settings by using the top five algorithms A$i$, $i\in\{1,\ldots,5\}$ for each problem based on the benchmark data in~\citet{doerr2020benchmarking}.

Figure~\ref{fig:refine-f10} presents the results for \textcolor{black}{OneMax with ruggedness (F10 of pbo}, see Appendix~\ref{appendix:bench}). Additional experimental results are available in Appendix~\ref{appendix:exp-ref}.  We can observe that Refine:A$i$ methods yield distinct performance trajectories, which confirms that different strong prior code can steer LLMs towards different search regions. Meanwhile, refining certain algorithms can outperform the 
\textcolor{black}{original baselines (LHNS and LLaMEA)}, achieving the best AUC across three tested LLMs. However, identifying the suitable example code that can guide LLMs toward the optimal performance remains an open question.

Overall, the results indicate that \emph{embedding specific strong prior code can constrain the search region of LLMs}, while raising the question of how to identify and utilize such code. 

\begin{figure}
\centering

\subfloat[LHNS\label{fig:refine-f10-lh}]{
    \includegraphics[width=0.972\linewidth]{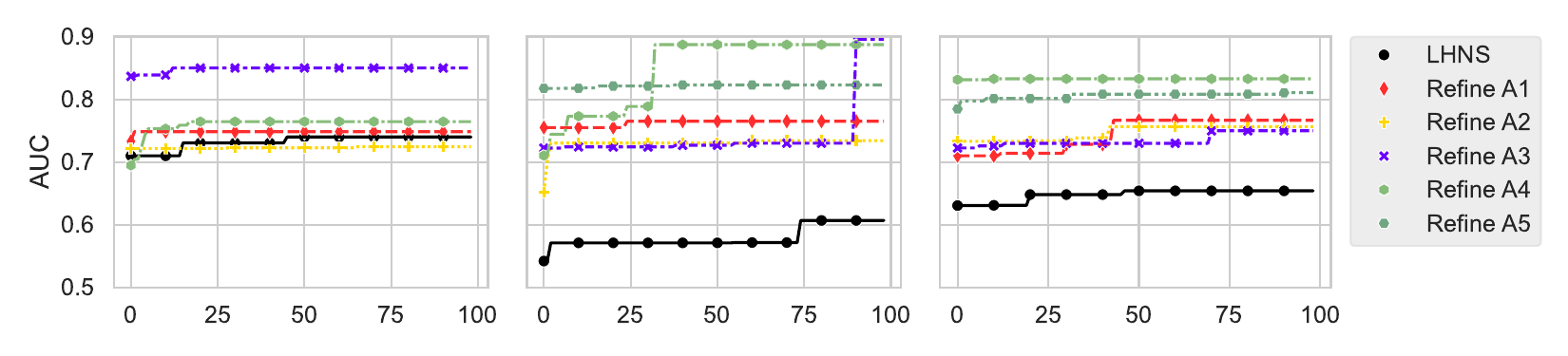}
}

\subfloat[LLaMEA\label{fig:refine-f10-ll}]{
    \includegraphics[width=0.972\linewidth]{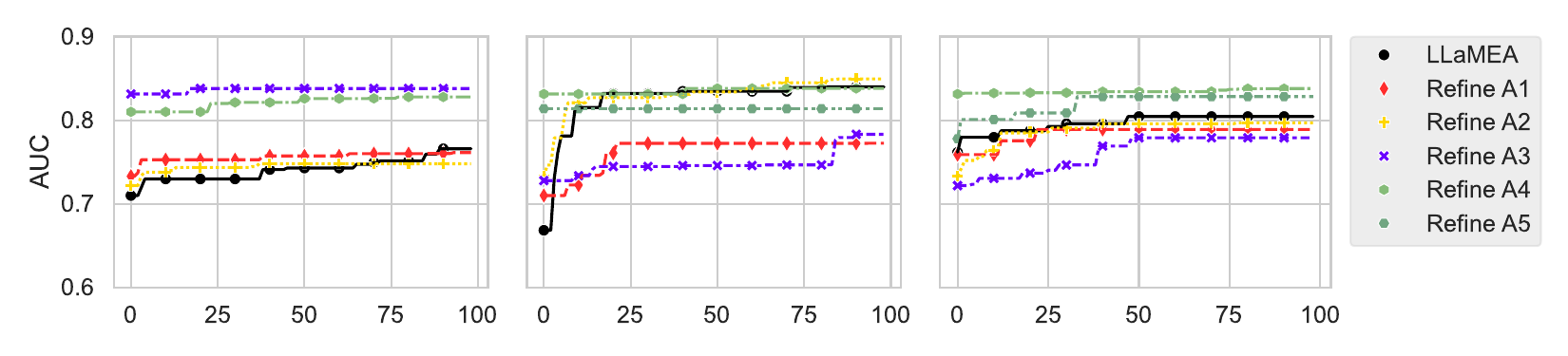}
}
\caption{\textcolor{black}{Convergence process of refinement-only LHNS and LLaMEA methods for the OneMax problem with ruggedness. $x$-axis represents the number of algorithms generated by the LLM, and $y$-axis indicates the best-so-far AUC value. Results are from using Gemini, GPT, and Qwen, respectively (from Left to Right). }}
\label{fig:refine-f10}
\end{figure}

\section{A Benchmark-Guided Approach}
\label{sec:bag experiment}
To address the research question \emph{Q3: how can we effectively control the search process of LLM-driven optimization approaches?}, we propose a benchmark-assisted guided evolutionary approach (BAG) following the evolutionary procedure as described in Algorithm~\ref{alg:LLM-bbo}. 
Specifically, the prompt $P(\mathcal{A})$ used to query the LLM is consistently embedded with an example code of the current algorithm $\mathcal{A}$. A benchmark set of algorithms, denoted as $\mathbf{{A}}_{bench}$, is leveraged to guide the prompt design during both the \emph{Initialization} and \emph{Adaptation} steps. \textcolor{black}{Recall that determining the optimal algorithms for an arbitrary BBO problem is challenging, and the conclusion may vary considering different performance measures. Therefore, to avoid being trapped in a particular algorithmic pattern, we work with a set of benchmarks, $\mathbf{{A}}_{bench}$, rather than relying on a single algorithm.} 
Note that the algorithm refers to its code in this context. The BAG method is detailed as follows.
\begin{itemize}
    \item We adopt the (1+1) elitist search strategy as proposed in LLaMEA, which generates a new algorithm at each iteration. Its effectiveness has been demonstrated by theoretical and empirical results~\citep{witt2006runtime,droste2002analysis,hoos2018stochastic} on mutation-based evolutionary algorithms and local search. These methods are analogous to our motivation of guiding LLMs to explore the neighborhood of the provided code example.
    \item \textcolor{black}{\emph{Initialization} is performed using a promising example code randomly selected from $\mathbf{A}_{bench}$.} 
    \item At each iteration, the LLM is queried either to \emph{refine} the current best algorithm $\mathcal{A^*}$ or to \emph{create} a novel algorithm with equal probability (lines 11-14), suggested by prior LLM-driven approaches~\citep{liuevolution,ye2024reevo,van2024llamea}. The \emph{refine} operator incrementally optimizes $\mathcal{A^*}$ under the elitist strategy, while the \emph{create} operator introduces diversity, though it heavily relies on the generative capacity of LLMs.
    \item To further exploit benchmark knowledge, every $q$ iterations the LLM is specifically queried to \emph{refine} an algorithm randomly selected from $\mathbf{A}_{bench}$ (lines 8-9). This encourages exploration of promising regions identified in prior benchmark while maintaining diversity.
\end{itemize}

The benchmark algorithm selection follows a sampling without replacement. Once all elements of $\mathbf{A}_{bench}$ have been selected, the process restarts with the full set $\mathbf{A}_{bench}$. 
$rand \in [0,1]$ denotes a value sampled uniformly at random. 
If a generated algorithm $\mathcal{A}$ fails to execute or does not return a solution within the timeout, its fitness $F(\mathcal{A})$ is assigned the worst possible value. \textcolor{black}{We set $q=10$ as suggested by the experimental analysis in Appendix~\ref{appendix:q}.}

\begin{algorithm}[!htb]
\caption{A Benchmark-assisted LLM-driven BBO}
\label{alg:LLM-bbo}
\textbf{Input:} A single problem and a fitness measure $F$, a set of benchmark algorithms $\mathbf{A}_{bench}$, a prompt template $P$, an LLM model, a frequency factor $q$, and a maximal budget $\mathcal{B}$ \textcolor{black}{for algorithm proposals}\;
\textbf{Initialization: } \textcolor{black}{Select} a benchmark algorithm code $\mathcal{A}^*$ from $\mathbf{A}_{bench}$ \textcolor{black}{uniformly at random}, generate an algorithm code $\mathcal{A}$ by querying the LLM with the template prompt $P(\mathcal{A}^*)$
(\textbf{\emph{Initialization}})\;
Evaluate the initial algorithm by $F(\mathcal{A^*})$,\;
Evaluate the algorithm by $F(\mathcal{A})$, $t \leftarrow 1$\;
\lIf{$F(\mathcal{A})$ outperforms $F(\mathcal{A^*)}$}{$\mathcal{A^*} \leftarrow \mathcal{A}$}
\While{
$t < \mathcal{B}$
}{
(\textbf{\emph{Adaptation \& Variation}})\;
\eIf{$t\mod q = 0$ } 
{
Select an algorithm randomly from $\mathcal{A'} \in \mathbf{A}_{bench}$\;
Generate an algorithm $\mathcal{A}$ by querying the LLM with the prompt $P(\mathcal{A}')$ to \emph{refine} $\mathcal{A'}$\; 
}
{
\eIf{$rand < 0.5$}
{
Generate an algorithm $\mathcal{A}$ by querying the LLM with the prompt $P(\mathcal{A}^*)$ to \emph{refine} $\mathcal{A}^*$\;
}
{
Generate an algorithm $\mathcal{A}$ by querying the LLM with the prompt $P(\mathcal{A}^*)$ to \emph{create a novel algorithm}\;
}
} 
Evaluate the generated algorithm by $F(\mathcal{A})$, $t \leftarrow t+1$ (\textbf{\emph{Evaluation}})\;
}
\lIf{$F(\mathcal{A})$ outperforms $F(\mathcal{A^*)}$}{$\mathcal{A^*} \leftarrow \mathcal{A}$}
\textbf{Output:} Return the best algorithm $\mathcal{A^*}$.
\end{algorithm}

\paragraph{Technique differences from existing methods} The key innovation of our proposed BAG framework lies in leveraging prior benchmark knowledge to effectively guide LLMs toward specific regions and discover improved algorithms. To this end, BAG applies a (1+1) mutation-based elitist scheme, chosen to clearly highlight this motivation. By contrast, EoH and ReEvo were originally introduced with population-based schemes that incorporate both mutation and crossover-like operators on algorithmic code, and LLaMEA exhibits both the (1+1) and population-based strategies. \textcolor{black}{We also note that a recent work, MCTS-AHD~\citep{zheng2025monte}, replaces the population update with a tree-based search, where heuristics are treated as nodes in a Monte Carlo Tree Search and entire root-to-leaf evolutionary trajectories are preserved and used during generation. Regardless of the difference, existing approaches mainly adapt the \emph{Strategy} component of prompts to steer the search process.} For example, EoH is motivated by the idea that \emph{the evolution of thought, a linguistic description representing a high-level idea of a heuristic, is important}~\citep{liuevolution}.
However, BAG emphasizes adapting \emph{code} itself as the core of prompt design and search guidance, while relying on only two concrete strategies, refining the current algorithm or generating a novel one.

\paragraph{Experimental settings}
We compare BAG with EoH, \textcolor{black}{LHNS}, LLaMEA, \textcolor{black}{MCTS-AHD}, ReEvo, and~\textcolor{black}{ShinkaEvolve} across $47 \times 5$ problem instances, including 23 100-dimensional pbo problems and 24 5-dimensional bbob problems. 
The details of the benchmarks are described in Appendix~\ref{appendix:bench}. 
\textcolor{black}{By design, BAG utilizes benchmark knowledge that already exists for well-studied problem classes, while the baseline methods start from the task description and a random-search example.}
For the construction of $\mathbf{A}_{bench}$, we select the top five algorithms reported in the repository of~\citep{doerr2020benchmarking} for each pbo problem\textcolor{black}{, ranked by their AUC performance on the training instances}. Since no repository of standard formatted code that supports effective LLM interaction for bbob, we form $\mathbf{A}_{bench}$ using five widely adopted algorithms in continuous BBO: covariance matrix adaptation evolution strategy (CMA-ES)~\citep{hansen2016cma}, Cholesky CMA-ES~\citep{krause2016cma}, evolution strategy with cumulative stepsize adaptation~\citep{chotard2012cumulative}, differential evolution~\citep{das2010differential}, and particle swarm optimization~\citep{wang2018particle}.
\textcolor{black}{This set is identical for all 24 bbob problems and is not selected per function, so the bbob results do not rely on function-specific benchmark data. In both suites, $\mathcal{A}^*$ is chosen without being evaluated on the test instances.}

We apply the default configurations of 
\textcolor{black}{EoH, LHNS, LLaMEA, MCTS-AHD, and ReEvo}
as specified in their publications. \textcolor{black}{To ensure a fair comparison, we adapt ShinkaEvolve to our AAD task. Additional explanations are provided in the paragraph \textit{Setup of Software} of the Appendix~\ref{sec:app:setup}.}
Experiments are conducted with three LLMs: Google's \textcolor{black}{Gemini 2.5 Flash-Lite} (Gemini), OpenAI's GPT 5 Nano (GPT), and Alibaba's Qwen3 Coder Flash (Qwen) (see Appendix~\ref{sec:app:setup} for details). The cutoff time (i.e., maximal function evaluations) for evaluating algorithms on each problem instance is set to $10^4$ and $10^6$ for pbo and bbob, respectively, consistent with common practice for these two benchmarks~\citep{doerr2020benchmarking,hansen2021coco}. \textcolor{black}{The budget is defined as the number of algorithm proposals, limited to $100$, rather than the number of LLM queries. Since the compared LLM-driven AAD frameworks differ in how many LLM queries they need to produce one complete proposal, whereas each proposal corresponds to exactly one candidate solver evaluated on the objective function, this choice provides a fair and framework-agnostic comparison.}
The performance of algorithms (i.e., fitness $F$) is evaluated by AUC (see Appendix~\ref{appendix:bench}), $T$ is formed by 100 log-scaled samples from 1 to the cutoff time, and the target set is introduced in Appendix~\ref{appendix:bench}. Since this measure is defined in terms of function evaluations, the results reported in this paper are independent of the underlying hardware\textcolor{black}{~\citep{hansen2021coco,doerr2020benchmarking}. Nevertheless, each generated algorithm still has a timeout of 600 seconds per problem instance.} Additional information can be found in Appendix~\ref{sec:app:setup}.

\begin{table*}[!htb]
\caption{\textcolor{black}{Normalized best-achieved AUC results on 23 pbo problems and 24 bbob problems. We report the mean $\pm$ standard deviation (average rank).The best-performing entries in the ranking are shown in bold.}}
\label{tab:main result}
\centering
\resizebox{\textwidth}{!}{
\begin{tabular}{c|ccccccc}
\toprule
\multicolumn{8}{c}{Results on 23 pbo benchmark problems} \\
\midrule
LLM & BAG & EoH & LHNS & LLaMEA & MCTS-AHD & ReEvo & ShinkaEvolve\\
\midrule
Gemini & $\mathbf{0.967_{\pm 0.043}(2.125)}$ & $0.925_{\pm 0.076}(3.167)$ & $0.711_{\pm 0.224}(6.250)$ & $0.923_{\pm 0.070}(3.500)$ & $0.869_{\pm 0.087}(5.042)$ & $0.920_{\pm 0.126}(3.042)$ & $0.852_{\pm 0.144}(4.792)$ \\
GPT & $0.918_{\pm 0.088}(3.292)$ & $\mathbf{0.924_{\pm 0.086}(3.125)}$ & $0.780_{\pm 0.206}(5.292)$ & $0.897_{\pm 0.112}(3.625)$ & $0.861_{\pm 0.206}(4.250)$ & $0.892_{\pm 0.117}(3.875)$ & $0.860_{\pm 0.153}(4.542)$ \\
Qwen & $\mathbf{0.950_{\pm 0.083}(2.542)}$ & $0.943_{\pm 0.068}(2.958)$ & $0.823_{\pm 0.164}(5.917)$ & $0.921_{\pm 0.076}(3.958)$ & $0.912_{\pm 0.094}(4.125)$ & $0.882_{\pm 0.103}(5.042)$ & $0.914_{\pm 0.157}(3.417)$ \\
\midrule
\multicolumn{8}{c}{Results on 24 bbob benchmark problems} \\
\midrule
LLM & BAG & EoH & LHNS & LLaMEA & MCTS-AHD & ReEvo & ShinkaEvolve\\
\midrule
Gemini & $\mathbf{0.871_{\pm 0.173}(2.600)}$ & $0.796_{\pm 0.179}(3.120)$ & $0.493_{\pm 0.264}(6.200)$ & $0.880_{\pm 0.133}(2.920)$ & $0.768_{\pm 0.231}(3.920)$ & $0.793_{\pm 0.180}(3.280)$ & $0.464_{\pm 0.247}(5.960)$ \\
GPT & $\mathbf{0.891_{\pm 0.218}(1.960)}$ & $0.610_{\pm 0.287}(4.080)$ & $0.429_{\pm 0.294}(5.880)$ & $0.727_{\pm 0.294}(3.000)$ & $0.607_{\pm 0.267}(4.440)$ & $0.543_{\pm 0.293}(4.560)$ & $0.579_{\pm 0.282}(3.960)$ \\
Qwen & $\mathbf{0.959_{\pm 0.076}(1.920)}$ & $0.787_{\pm 0.198}(3.080)$ & $0.650_{\pm 0.225}(4.920)$ & $0.779_{\pm 0.189}(3.640)$ & $0.583_{\pm 0.275}(5.160)$ & $0.736_{\pm 0.258}(3.960)$ & $0.582_{\pm 0.306}(5.320)$ \\
\bottomrule
\end{tabular}
}
\end{table*}

\begin{figure*}[!htb]
\centering

\includegraphics[width=0.3\textwidth]{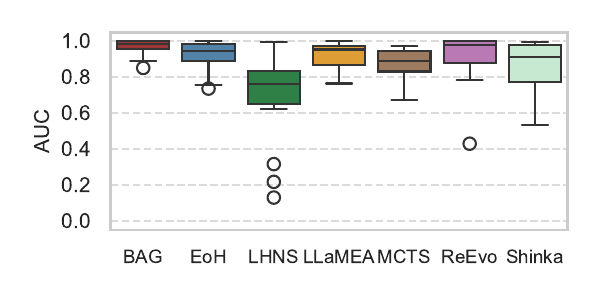}
\includegraphics[width=0.3\textwidth]{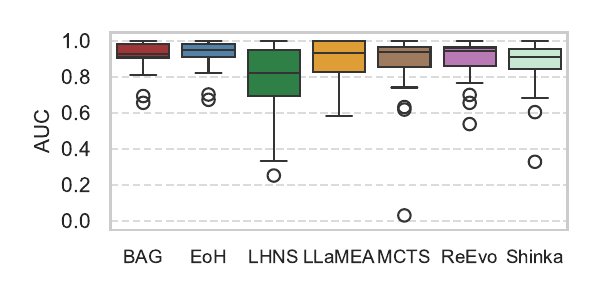}
\includegraphics[width=0.3\textwidth]{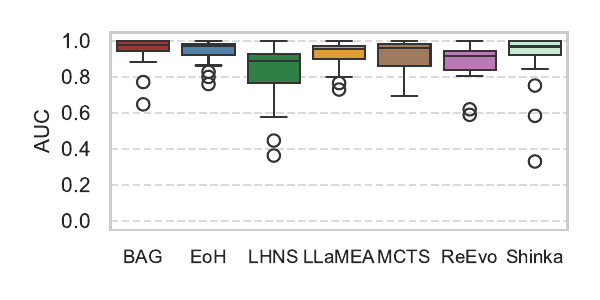}

\includegraphics[width=0.3\textwidth]{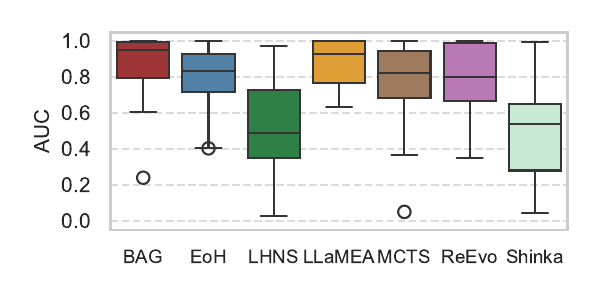}
\includegraphics[width=0.3\textwidth]{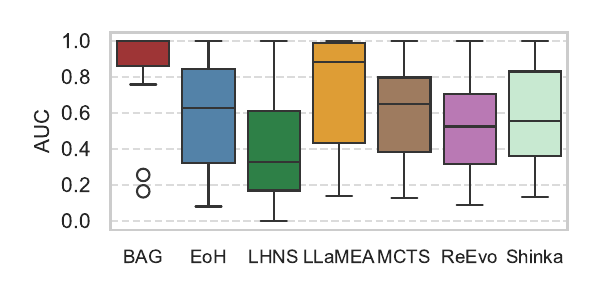}
\includegraphics[width=0.3\textwidth]{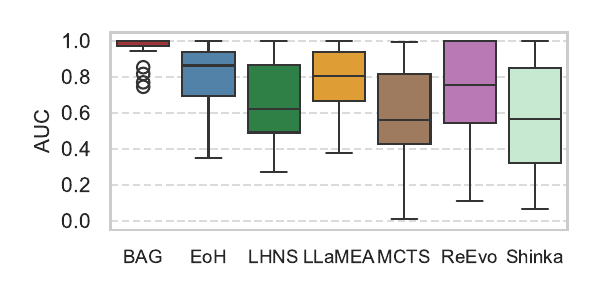}
\caption{\textcolor{black}{Boxplots of the best normalized AUC values obtained by the \textcolor{black}{seven} approaches on 23 problems of pbo (Top) and 24 problems of bbob (Bottom). The LLM-driven approaches have been tested on  Gemini, GPT, and Qwen, respectively (from Left to Right). For conciseness, MCTS-AHD and ShinkaEvolve are abbreviated as \textit{MCTS} and \textit{Shinka}, respectively.}}
\label{fig:boxplot}
\end{figure*}

\begin{figure}
\centering
\includegraphics[width=0.9\linewidth]{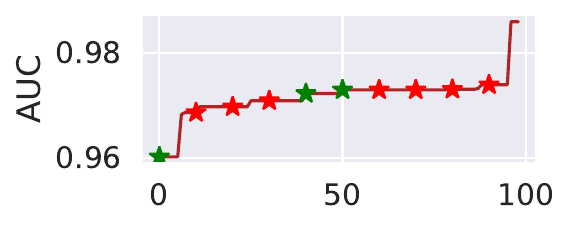}
\caption{An illustration of the contribution of queries \emph{refining} prior benchmark algorithms in BAG. The corresponding obtained results are marked by stars, and green indicates triggering improvements. Results are from using Gemini.}
\label{fig:refine-mark}
\end{figure}

\paragraph{Performance}
Table~\ref{tab:main result} presents the average normalized AUC and ranks of BAG and the compared LLM-driven approaches across the pbo and bbob suites, and Figure~\ref{fig:boxplot} displays boxplots of normalized AUC distributions. Normalization is performed as $\text{AUC}/\text{AUC}_{\text{best}}$, such that the best approach obtains a value of 1. \textcolor{black}{The standard deviation is taken across problems and therefore shows how consistently a method performs over a benchmark suite.} Detailed results for each problem, including AUC values and the convergence trajectories, are provided in Appendix~\ref{appendix:ref-com}.

\emph{Overall Performance.} 
\textcolor{black}{BAG achieves the best average rank in five of the six benchmark–LLM settings.} Specifically, it outperforms all baselines when using Gemini and Qwen for pbo, while it ranks second when using GPT, showing only a $-0.65\%$ gap in average AUC compared to EoH.
Importantly, BAG requires only a single LLM query to obtain a novel algorithm candidate before evaluation, whereas EoH relies on multiple (five) prompts to generate an algorithm candidate. 
\textcolor{black}{On bbob, BAG achieves the best average rank with all three LLMs. With GPT and Qwen, it improves the mean normalized AUC over the strongest baseline by 23\% and 22\%, respectively. With Gemini, BAG is within 1\% of LLaMEA while obtaining a better average rank.}
\textcolor{black}{Note that all methods are compared under the same amount of full algorithm proposals, i.e., candidate algorithms evaluated on the objective. Under the same amount of full algorithm proposals, BAG therefore consumes fewer LLM queries, and it has the potential to achieve even better results under an equal LLM query budget.}
\begin{figure}
    \centering
    \includegraphics[width=\linewidth, trim=0 1.5cm 0 0, clip]{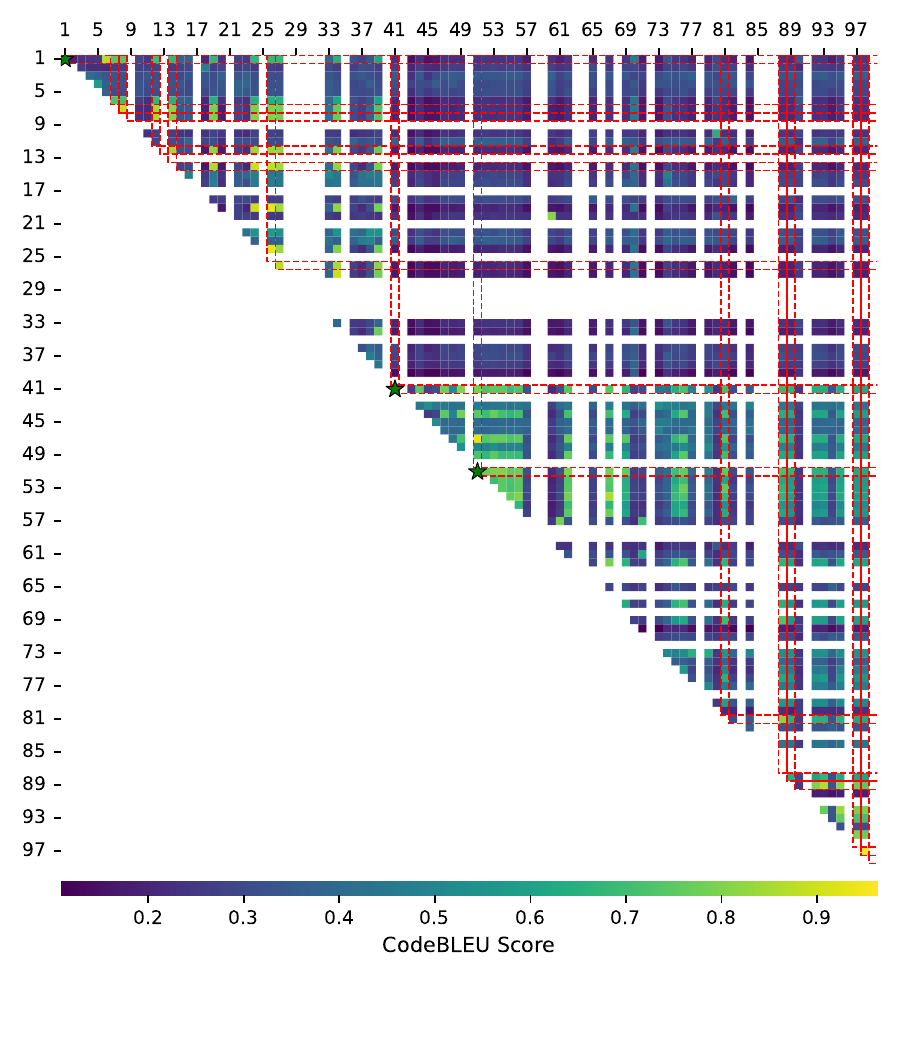}
    \caption{\textcolor{black}{
    CodeBLEU similarity scores between ordered pairs of generated algorithms (see Equation~\ref{eq:CodeBLEU}). Axes present time (i.e., the number of algorithms generated). Each column shows the similarity scores between a newly generated algorithm and all previous ones. Larger scores indicate higher relevance.}}
    \label{fig:code relevance}
\end{figure}
The competitive performance of BAG confirms that benchmark knowledge can effectively guide LLMs toward promising regions of search space.
Using prior benchmark algorithm code, BAG can initialize from a promising candidate (as presented in Figures~\ref{fig:convergence_pbo_gemini}-\ref{fig:convergence_bbob_qwen} in Appendix~\ref{appendix:ref-com}), ensuring final result quality and accelerating improvement. \textcolor{black}{The experimental results presented here are examined on 235 training problem instances ($47 \text{ problems } \times 5$ runs on different instances). Following the common practice of benchmarking BBO algorithms, we further evaluate the final generated algorithms on 20 unseen testing instances for each of the 47 benchmark problems, totaling 940 problem-instance pairs. Statistical significance tests show that BAG achieves more wins than losses against the six baselines in 35 out of 36 pairwise comparisons across the two suites and three backend LLMs. These results support the better instance-level transferability of BAG-generated algorithms, with detailed statistical results presented in Appendix~\ref{appendix:testResult}.}

\emph{Convergence Analysis.} Figure~\ref{fig:refine-mark} illustrates the convergence process of BAG on Sphere (F1 of bbob), where the star markers denote the fitness obtained from \emph{refining} a new benchmark algorithm at fixed intervals (line 9 in Algorithm~\ref{alg:LLM-bbo}). The three green markers indicate triggered improvements, evidencing the effectiveness of our design in BAG. \textcolor{black}{Apart from the convergence analysis in Figure~\ref{fig:refine-mark}, we also examine the long-term impact of prompt design by evaluating the similarity among generated algorithm code using the CodeBLEU~\citep{ren2020codebleu} metric (see Appendix~\ref{appendix:CodeBLEU}). 
Figure~\ref{fig:code relevance} presents the CodeBLEU scores for all ordered pairs of algorithms generated within the same evolutionary search procedure shown in Figure~\ref{fig:refine-mark}. We observe that the $41$st generated algorithm exhibits low similarity scores to all previously generated algorithms, as it is produced by introducing a new algorithm in $\mathbf{A}_{bench}$ for prompt design. However, subsequent algorithms show high similarity to this $41$st algorithm while obtaining low similarity to the first $40$ ones. A Similar pattern can be observed for the $51$st generated algorithm, which is also generated by introducing new example code. 
These results indicate that the benchmark algorithms we incorporate play a crucial role in guiding the evolutionary search and substantially contribute to the \textcolor{black}{strong} performance of our BAG method.}

\begin{figure}[htb]
\centering
\includegraphics[width=\linewidth]{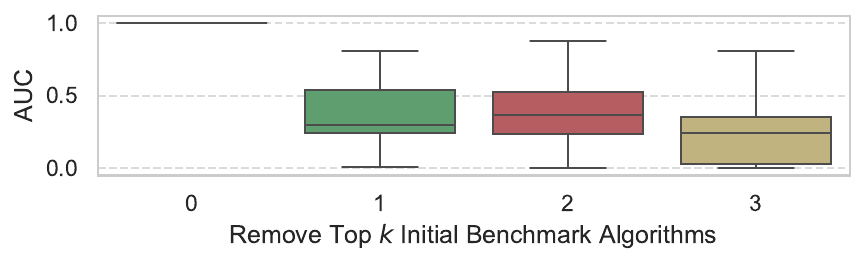}
    \caption{\textcolor{black}{Boxplots of the best normalized AUC values obtained by BAG with each of the top-$k$ example benchmark algorithms ($\mathbf{A}_{bench}$) \underline{removed}. The results are aggregated across three LLMs and six bbob problems. The leftmost box-plot ($k$=0) corresponds to the default setting, where no benchmark algorithms are removed.}}
    \label{fig:ablation k main}
\end{figure}

\emph{\textcolor{black}{Design Choices}}
BAG contains two design choices whose effects on performance are worth examining: the quality of the benchmark algorithm set $\mathbf{A}_{bench}$ and the frequency factor $q$ in Algorithm~\ref{alg:LLM-bbo}. We conduct sensitivity analysis of BAG to both choices on six bbob and seven pbo problems. Figures~\ref{fig:ablation k main} and~\ref{fig:ablation q main} report the results on bbob, while the full results, including pbo, are provided in Appendices~\ref{appendix:q} and~\ref{appendix:k} , respectively. Figure~\ref{fig:ablation k main} shows that BAG obtains both the highest average performance and the greatest robustness under the default setting ($k=0$) on the six bbob problems, where no example algorithms are removed and BAG has access to the complete set of five benchmark algorithms. Removing one or more of the best-performing example algorithms drastically degrades performance and robustness. Since the example algorithms are presented to the LLM precisely through the \textit{code descriptions}, this reinforces our main finding on the importance of that prompt component. Figure~\ref{fig:ablation q main} shows that BAG achieves the best and most robust performance at $q=10$, under which each of the five algorithms in $\mathbf{A}_{bench}$ is utilized approximately twice on average. \textcolor{black}{The two extreme choices of $q$ are simpler uses of $\mathbf{A}_{bench}$. With $q=100$, the search is initialized from $\mathbf{A}_{bench}$ but never returns to it. The refinement-only runs in Section~\ref{sec:refine experiments} are actually a further variant of this setting, starting from a single example algorithm. With $q=1$, every proposal refines an algorithm drawn from $\mathbf{A}_{bench}$, while the current $\mathcal{A}^*$ is never refined and no novel algorithm is created, this is equivalent to restarting from the benchmark set at every iteration. BAG with $q=10$ outperforms both extreme cases, suggesting that the gains come neither from the initial benchmark algorithm alone nor from repeatedly refining benchmark algorithms, but from combining refinement of the proposed solver with periodic injection of benchmark algorithms.} These two design choices jointly determine how well BAG can use the example algorithm set $\mathbf{A}_{bench}$. The results suggest that using a moderate $q$ between $5$ and $10$ with the full initial set ($k=0$) gives each example algorithm a fair chance to guide the search.

\begin{figure}[htb]
\centering
\includegraphics[width=\linewidth]{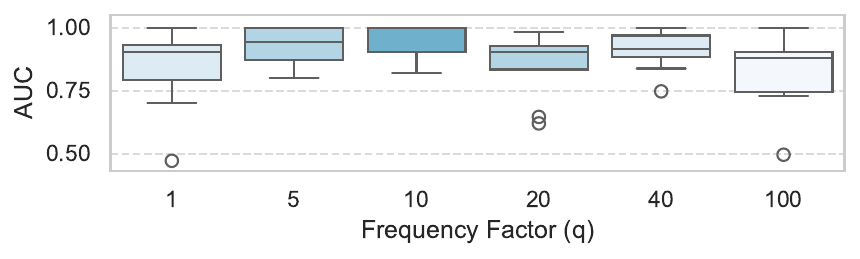}
    \caption{\textcolor{black}{Boxplots of the best normalized AUC values obtained by BAG with different frequency factor $q$. The results are aggregated across three LLMs and six bbob problems.}}
    \label{fig:ablation q main}
\end{figure}

Despite its simple search strategy design, BAG outperforms the state-of-the-art LLM-driven approaches, demonstrating potential for improvements with benchmark knowledge. For instance, the five benchmark algorithms used in our experiments are drawn from a limited dataset for each problem suite, while improvements could be expected by incorporating tailored code, as discussed in Section~\ref{sec:refine experiments}.
\textcolor{black}{Therefore, future work may explore integrating LLM-driven optimization with benchmarks such as Nevergrad}~\citep{bennet2021nevergrad}, which provides extensive algorithm collections but uses complex frameworks that hinder effective LLM interaction. 

In summary, BAG demonstrates that a benchmark-guided search strategy can substantially outperform existing LLM-driven optimization methods.~\textcolor{black}{We also note that BAG is fundamentally based on LLaMEA with periodic injection of benchmark knowledge. Comparative experimental results demonstrate that this injection effectively boosts LLaMEA's performance in BBO. More importantly, BAG is essentially a unified strategy that could potentially be integrated into most population-based LLM-driven AAD approaches. This encouragingly highlights that \emph{fusing benchmark data can enhance the efficiency and robustness of LLM-driven BBO approaches.} }

\section{Conclusions}
In this paper, we addressed the challenges of \emph{understanding the impact of prompt design} and \emph{controlling the search process in LLM-driven optimization.} We conducted the first relevance analysis of prompt components in code generation of LLM-driven optimization approaches by using AttnLRP, confirming that the code-related content obtains the strongest influence. Motivated by this observation, we propose the benchmark-assisted guided evolutionary approach (BAG). By combining prior benchmark knowledge with a simple yet effective (1+1) elitist search strategy, BAG provides a simple way of guiding LLMs towards promising search regions of algorithm code.

Extensive experiments on two BBO suites demonstrated the \textcolor{black}{stronger}
performance of BAG, compared to \textcolor{black}{six} state-of-the-art LLM-driven optimization approaches, EoH, \textcolor{black}{LHNS}, LLaMEA, \textcolor{black}{MCTS-AHD}, ReEvo, and~\textcolor{black}{ShinkaEvolve}. These results were largely consistent across \textcolor{black}{235 training and 940 testing} problem instances using three LLMs.

These findings confirm that benchmark knowledge can effectively guide LLM-driven optimization, offering both practical improvements and a novel perspective on prompt design in LLM-driven black-box optimization. While BAG has achieved competitive performance under the (1+1) elitist scheme, EoH shows advantages in particular scenarios, benefiting from the population-based design. A natural next step is to integrate benchmark knowledge into a population-based framework, which can enhance both the performance and robustness of LLM-driven approaches by enabling more diverse and efficient search dynamics. \textcolor{black}{Such frameworks also use richer prompts than the (1+1) setting
studied here, including reflection text, tool outputs, and longer evolutionary history. Extending our token-wise analysis to these components would reveal whether example code remains the dominant influence on the generated algorithms.} \textcolor{black}{Furthermore, this study highlights the potential to connect classical automated algorithm generation and LLM-driven black-box optimization through the integration of benchmark knowledge.}

\clearpage

\appendices

\section{Benchmark Settings}
\label{appendix:bench}
\subsection{Black-box Optimization}
\label{appendix:bbo}
We first define the general task of (single-objective) Black-box Optimization (BBO). Let $f : \mathcal{X} \to \mathbb{R}$ be an objective function defined over a search space 
$\mathcal{X} \subseteq \mathbb{R}^n$, where $n$ denotes the dimensionality. In BBO, the explicit form, derivatives, or structural properties of $f$ are unknown. The optimization algorithm can only query $f$ through evaluations of the type
\[
    y = f(x), \quad x \in \mathcal{X},
\]
where $y$ is commonly mentioned as objective value or fitness value. The optimizer seeks to find
\[
    x^\star \in \arg\min_{x \in \mathcal{X}} f(x)
    \quad \text{or} \quad
    x^\star \in \arg\max_{x \in \mathcal{X}} f(x).
\]
A \emph{black-box optimization algorithm} is thus an iterative procedure that generates 
a sequence $\{x_t\}_{t=1}^T \subseteq \mathcal{X}$ based solely on past queries 
$\{(x_i, f(x_i))\}_{i=1}^{t-1}$, without requiring analytical knowledge of $f$.

\subsection{The Benchmark Suites}
We provide in this section the detailed problem list of the pbo and bbob suites, including the corresponding target sets used to calculate AUC values~\cite{ye2022automated}.

\noindent \emph{Definition. AUC: area under the ECDF running time curve} Given a predefined target set $\Phi = \{\phi_i \in \R \mid i \in [m]\}$ and a budget set (e.g., function evaluations in our paper) $T=\{t_j \in [B] \mid j \in [z]\}$, the AUC~$\in [0,1]$ (normalized over $B$) of algorithm $A$ on problem $P$ is the (approximate) area under the ECDF curve of the running time over multiple targets. For maximization, it is defined by
\begin{equation*}
\text{AUC}(A,P,\Phi,T) = \frac{\sum\limits_{h=1}^{r}\sum\limits_{i=1}^{m}\sum\limits_{j=1}^{z} \mathds{1} \{\phi_h(A,P,t_j) \ge \phi_i\}} {r m z} \;,
\label{def:AUC}
\end{equation*}
where $r$ is the number of independent runs of $A$ and $\phi_h(A,P,t)$ denotes the value of the best-so-far solution that $A$ obtained within its first $t$ evaluations of the run $h$.

The pbo suite covers a wide range of discrete problems, including the theory-oriented OneMax, LeadingOnes, and their variants, as well as practical problems. The pbo problems list is as follows:
\begin{itemize}
    \item \textbf{F1}: OneMax maximizes the number of one-bits in a bitstring. For the 100-dimensional problem, we use the target set $\{50,\ldots,100\}$ to calculate AUC in this paper. 
    \item \textbf{F2}: LeadingOnes maximizes the number of consecutive one-bits from the start of a bitstring until the first zero-bit. We use the target set $\{0,\ldots,100\}$.
    \item \textbf{F3}: Harmonic maximizes the weighted sum of one-bits, where the weight of each bit is $i, i\in \{1,\ldots,n\}$. The target set is $\{2525 + 5i\} \mid i \in \{0,\dots,505\}$. 
    \item \textbf{F4-F10}: OneMax variants with dummy, neutrality, and epistasis transformations. The target sets are $\{25, \ldots, 50\}$,  $\{45, \ldots, 90\}$,  $\{11, \ldots, 33\}$,  $\{50, \ldots, 100\}$, $\{20, \ldots, 51\}$, $\{0, \ldots, 100\}$, and $\{0, \ldots, 100\}$, respectively.
    \item \textbf{F11-F17}: LeadingOnes variants with dummy, neutrality, and epistasis transformations. The target sets are $\{0, \ldots, 50\}$,  $\{0, \ldots, 90\}$,  $\{0, \ldots, 33\}$,  $\{0, \ldots, 100\}$, $\{0, \ldots, 51\}$, $\{0, \ldots, 100\}$, and $\{0, \ldots, 100\}$, respectively.
    \item \textbf{F18}: LABS (Low Autocorrelation Binary Sequences). The target set is $\{0.5 + 0.1i\} \mid i \in \{0,\dots,450\}$. 
    \item \textbf{F19-F21}: Ising models maximize the energy of a lattice model, considering the one, two, and three-dimensional instances, respectively. The target sets are $\{50, \ldots, 100\}$,  $\{100, \ldots, 200\}$, and $\{150, \ldots, 300\}$, respectively.
    \item \textbf{F22}: MIVS (Maximum Independent Vertex Set). The target set is $\{-1,\ldots,51\}$.
    \item \textbf{F23}: N-Queens, in pbo, the 100-dimensional problem corresponds to 10-Queens. The target set is $\{-2,\ldots,10\}$.
\end{itemize}
We refer to the detailed definitions of problems to~\citet{doerr2020benchmarking}. The target sets are determined based on the corresponding benchmark data. 
To prevent confusion in our discussion, we provide the definition of OneMax (F1) here.
$$
f_{\text{OneMax}}: \{0,1\}^n \rightarrow \{0,\ldots,n\}, x \mapsto \sum_{i=1}^n x_i
$$

The bbob suite consists of five categories of continuous problems. When calculating the ECDFs of algorithms, we commonly consider the objective domain $[10^{-8}, 100]$ and compute the \emph{fraction} (see Section~\ref{sec:bench}) by the distance to $10^{-8}$, following the common setup of using bbob.
The list of bbob problems is as follows:
\begin{itemize}
    \item \textbf{F1-F5}: Separable functions including: Sphere, Separable Ellipsoidal, Rastrigin, B\"uche-Rastrigin, and Linear Slope
    \item \textbf{F6-F9}: Functions with low or moderate conditioning: Attractive Sector, Step Ellipsoidal, original Rosenbrock, and rotated Rosenbrock.
    \item \textbf{F10-F14}: Functions with high conditioning and unimodal: Ellipsoidal, Discus, Bent Cigar, Sharp Ridge, and Different Powers.
    \item \textbf{F15-F19}: Multi-modal functions with adequate global structure: Rastrigin, Weierstrass, Schaffer's F7, ill-conditioned Schaffer's F7, and Composite Griewank-Rosenbrock.
    \item \textbf{F20-F24}: Multi-modal functions with weak global structure: Schwefel, 	Gallagher's Gaussian 101-me Peaks Function, Gallagher's Gaussian 21-hi Peaks Function, Katsuura, and Lunacek bi-Rastrigin.
\end{itemize}

For detailed definitions of the problems, we refer to~\citet{hansen2010comparing}. We provide the definition of Sphere (F1) for discussion in the paper. 
$$
f_{\text{Sphere}}: \mathcal{R}^n \rightarrow \mathcal{R}, \mathbf{x} \mapsto \parallel z\parallel^2 + f_{\text{opt}},
$$
where $ \mathbf{z} = \mathbf{x} - \mathbf{x}_{\text{opt}}$ and $\mathcal{R} \in [-5,5]$. 

\newpage
\section{Prompt Design}
\label{appendix:prompt}
Prompt design is critical to all LLM-driven optimization approaches. In this section, we first describe the general structure of the prompts used in all our experiments, and then we further discuss the specific parts that are customized for each of the three experiments (Section~\ref{sec:token importance analysis}-\ref{sec:bag experiment}). The general template is provided as follows:

\begin{promptbox}{The Prompt Template}

\begin{subsectionbox}{black}
\textbf{$\blacksquare$ Role}

$\vartriangleright$ You are an excellent Python programmer. You are a Python expert working on a new optimization algorithm.
\end{subsectionbox}

\begin{subsectionbox}{black}
\textbf{$\blacksquare$ Problem and Task Descriptions}

\textit{$/*$ Provide here, the descriptions of the black-box optimization problems, along with the specific task assigned to the LLM. $*/$}

\end{subsectionbox}

\begin{subsectionbox}{magenta}
\textbf{$\square$ Linguistic Descriptions of the Parent Heuristic}

\textit{$/*$ Include the descriptions (also the fitness value) of the generated parent heuristic here. This component does not appear in the initial prompt. It is partially produced by the LLM during the optimization loop and organized according to a template, and then used together with the reference code section that follows.$*/$}
\end{subsectionbox}

\begin{subsectionbox}{black}
\textbf{$\blacksquare$ Reference Code}

\textit{$/*$ Place the reference code here. The code should serve as a feasible solver for the black-box problem, meaning it is free of errors at a minimum.$*/$}
\end{subsectionbox}

\begin{subsectionbox}{orange}
\textbf{$\square$ Strategy}

\textit{$/*$ This component defines the instructions given to the LLM for deriving a new solution from the provided parent heuristic.$*/$}
\end{subsectionbox}

\begin{subsectionbox}{Brown}
\textbf{$\blacksquare$ Expected Outputs}\\
$\vartriangleright$ 
Provide the Python code and also give it a one-line description, describing the main idea. Give the response in the format:\\
\# Description: $<$short-description$>$\\
\# Code:\\
\textasciigrave\textasciigrave\textasciigrave python\\
$<$code$>$\\
\textasciigrave\textasciigrave\textasciigrave

\end{subsectionbox}
\end{promptbox}

Our prompts implement the concept of the (1+1) elitist search strategy. 
All initial prompts consist of four components, each marked by $\blacksquare$: \emph{Task Description}, including a Role instruction as well as the Problem and Task Descriptions, the Reference Code, and the instructions specifying the \emph{Expected Outputs}. During the optimization loop, the prompt includes two additional components, each marked with $\square$: \emph{Parent Heuristic(s)}, including the linguistic description (together with its fitness value \emph{Note}) and updated Reference code, as well as the \emph{Strategy} sepcifying the instruction to the LLM for deriving a new solution based on the parent. The \textit{italicized} text enclosed within $/*$ and $*/$ provides auxiliary explanations, while \textbf{the regular text marked with $\vartriangleright$ is the actual prompt}. 

With the overall prompt template, we first introduce the template of the linguistic descriptions of the parent heuristic, a shared component across all experiments:

\begin{promptbox}{The Template for Descriptions of the Parent}
\begin{subsectionbox}{magenta}
\textbf{$\square$ Linguistic Descriptions of the Parent Heuristic}

$\vartriangleright$ The current population of algorithms already evaluated (name, description, score) is:

\textit{$/*$ Names and Descriptions of the parent given by the LLM$*/$} (Score: \textit{$/*$ the fitness value$*/$})

The selected solution to update is:

\textit{$/*$ Descriptions of the parent given by the LLM$*/$}

With code:

\textit{$/*$ Immediately followed by the component Reference Code.$*/$}
\end{subsectionbox}
\end{promptbox}

Next, we introduce the two task-specific description components for the pbo and bbob suites. The task description for these two suites shares much common content, \textbf{with the main differences (highlighted in bold)} lying in the problem characteristics (pseudo-Boolean vs. continuous), aligning with the variable types (Boolean variables vs. bounded continuous space). \textbf{Note that the suite and problem names are omitted from the prompt to ensure proper alignment with the properties of BBO.} Details of the prompts are given as follow:

\begin{promptbox}{Task Prompt for Pbo Benchmark Suite - Part 1}

\begin{subsectionbox}{black}
$\blacksquare$ \textbf{Problem and Task Descriptions}

$\vartriangleright$ Your task is to develop a novel heuristic optimization algorithm for \textbf{pseudo-boolean maximization problems}.
The optimization algorithm should work on different instances of noiseless unconstrained functions. Your task is to write the optimization algorithm in Python code.
Each optimization problem has a search space \textbf{consisting of binary decision variables that can take values 0 or 1}. The dimensionality can be varied.

The code should contain an \textasciigrave \_\_init\_\_(self, budget, dim, seed)\textasciigrave\ function with optional additional arguments and the function \textasciigrave\ def \_\_call\_\_(self, func)\textasciigrave, which should optimize the black box function \textasciigrave func\textasciigrave\ using \textasciigrave self.budget\textasciigrave\ function evaluations.
The func() can only be called as many times as the budget allows, not more. 
\end{subsectionbox}
\end{promptbox}
\newpage

\begin{promptbox}{Task Prompt for Bbob Suite - Part 1}

\begin{subsectionbox}{black}
$\blacksquare$ \textbf{Problem and Task Descriptions}

$\vartriangleright$ Your task is to write a heuristic optimization algorithm for \textbf{continuous problems}. The optimization algorithm should work on different instances of noiseless unconstrained functions. Your task is to write an optimization algorithm in Python.
Each of the optimization functions has a \textbf{search space between -5.0 (lower bound) and 5.0 (upper bound)}. The dimensionality can be varied.

The code should contain an \textasciigrave \_\_init\_\_(self, budget, dim, seed)\textasciigrave\ function with optional additional arguments and the function \textasciigrave\ def \_\_call\_\_(self, func)\textasciigrave, which should optimize the black box function \textasciigrave func\textasciigrave\ using \textasciigrave self.budget\textasciigrave\ function evaluations.
The func() can only be called as many times as the budget allows, not more. 
\end{subsectionbox}
\end{promptbox}

\subsection{Initial Prompts for Pbo and Bbob Suites}
The initial reference code for all LLM-driven optimizers except for BAG is given as follow:

\begin{promptbox}{Prompt Components for Pbo Suite - Part 2}
\begin{subsectionbox}{black}
$\blacksquare$ \textbf{Reference Code for the Initial Prompt}

$\vartriangleright$ An example of such code (a simple random search), is as follows:

\textasciigrave\textasciigrave\textasciigrave python

\begin{lstlisting}[style=pythonstyle, numbers=none, frame=none]
import numpy as np
class RandomSearch:
    def __init__(self, budget, dim, seed=None):
        self.budget = budget
        self.dim = dim
        self.rng = np.random.default_rng(seed)

    def __call__(self, func):
        f_opt = -np.Inf
        x_opt = None
        while func.state.evaluations < self.budget:
            x = self.rng.integers(0, 2, size=self.dim)
            f = func(x)
            if f > f_opt:
                f_opt = f
                x_opt = x
            if f_opt >= func.optimum.y:
                break
        return f_opt, x_opt
\end{lstlisting}
\textasciigrave\textasciigrave\textasciigrave
\end{subsectionbox}
\end{promptbox}

\newpage

\begin{promptbox}{Prompt Components for Bbob Suite - Part 2}
\begin{subsectionbox}{black}
$\blacksquare$ \textbf{Reference Code for Initial Prompt}

$\vartriangleright$ An example of such code (a simple random search), is as follows:

\textasciigrave\textasciigrave\textasciigrave python

\begin{lstlisting}[style=pythonstyle, numbers=none, frame=none]
import numpy as np
class RandomSearch:
    def __init__(self, budget, dim, seed=None):
        self.budget = budget
        self.dim = dim
        self.rng = np.random.default_rng(seed)

    def __call__(self, func):
        f_opt = np.Inf
        x_opt = None
        for i in range(self.budget):
            x = self.rng.uniform(func.bounds.lb, \
                                    func.bounds.ub)
            f = func(x)
            if f < f_opt:
                f_opt = f
                x_opt = x
            if f_opt <= func.optimum.y:
                break
        return f_opt, x_opt
\end{lstlisting}
\textasciigrave\textasciigrave\textasciigrave
\end{subsectionbox}
\end{promptbox}

By default, we adopt (global) random search as the code-formatting guideline for EoH, \textcolor{black}{LHNS}, LLaMEA, \textcolor{black}{MCTS-AHD}, and ReEvo, due to its simplicity. 

\subsection{Common Strategies}
Below we list the two common strategies that are used in all three experiments for BAG (Sections~\ref{sec:token importance analysis},~\ref{sec:refine experiments}, and~\ref{sec:bag experiment}). This is to say, besides the initial and refinement iterations, BAG dynamically chooses one of two strategies to be the strategy for the offspring (see lines 11 to 14 of Algorithm~\ref{alg:LLM-bbo}).

\begin{promptbox}{Prompt Components - Strategy}

\begin{subsectionbox}{orange}
\textbf{$\square$ Strategy 1}

$\vartriangleright$ Refine the strategy of the selected solution to improve it.
\end{subsectionbox}

\begin{subsectionbox}{orange}
\textbf{$\square$ Strategy 2}

$\vartriangleright$ Generate a new algorithm that is different from the algorithms you have tried before.
\end{subsectionbox}

\end{promptbox}

\subsection{Prompts Design for the Refinement-only Optimization}
In this section, we present the prompts used in the experiments to guide the LLM toward specific search regions (see Section~\ref{sec:refine experiments}) and also in the final BAG optimizer (see section~\ref{sec:bag experiment}). In this setting, each initial prompt and each subsequent prompt that refers to the $\mathbf{A}_{bench}$ contains a promising starting code, and the LLM is required to refine this code as specified in the strategy block. An example of a reference code block and the corresponding strategy block for a pbo problem is shown below:

\begin{promptbox}{Prompt Components for the Refinement Experiment}
\begin{subsectionbox}{black}
$\blacksquare$ \textbf{Reference Code}
\textit{$/*$ Here, the greedy hill climber is used as the example.$*/$}

$\vartriangleright$ An example of such code for a good optimization algorithm is as follows:

\textasciigrave\textasciigrave\textasciigrave python

\begin{lstlisting}[style=pythonstyle, numbers=none, frame=none]
import numpy as np
class GreedyHillClimber:
    def __init__(self, budget, dim, seed=None):
        self.budget = budget
        self.dim = dim
        self.rng = np.random.default_rng(seed)

    def __call__(self, func):
        self.mutation_rate = (1.0 / self.dim)
        x = self.rng.integers(0, 2, self.dim)
        fx = func(x)
        idx = 0
        while func.state.evaluations < self.budget:
            y = x.copy()
            y[idx] = 1 - y[idx]  # flip bits for 0/1 domain
            idx = (idx + 1) % self.dim
            fy = func(y)
            if fy > fx:
                x, fx = y, fy
            if fy >= func.optimum.y:
                break
        return fx, x
\end{lstlisting}
\textasciigrave\textasciigrave\textasciigrave
\end{subsectionbox}

\begin{subsectionbox}{orange}
\textbf{$\square$ Strategy}

$\vartriangleright$ Refine the example algorithm to improve its performance on the given task. Focus only on algorithmic changes, not formatting or comments.
\end{subsectionbox}

\end{promptbox}


\section{\textcolor{black}{Additional Information on Experimental Setup}}
\label{sec:app:setup}
\paragraph{Details on the LLMs}
We conduct the token-wise explanation experiments on \textcolor{black}{six open-source LLMs from two model families, Gemma~3 (4B, 12B and 27B) and Qwen3 (8B, 14B and 32B),} and evaluate all LLM-driven optimization methods on three proprietary LLMs, as summarized in Table~\ref{tab:LLMs}. All models are used with their default (sampling) parameters provided by the respective versions. The open-source models are loaded using the transformers library provided by Hugging Face~\citep{wolf2020transformers}. \textcolor{black}{We select these two families for two reasons. AttnLRP requires access to model weights, which the proprietary models do not provide. In addition, they are open-source counterparts of two of the proprietary models: Gemma~3 is built from the same research and technology as Gemini, and Qwen3 Coder Flash belongs to the Qwen3 family.}

\begin{table}[ht]
\centering
\caption{\textcolor{black}{The used LLMs}}
\label{tab:LLMs}
\resizebox{0.48\textwidth}{!}{
\begin{tabular}{l|c|c}
\toprule
\multicolumn{3}{c}{Open-source models used in Section~\ref{sec:token importance analysis}. Models acquired from Hugging Face.} \\
\midrule
\textbf{Name} & \textbf{Version} & \textbf{Provider}\\
\midrule
\multirow{3}{*}{Gemma 3}
  & google/gemma-3-4b-it
  & \multirow{3}{*}{Google DeepMind} \\
  & google/gemma-3-12b-it
  & \\
  & google/gemma-3-27b-it
  & \\
  \midrule
\multirow{3}{*}{Qwen 3}
  & Qwen/Qwen3-8b
  & \multirow{3}{*}{Alibaba Cloud} \\
  & Qwen/Qwen3-14b
  & \\
  & Qwen/Qwen3-32b
  & \\

\midrule
\multicolumn{3}{c}{Proprietary models used in Sections~\ref{sec:refine experiments} and~\ref{sec:bag experiment}} \\
\midrule
\textbf{Name} & \textbf{Version} & \textbf{Provider}\\
\midrule
\textcolor{black}{Gemini 2.5 Flash-Lite} & gemini-2.5-flash-lite & Google DeepMind\\
GPT 5 Nano & gpt-5-nano-2025-08-07 & OpenAI\\
Qwen3 Coder Flash & qwen3-coder-flash-2025-07-28 & Alibaba Cloud\\
\bottomrule
\end{tabular}
}
\end{table}

\paragraph{Setups of pbo and bbob suites}
For each problem (function) in pbo and bbob (see Appendix~\ref{appendix:bench}), we evaluate every generated algorithm on five instances of the problem to compute the mean AUC performance. To ensure fairness in CPU time, we set an evaluation timeout for each instance. Table~\ref{tab:app:setup pbo and bbob} summarizes the setup of the pbo and bbob problem suites.

\begin{table}[ht]
    \centering
    \caption{Setups of the pbo and bbob suites}
    \label{tab:app:setup pbo and bbob}
    \resizebox{0.48\textwidth}{!}{
    \begin{tabular}{c|c|c|c}
    \toprule
    Suite & Dimensionality & Budget & Evaluation timeout per instance\\
    \midrule
     pbo &  100 &  1,000,000 & 600 seconds\\ 
     \midrule
     bbob & 5 & 10,000 & 600 seconds\\
     \bottomrule
    \end{tabular}
    }
\end{table}

\paragraph{\textcolor{black}{Example algorithms in $\mathbf{A}_{bench}$}}
\textcolor{black}{For pbo, the five algorithms of each problem use the parameter settings reported in~\cite{doerr2020benchmarking}. For bbob, all five algorithms use the default settings of their original implementations. Constructing $\mathbf{A}_{bench}$ required no parameter tuning and no additional optimization runs, beyond adapting each implementation to the common interface described in Appendix~\ref{appendix:prompt}.}

\paragraph{Setups of software}
We use the default implementations and hyperparameters for EoH\footnote{\url{https://github.com/FeiLiu36/EoH}} (see Section 4.1 of~\citet{liuevolution}), LHNS\footnote{https://github.com/Acquent0/LHNS} (see the official repository), LLaMEA\footnote{\url{https://github.com/XAI-liacs/LLaMEA}} (see Section IV of~\citet{van2024llamea}), MCTS-AHD\footnote{https://github.com/zz1358m/MCTS-AHD-master} (See Section 4 of~\citet{zheng2025monte}), and ReEvo\footnote{\url{https://github.com/ai4co/reevo}} (see Appendix C of~\citet{ye2024reevo}). \textcolor{black}{ShinkaEvolve~\citet{lange2026shinkaevolve} is a state-of-the-art framework for general scientific discovery\footnote{https://github.com/SakanaAI/shinkaevolve}, it is a complex system that goes beyond the LLM-based automated algorithm design that the other methods falls in. Therefore, to ensure a fair comparison with the other methods, we adapted ShinkaEvolve into a controlled evolutionary code-generation baseline with three different settings and report its best-performed variant in the main results (see the further discussion in Appendix~\ref{sec:app:shinka}). Candidate optimizers were evaluated using the same AUC metric and benchmark evaluation budget as the other methods.} The internal prompt engineering of all six baseline models is kept unchanged.

For fairness and consistency, we provide only the \textit{Task Description} along with the initial random search code to each method. We use the default implementation of AttnLRP\footnote{\url{https://github.com/rachtibat/LRP-eXplains-Transformers}} for the experiment on token-wise analysis of prompt design.

\paragraph{Hardware Specifications}
The token-wise analysis experiment is conducted on clusters equipped with \textcolor{black}{NVIDIA L40S GPUs and Intel Xeon Gold 6438M CPUs.}
For large-scale benchmarking on the pbo and bbob problem suites, each problem (function) is assigned a single core of an AMD EPYC 7662 CPU for every LLM-driven optimizer. Importantly, as discussed in Section~\ref{sec:bag experiment}, the final experimental results are independent of the underlying hardware.

\section{\textcolor{black}{Impact of the Frequency Factor}}
\label{appendix:q}
\begin{figure}[ht]
    \centering
    \includegraphics[width=0.9\linewidth]{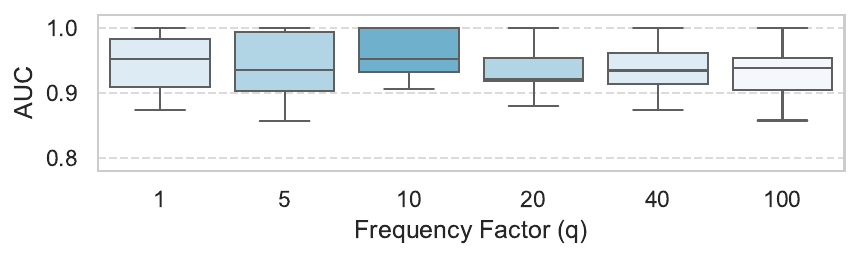}
    \caption{\textcolor{black}{Average performance of BAG with different frequency factor $q$. Results are aggregated across three LLMs and seven pbo benchmarks.}}
    \label{fig:ablation frequency factor pbo}
\end{figure}

\begin{figure}[ht]
    \centering
    \includegraphics[width=0.9\linewidth]{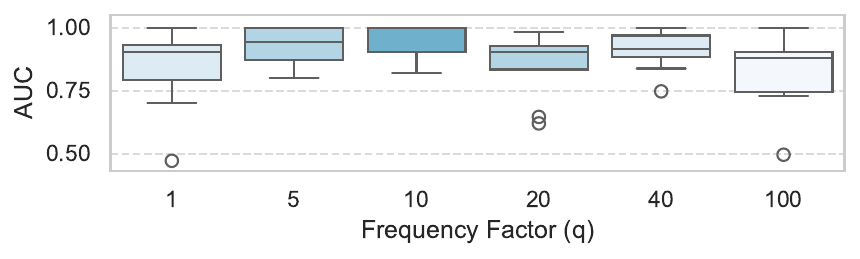}
    \caption{\textcolor{black}{Average performance of BAG with different frequency factor $q$. The results are aggregated across three LLMs and six bbob problems.}}
    \label{fig:ablation frequency factor bbob}
\end{figure}

\textcolor{black}{In this section, we study the impact of the frequency factor $q$ in Algorithm~\ref{alg:LLM-bbo}. Specifically, we compare the AUC values obtained with different $q \in \{1, 5, 10, 20, 40, 100\}$. Figures~\ref{fig:ablation frequency factor pbo} and \ref{fig:ablation frequency factor bbob} present the averaged (normalized) performance of the three LLMs on the pbo and bbob benchmarks, respectively. These results show that BAG achieves the best and most robust results when $q=10$. Under this setting, BAG is expected to utilize each algorithm in $\mathbf{A}_{bench}$ (of size 5) approximately twice on average. }

\section{\textcolor{black}{Impact of the Quality of the Example Algorithms}}
\label{appendix:k}
\begin{figure}[ht]
    \centering
    \includegraphics[width=0.9\linewidth]{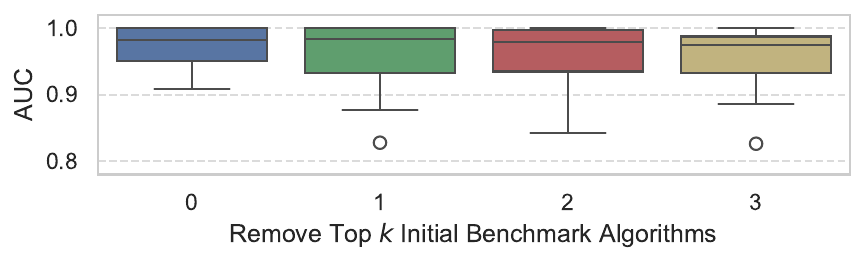}
    \caption{\textcolor{black}{Average performance of BAG with each of the top-$k$ example benchmark algorithms ($\mathbf{A}_{bench}$) \underline{removed}. The results are aggregated across three LLMs and seven pbo problems. The leftmost box-plot ($k$=0) corresponds to the default setting, where no benchmark algorithms are removed.}}
    \label{fig:ablation top-K pbo}
\end{figure}

\begin{figure}[ht]
    \centering
    \includegraphics[width=0.9\linewidth]{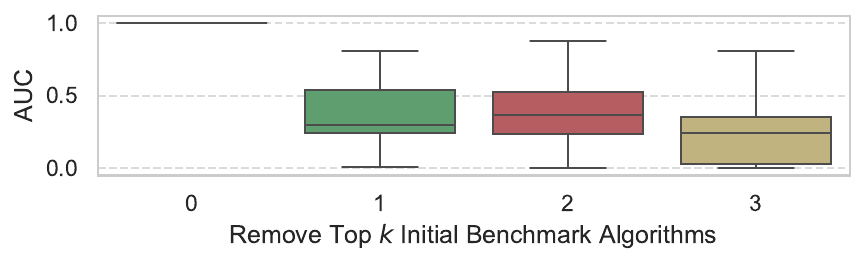}
    \caption{\textcolor{black}{Average performance of BAG with each of the top-$k$ example benchmark algorithms ($\mathbf{A}_{bench}$) \underline{removed}. The results are aggregated across three LLMs and six bbob problems. The leftmost box-plot ($k$=0) corresponds to the default setting, where no benchmark algorithms are removed.}}
    \label{fig:ablation top-K bbob}
\end{figure}

\textcolor{black}{In this section, we investigate how the quality of the example benchmark algorithms in Algorithm~\ref{alg:LLM-bbo} affect BAG. For each problem in the two benchmark suites, we rank the five example benchmark algorithms in $\mathbf{A}_{bench}$ according to their individual AUC (see section~\ref{appendix:bench}).} \textcolor{black}{This ranking is used only in this analysis, the proposed BAG itself does not use it.}
\textcolor{black}{We then remove the top-$k$ algorithms, namely the $k$ best-performing ones, for $k\in \{1,2,3\}$, and compare the resulting AUC against the default setting ($k=0$), in which all five are kept. Removing the \textit{strongest} algorithms, rather than random or weak ones, is the clearest way to test whether BAG depends on the quality of its example set. If performance does not change after removing the best seeds, then the composition of $\mathbf{A}_{bench}$ does not matter much.}
\textcolor{black}{Figures~\ref{fig:ablation top-K pbo} and \ref{fig:ablation top-K bbob} show boxplots of the best obtained AUC across the three LLMs on the pbo and bbob suites, respectively. In both suites, BAG achieves its highest average performance and greatest robustness when $k=0$, although the magnitude of this effect depends strongly on the benchmark suite. On pbo (Fig.~\ref{fig:ablation top-K pbo}), BAG degrades gracefully: the median AUC remains high through $k=3$, while the worst-case performance declines steadily as more top algorithms are removed. On BBOB (Fig.\ref{fig:ablation top-K bbob}), however, BAG shows a much stronger dependence on the example set. Removing even the single best example algorithm, i.e., setting $k=1$, reduces the median normalized performance from $1.0$ to roughly $0.4$ and introduces large variance. Together, these results confirm that the quality of the example algorithms matters, and that on bbob suite, it becomes decisive.}

\textcolor{black}{These findings should be considered together with the frequency factor $q$ (see Appendix~\ref{appendix:q}). The two design choices jointly affect whether each example algorithm contributes to the search. Specifically, $q$ controls how often each algorithm in $\mathbf{A}_{bench}$ is selected as the basis for LLM refinement under a fixed proposal budget, while $k$ controls how many of the strong algorithms remain in the example set.
A reasonable configuration, with $q$ between $5$ and $10$ and the full example set retained ($k=0$), allows each of the five example algorithms to be used as a basis at least once on average. This helps BAG benefit from the full quality of $\mathbf{A}_{bench}$. The BBOB results highlight why this coverage matters. When each algorithm has a fair chance to guide the search, removing the single most effective algorithm has an immediate effect, leading to the sharp performance drop observed at $k=1$.}

\section{Analysis of Impact of Benchmark(-Induced) Algorithm to Generated Heuristics}
\label{appendix:CodeBLEU}

To analyze the impact of benchmark algorithm on the generated heuristics, we use CodeBLEU~\cite{ren2020codebleu}, a metric for measuring the similarity (relevance) between two pieces of code. CodeBLEU was originally proposed to evaluate the logical and structural quality of generated code by comparing it with a reference implementation. Given a reference code snippet (R) and a generated code snippet (S), CodeBLEU computes a weighted sum of four sub-metrics derived from R and S. These include: 1) BLEU, which measures exact sequential token matches between R and S; 2) weighted n-gram matching of BLEU, which emphasizes the influence of key tokens such as data types and keywords; 3) abstract syntax tree (AST) matching, which captures structural consistency across code lines; and 4) data-flow matching, which evaluates semantic similarity based on variable dependencies, such as whether later code correctly references earlier definitions of variable.

In our setting, CodeBLEU\footnote{\textcolor{black}{We use the default setup of \url{https://github.com/k4black/codebleu}.}} can provide a quantitative measure of how much textual, structural, and semantic information the generated heuristics inherit from earlier heuristics. This allows us to assess how the injected benchmark algorithm influences solutions produced at later stages of the evolutionary search process.

In our setup, new algorithms are generated in a (1+1) evolutionary manner, resulting in an ordered sequence with one-directional influence: later algorithms inherit information from their promising predecessors, but not the other way around. To check this property, for each problem and each LLM, we compute pairwise similarity scores for all ordered pairs of algorithms within the same search trajectory of BAG using an upper-triangular scheme. Specifically, given a sequence of algorithms [$\mathcal{A}_1$, \dots, $\mathcal{A}_N$], we iterate over indices $i \in [1, N-1]$ and, for each $i$, compute the similarity between $\mathcal{A}_i$ and all subsequent heuristics $\mathcal{A}_j$ where $j \in [i+1, N]$. This procedure evaluates each valid pair exactly once while avoiding redundant and self-comparisons, and preserves the temporal direction of influence inherent in the generation process of BAG.

\section{Addition Comparison Results}
\label{appendix:ref-com}
We present the detailed performance of our BAG method on each individual benchmark problem, compared to EoH, \textcolor{black}{LHNS}, LLaMEA, \textcolor{black}{MCTS-AHD}, ReEvo, and \textcolor{black}{ShinkaEvolve}, in Tables~\ref{tab:results_PBO_gemini-2.5-flash-lite} to~\ref{tab:results_BBOB_qwen3-coder-flash}. The results include the normalized AUC values, the corresponding ranks based on AUC, and the convergence process of the obtained AUC throughout the search process of each method.

\subsection{\textcolor{black}{Remarks on ShinkaEvolve Configurations}}
\label{sec:app:shinka}
\textcolor{black}{As described in Appendix~\ref{sec:app:setup}, we implemented and evaluated three variants of ShinkaEvolve to ensure a fair comparison. The configurations of these variants are summarized in Table~\ref{tab:shinka-variants}.}
\textcolor{black}{
The \textit{minimal} variant is designed to closely approximate the LLM utilization of the baseline methods and our BAG method. The \textit{single LLM} variant restricts the system to a single LLM, enabling a fair comparison with previously reported methods, all of which rely on only one LLM. Finally, the \textit{native} variant represents the default configuration.}
\begin{table}[t]
\centering
\caption{\textcolor{black}{Configuration Summary of the Three ShinkaEvolve Variants}}
\label{tab:shinka-variants}
\resizebox{0.8\linewidth}{!}{
\begin{tabular}{l|ccc}
\toprule
Variant & Minimal & Single LLM & Native \\
\midrule
Backend models & 1 & 1 & Multi-model ensemble \\
LLM identity & Single LLM & Single LLM & Gemini/GPT/Qwen \\
\# Islands & 1 & 2 & 2 \\
Bandit model selection & No & Yes, one model & Yes, multi-model \\
Text embeddings & No & \multicolumn{2}{c}{Yes, OpenAI text-embedding-3-small} \\
Novelty LLM judge & No & Yes, same model & Yes, GPT 5 Nano \\
Meta recommendation & No & Yes, same model & Yes, GPT 5 Nano \\
Patch types & Full only & \multicolumn{2}{c}{Yes, Diff/Full/Cross} \\
Patch resampling & Low & Native & Native \\
\bottomrule
\end{tabular}
}
\end{table}

\begin{figure}[ht]
    \centering
    \includegraphics[width=0.95\linewidth]{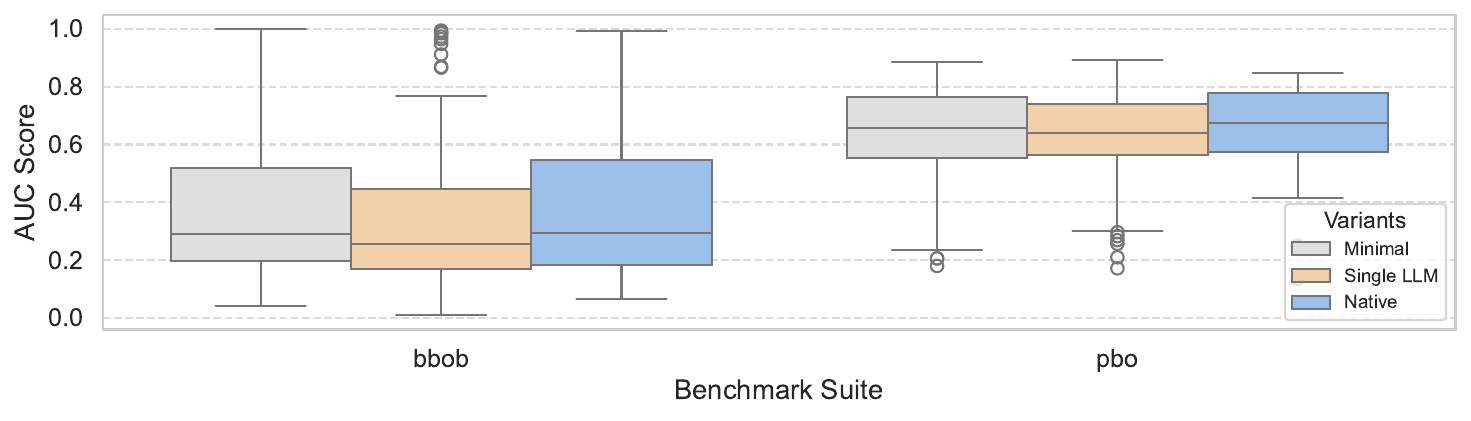}

    \vspace{0.5em}

    \includegraphics[width=0.95\linewidth]{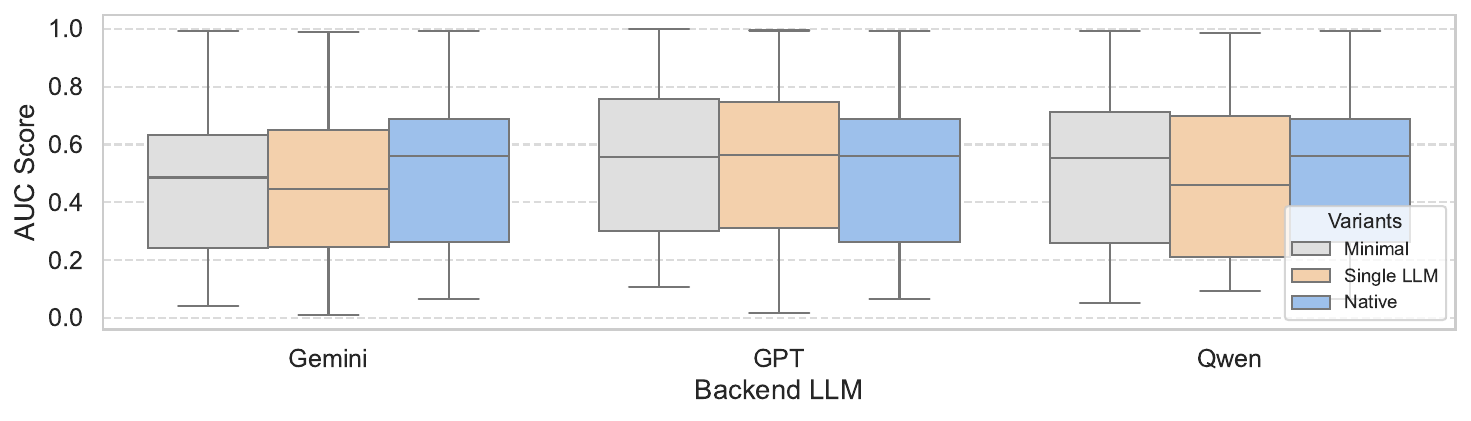}

    \caption{\textcolor{black}{Performance comparison of the three ShinkaEvolve variants by benchmark (top) and by backend LLM (bottom).}}
    \label{fig:results_shinka}
\end{figure}

\textcolor{black}{Figure~\ref{fig:results_shinka} presents the results of the internal comparison among the three ShinkaEvolve variants, while Table~\ref{tab:shinka results} summarizes their overall performance. Although the \textit{minimal} and \textit{native} variants show comparable performance, we selected the \textit{minimal} variant because it achieved the best overall ranking across all problems in the two benchmark suites. This selection is consistent with the evaluation criterion adopted in our other experiments.}

\begin{table}[htbp]
    \centering
    \caption{\textcolor{black}{Overall Performance of ShinkaEvolve Variants}}
    \label{tab:shinka results}
    \resizebox{0.95\linewidth}{!}{
    \begin{tabular}{lrrrrr}
    \toprule
    Variant & Avg. Rank & Median Rank & \# Wins & Mean Raw AUC & Std. Raw AUC\\
    \midrule
    Minimal & 1.947 & 2.000 & 50/141 & 0.497 & 0.256 \\
    Single LLM & 2.092 & 2.000 & 36/141 & 0.483 & 0.264\\
    Native & 1.961 & 2.000 & 54/141 & 0.509 & 0.260 \\
    \bottomrule
    \end{tabular}}
\end{table}

\section{\textcolor{black}{Results for Instance-level Testing}}
\label{appendix:testResult}
\begin{figure*}
    \centering
    \includegraphics[width=\linewidth]{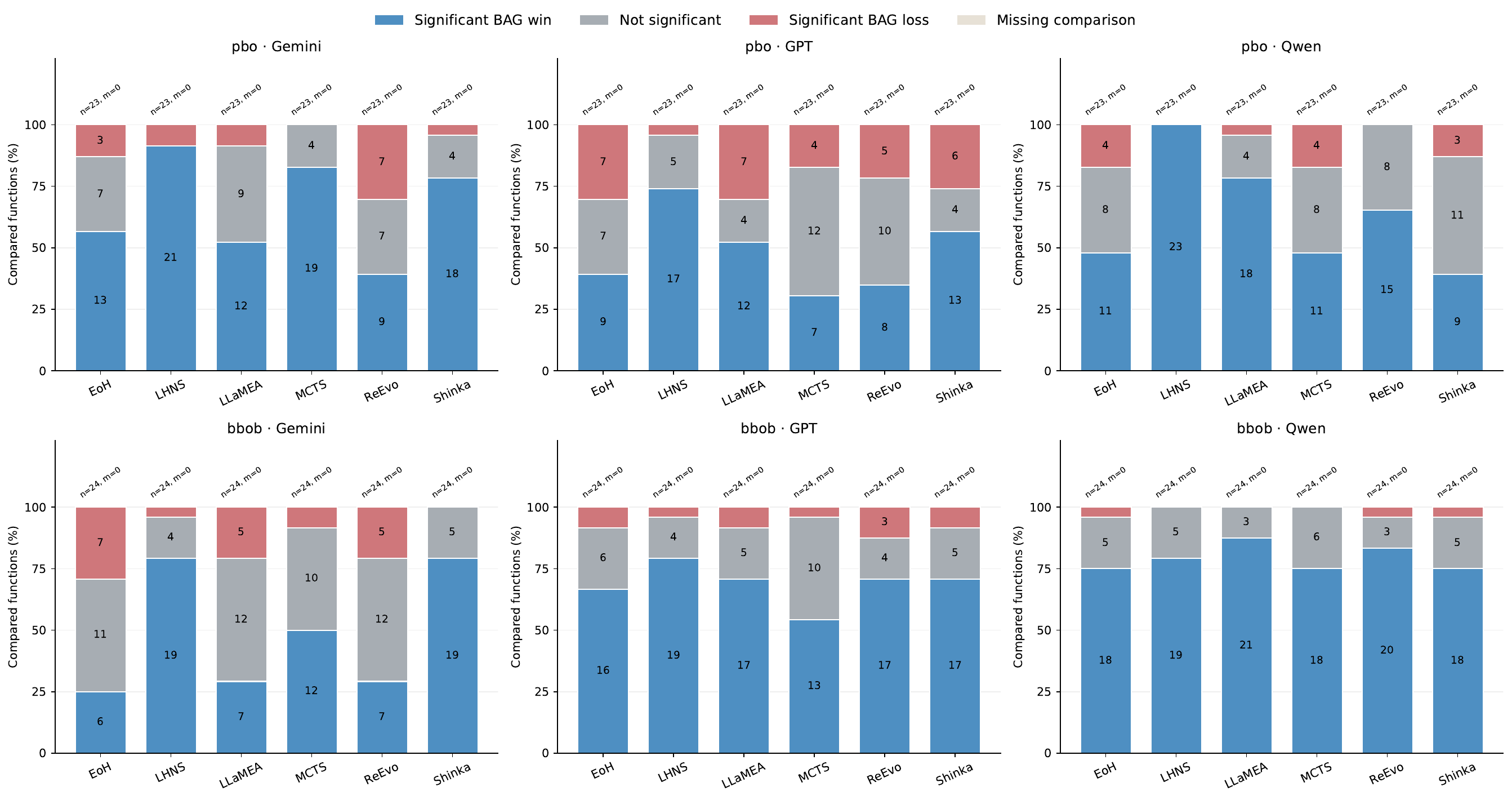}
    \caption{\textcolor{black}{Summary of pairwise function-level comparison between BAG and each of the six baseline LLM-driven automated algorithm design frameworks for both pbo and bbob suites under each backend LLM. Results of statistical significance are determined using two-sided exact Wilcoxon signed-rank tests with Holm correction within each suite–LLM–function combination (\(\alpha=0.05\)). Wins and losses indicate significant differences favoring BAG and the compared baseline framework, respectively, with ties indicating nonsignificant differences.}}
    \label{fig:bag_vs_baseline}
\end{figure*}

\begin{figure*}
    \centering
    \includegraphics[width=\linewidth]{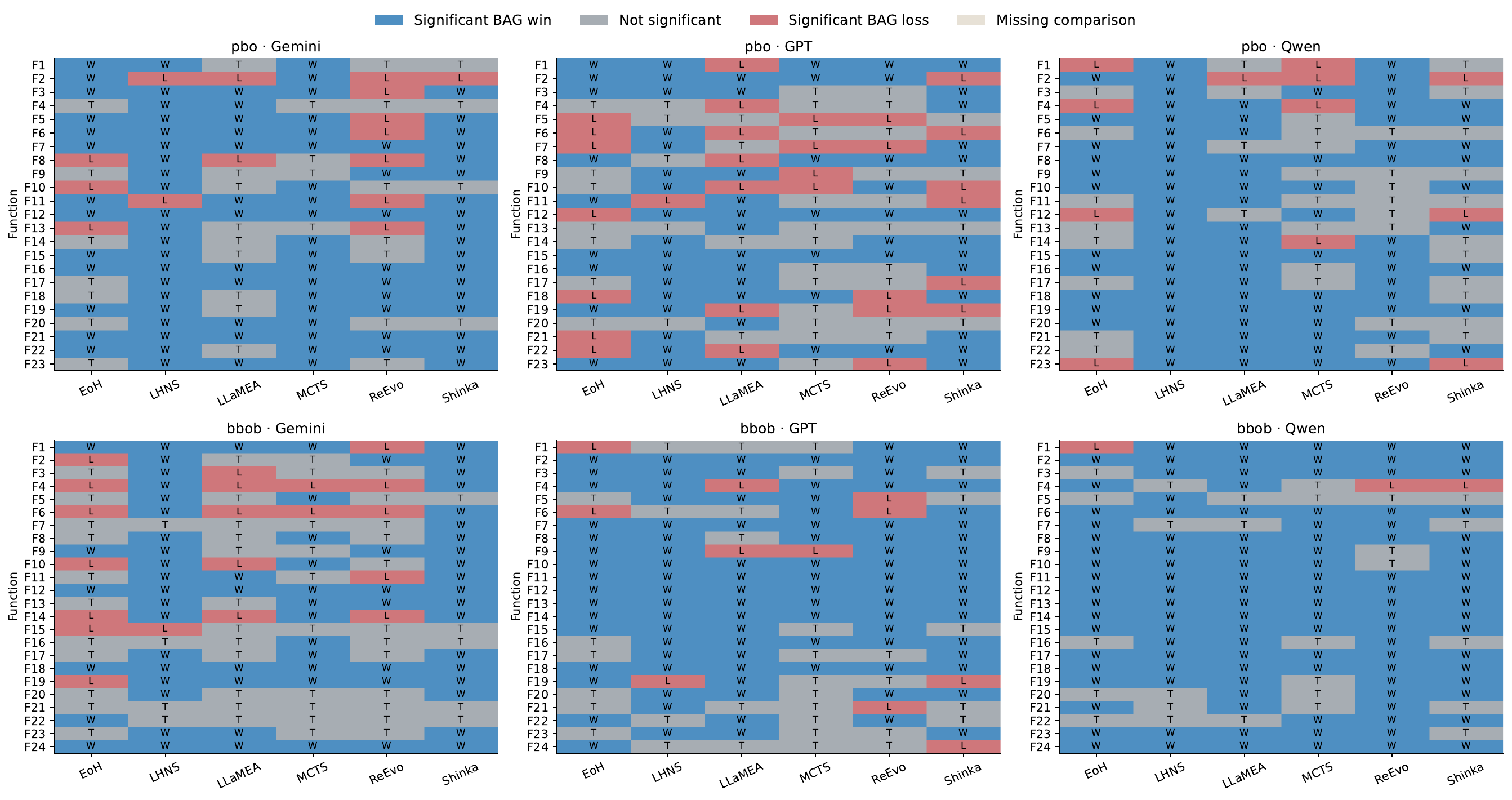}
    \caption{\textcolor{black}{Function-level view of comparison between BAG and each of the six baseline LLM-driven automated algorithm design frameworks for both pbo and bbob suites under each backend LLM. Results of statistical significance are determined using two-sided exact Wilcoxon signed-rank tests with Holm correction within each suite–LLM–function combination (\(\alpha=0.05\)). Wins (W) and losses (L) indicate significant differences favoring BAG and the baseline framework, respectively, with ties (T) indicating nonsignificant differences.}}
    \label{fig:bag_vs_baseline_heatmap}
\end{figure*}

In this section, we evaluate the performance of the final algorithms obtained by LLM-driven optimization methods. Recall that during the searching process, the candidate algorithms are assessed on five \textcolor{black}{training} problem instances. Here, we validate the algorithms' performance on a different set of \textcolor{black}{twenty} instances that are unseen during the search. \textcolor{black}{This is a common practice when benchmarking black-box optimization algorithms~\cite{hansen2022anytime,ye2022automated}. For each benchmark function, we apply a two-sided Wilcoxon signed-rank test to the paired AUC values over the $20$ instances, and adjust the resulting $p$-values with the Holm method within each combination of LLM and benchmark suite ($\alpha = 0.05$). A function counts as a win (loss) for BAG when its AUC value is significantly higher (lower) than that of the baseline algorithm, and as a tie otherwise. 
\\
The corresponding results are shown in Figure~\ref{fig:bag_vs_baseline}. For the pbo suite, BAG consistently achieves more statistically significant wins than losses against all six baselines across the three backend LLMs, demonstrating its favorable performance on unseen test instances. Similar results are observed on the BBOB suite, where BAG achieves more significant wins than losses in 17 out of 18 baseline–LLM comparisons. The only exception is against EoH with Gemini 2.5 Flash-lite, where BAG obtains six significant wins and seven significant losses. In contrast, BAG substantially outperforms EoH with GPT and Qwen, achieving 16 and 18 significant wins against only two and one significant losses, respectively. These results provide evidence that BAG-generated algorithms show stronger instance-level transferability across different backend LLMs.
\\
We additionally show the function-level heatmap of the comparative results in Figure~\ref{fig:bag_vs_baseline_heatmap}. The results show that BAG achieves statistically significant improvements over the baselines across a broad range of benchmark functions, rather than being limited to a few specific cases. This is particularly evident in the comparisons with LHNS, where BAG achieves significant wins on most functions across both benchmark suites and all three backend LLMs. Notably, on the BBOB suite, BAG significantly outperforms all six baselines on functions F12 and F18 across all three backend LLMs, indicating consistent performance advantages on these functions. Nevertheless, we also observe some function-dependent variations, particularly on BBOB with Gemini 2.5 Flash-lite, where BAG shows several significant losses on functions F4 and F6. Overall, these observations further support the effectiveness of BAG while highlighting that its relative performance can vary across benchmark functions and backend LLMs.}


\section{Additional Refinement Results}
\label{appendix:exp-ref}
\begin{figure*}[!htb]
    \centering
\subfloat[\label{fig:refine-f2-lh}]{
    \includegraphics[width=0.945\textwidth]{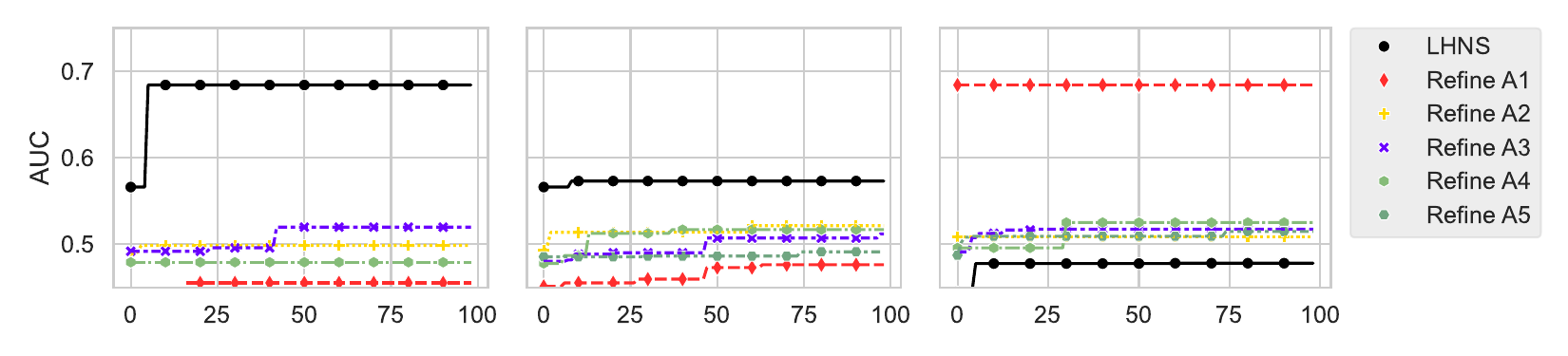}
}

\subfloat[\label{fig:refine-f10-lh-app}]{
    \includegraphics[width=0.945\textwidth]{figures_new/Refine/refine_pid_10_LH.pdf}
}

\subfloat[\label{fig:refine-f17-lh}]{
    \includegraphics[width=0.945\textwidth]{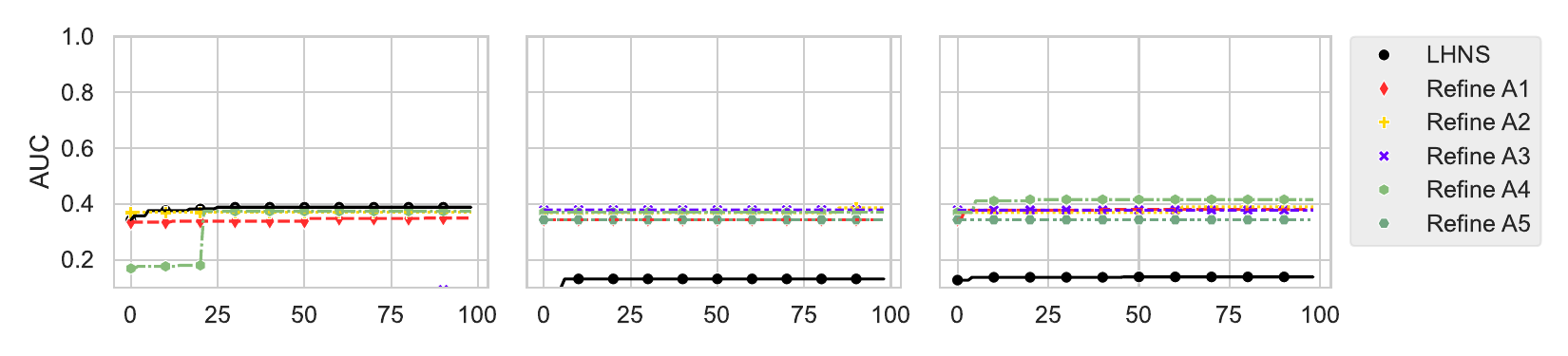}
}

\subfloat[\label{fig:refine-f19-lh}]{
    \includegraphics[width=0.945\textwidth]{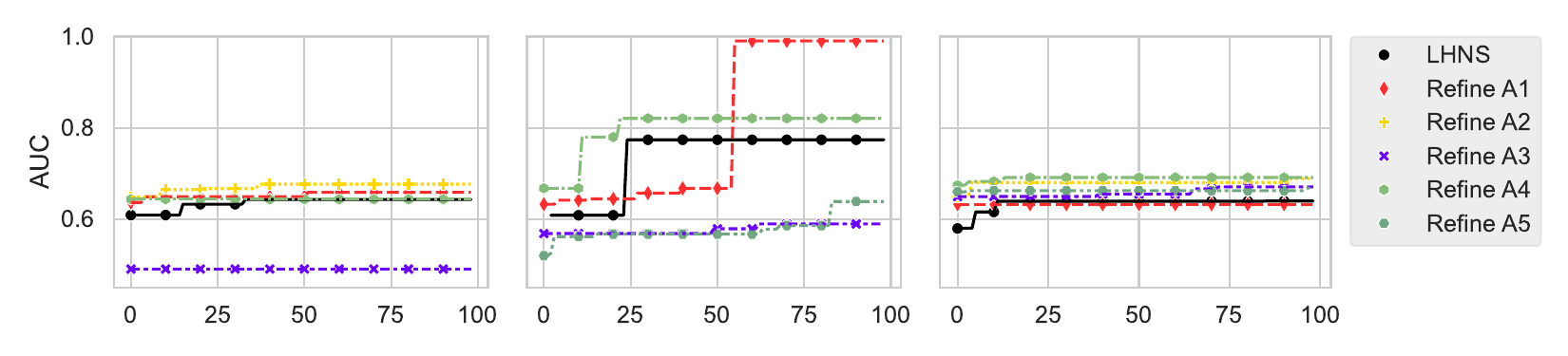}
}

\subfloat[\label{fig:refine-f22-lh}]{
    \includegraphics[width=0.945\textwidth]{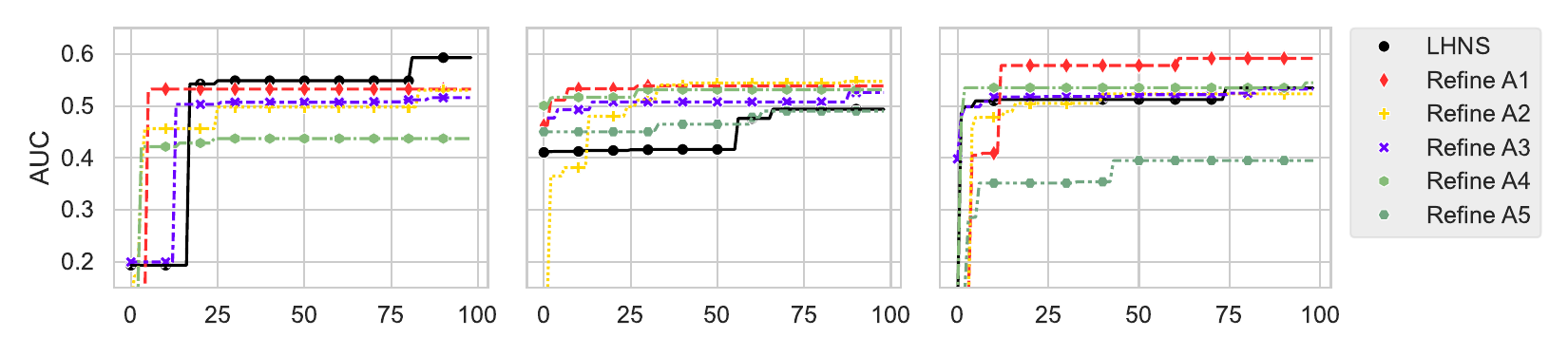}
}
    
    \caption{
    \textcolor{black}{AUC values obtained by local search method LHNS with six different initial baseline heuristics on F2, F10, F17, F19, and F22 of the pbo suite (from Top to Bottom). The $x$-axis represents the cumulative number of algorithms generated by the LLM, and the $y$-axis indicates the best-so-far AUC value. The results are obtained using  GPT 5 Nano, \textcolor{black}{Gemini 2.5 Flash-Lite}, and Qwen3 Coder Flash (from Left to Right).}}
    \label{fig:app:refine rest LHNS}

\end{figure*}

\begin{figure*}[!htb]
    \centering
\subfloat[\label{fig:refine-f2-ll}]{
    \includegraphics[width=0.945\textwidth]{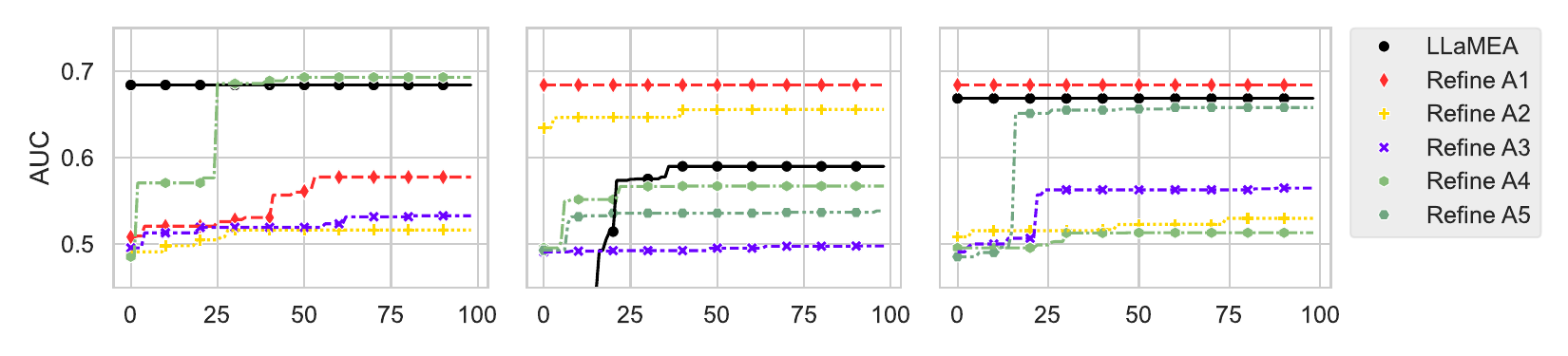}
}

\subfloat[\label{fig:refine-f10-ll-app}]{
    \includegraphics[width=0.945\textwidth]{figures_new/Refine/refine_pid_10_LL.pdf}
}
    
\subfloat[\label{fig:refine-f17-ll}]{
    \includegraphics[width=0.945\textwidth]{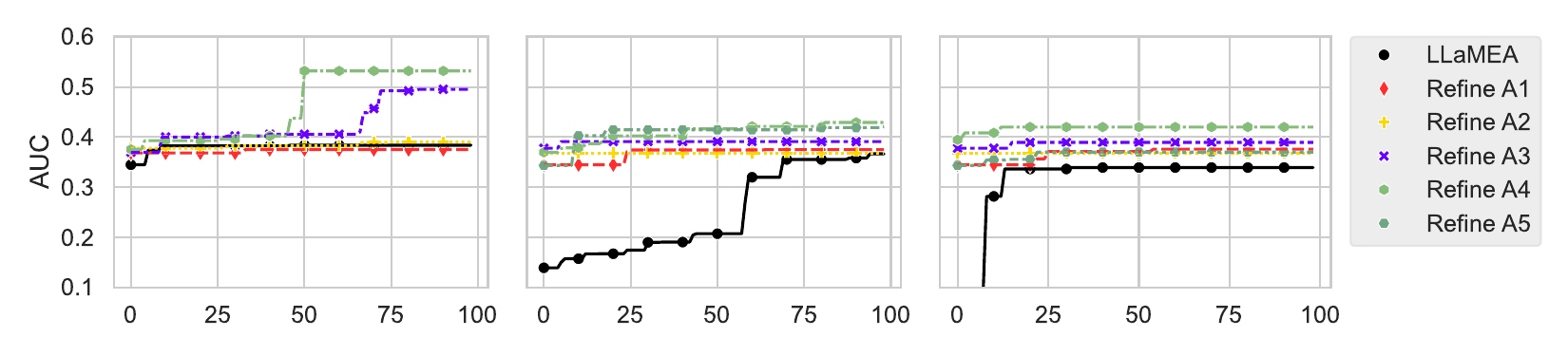}
}

\subfloat[\label{fig:refine-f19-ll}]{
    \includegraphics[width=0.945\textwidth]{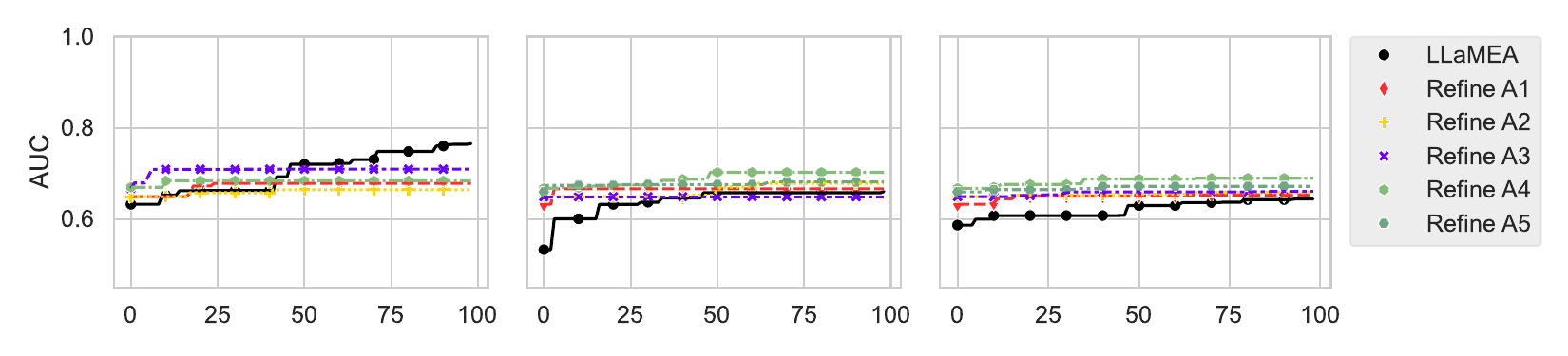}
}

\subfloat[\label{fig:refine-f22-ll}]{
    \includegraphics[width=0.945\textwidth]{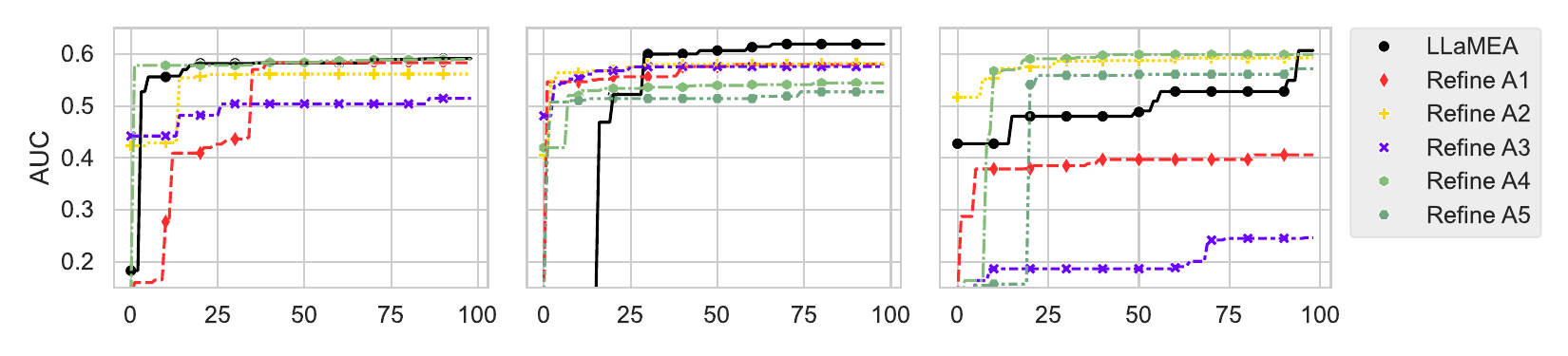}
}
    
    \caption{
    \textcolor{black}{AUC values obtained by \textit{refinement-only} LLaMEA with six different initial baseline heuristics on F2, F10, F17, F19, and F22 of the pbo suite (from Top to Bottom). The $x$-axis represents the cumulative number of algorithms generated by the LLM, and the $y$-axis indicates the best-so-far AUC value. The results are obtained using  GPT 5 Nano, Gemini 2.5 Flash-Lite, and Qwen3 Coder Flash (from Left to Right).}}
    \label{fig:app:refine rest LLaMEA}

\end{figure*}

We provide additional results in Figures~\ref{fig:app:refine rest LHNS} and~\ref{fig:app:refine rest LLaMEA}, comparing the AUC values of the algorithms obtained by \textcolor{black}{LHNS, LLaMEA} and our refinement strategy, which explicitly queries LLMs to refine the five provided benchmark codes, respectively. In addition to the results introduced in Section~\ref{sec:refine experiments}, we report further experiments on \textcolor{black}{OneMax (F1), LeadingOnes (F2) and its variant (F17)}, an Ising model (F19), and MIVS (F22), covering both theory-oriented and practical scenarios. 

\section{\textcolor{black}{Comparison with the Example Solvers}}
\label{appendix:comp-example}
\begin{figure*}
    \centering
    \includegraphics[width=\linewidth]{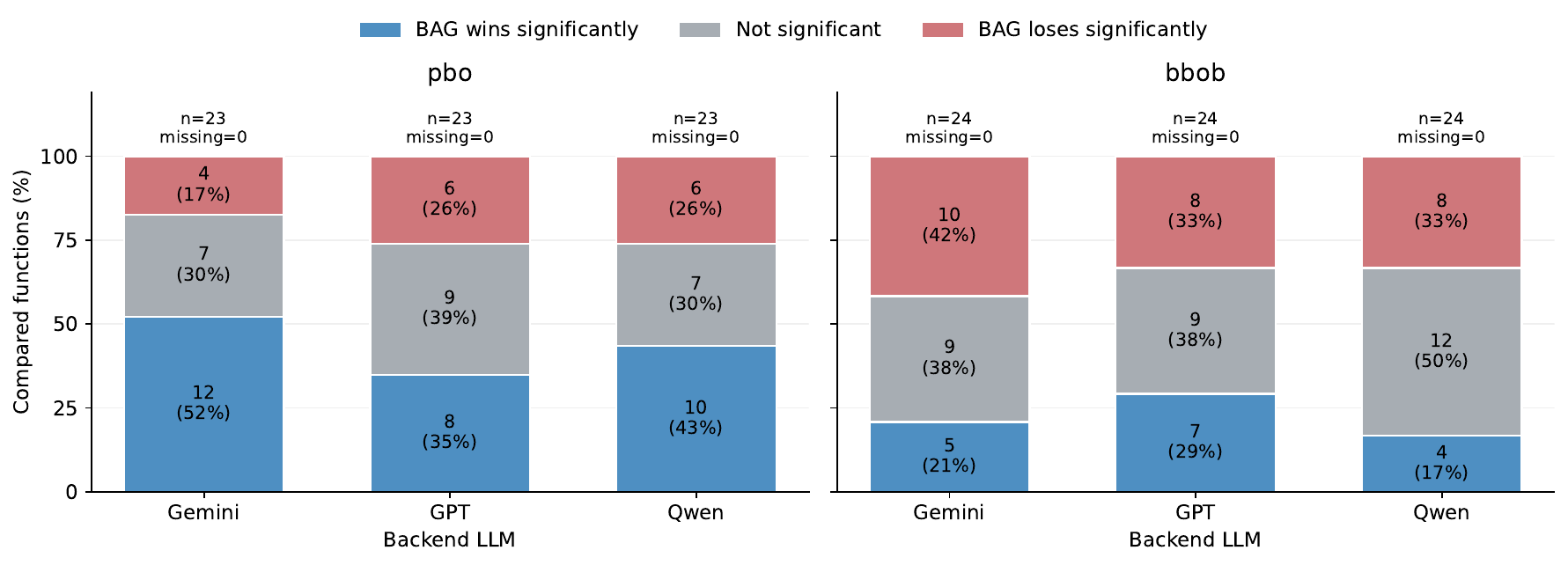}
    \caption{\textcolor{black}{
    Function-level comparison between BAG’s final solver and the best-performing example solver in $\mathbf{A}_{bench}$ on the pbo and bbob suites under each backend LLM. For each function and backend, the reference example is selected \textbf{retrospectively} according to the highest mean AUC over the same 20 unseen test instances used for the comparison. Bars report the counts and proportions of significant BAG wins, nonsignificant differences, and significant BAG losses, determined through two-sided exact Wilcoxon signed-rank tests with Holm-adjusted \(p\)-values \((\alpha=0.05)\). Wins and losses favor BAG and the selected example solver, respectively; nonsignificant differences do not establish performance equivalence.}}
    \label{fig:bag_vs_best_example}
\end{figure*}

\textcolor{black}{
To further analyze the relationship between the supplied solver knowledge and the performance of BAG-generated algorithms, we compare the final solver produced by BAG with the best-performing example solver in \(\mathbf{A}_{bench}\) for each benchmark function and backend LLM. This comparison examines whether BAG can generate solvers that improve upon the provided examples, rather than simply benefiting from their existing performance. BAG’s final solver and the five example solvers are evaluated on the same 20 unseen test instances (see Appendix~\ref{appendix:testResult}) under the same evaluation budget (see Table~\ref{tab:app:setup pbo and bbob}). The compared example is chosen according to the highest mean AUC over the 20 test instances after all results are known, which essentially provides us a function-specific best-example reference. Results of statistical significance are determined using two-sided exact Wilcoxon signed-rank tests on the paired AUC values, with Holm-corrected \(p\)-values at \(\alpha=0.05\). A function is counted as a win or loss when the difference is statistically significant in favor of BAG or the selected example solver, respectively, and as a tie otherwise.
\\
Figure~\ref{fig:bag_vs_best_example} presents the function-level comparison results. On pbo, BAG achieves more significant wins than losses across all three backend LLMs, with significant improvements on 8 to 12 of the 23 functions. On bbob, BAG achieves significant improvements on 4 to 7 of the 24 functions, although the best-performing example solvers show an advantage on a larger number of functions. These findings highlight the importance of the supplied solver knowledge, while also showing that BAG can generate algorithms that outperform the best observed example on particular functions. This is consistent with our core motivation: existing solver code provides useful prior knowledge for LLM-based algorithm generation and can serve as a starting point for further algorithmic improvement.
}

\section{\textcolor{black}{Proportion of Failed Code Generation}}
\label{appendix:bugcode}
\begin{figure*}[!ht]
    \centering
    \includegraphics[width=0.67\textwidth]{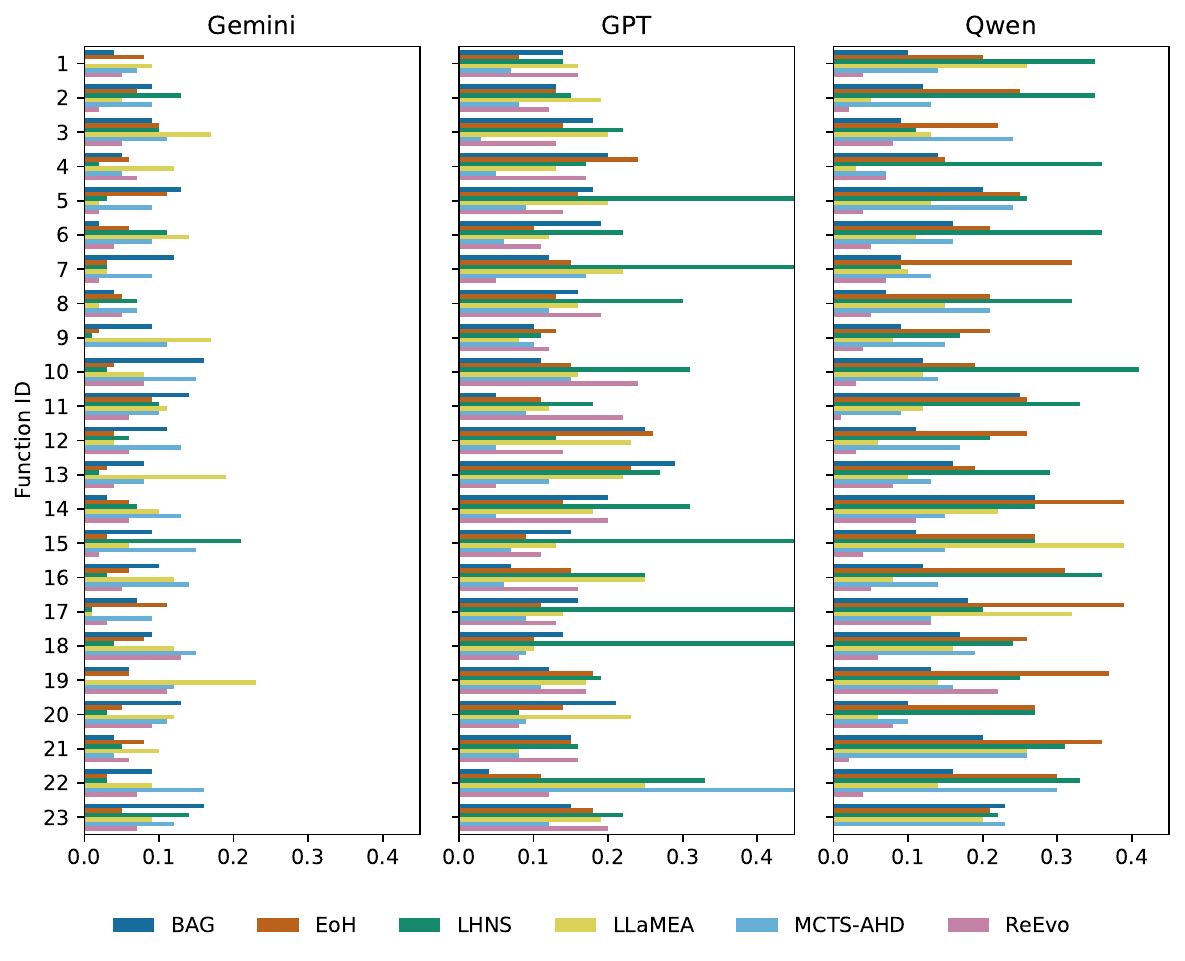}
    \caption{\textcolor{black}{Averaged proportions of failed code generation across 23 pbo problems for the compared LLM-driven approaches. The performances of using Gemini, GPT, and Qwen are plotted from left to right.}}
    \label{fig:bug rate pbo}
\end{figure*}

\begin{figure*}[!ht]
    \centering
    \includegraphics[width=0.67\textwidth]{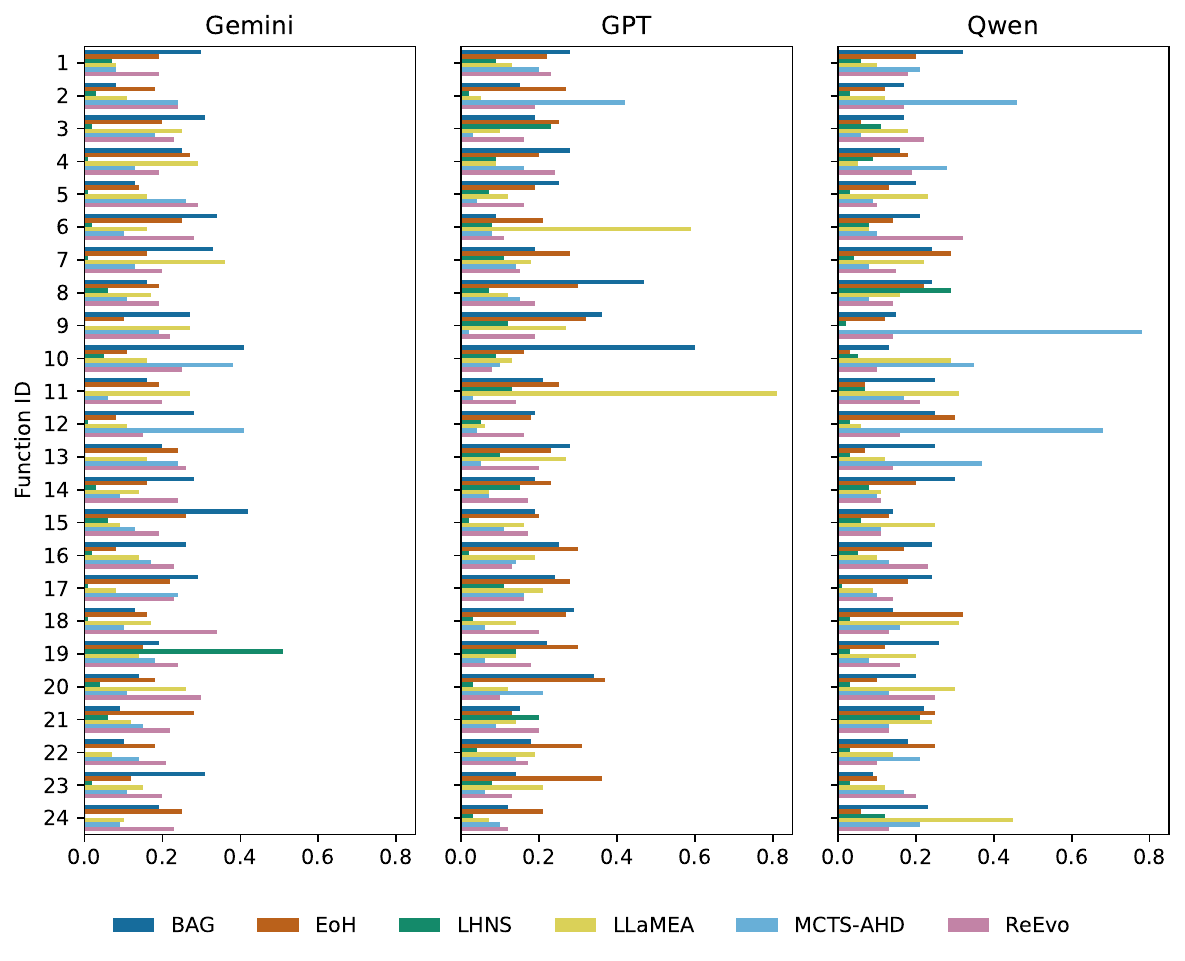}
    \caption{\textcolor{black}{Averaged proportions of failed code generation across 24 bbob problems for the compared LLM-driven approaches. The performances of using Gemini, GPT, and Qwen are plotted from left to right.}}
    \label{fig:bug rate bbob}
\end{figure*}

\textcolor{black}{Although LLMs are capable of generating algorithm codes, they may still produce programmes that fail to execute due to syntax errors or other logic issues. In our experiments, we set a maximum $3000$s CPU time limit for testing each problem. Note that we conduct 5 independent runs ($600$s) for each problem. Runs that either cannot be executed correctly or fail to produce any results within this limit are assigned a negative infinity fitness value (the worst possible). Figures~\ref{fig:bug rate pbo} and ~\ref{fig:bug rate bbob} present the average proportions of failed code generation within $100$ 
\textcolor{black}{full algorithm proposal budget}
for each tested problem. The plots do not reveal a clear difference among the compared LLM-driven optimization methods, as such failures primarily rely on the underlying LLM models rather than the proposed LLM-driven optimization frameworks.}

\newpage
\begin{table}
\centering
\caption{\textcolor{black}{The best normalized (higher is better) AUC achieved by four LLM-driven approaches on 23 pbo problems.
Results are obtained using Gemini 2.5 Flash Lite. 
Each normalized AUC value is followed by its corresponding rank in brackets. 
The best entries are underlined. 
}}
\label{tab:results_PBO_gemini-2.5-flash-lite}
\resizebox{0.48\textwidth}{!}{
\begin{tabular}{c|ccccccc}
\toprule
Function ID & BAG & EoH & LHNS & LLaMEA & MCTS-AHD & ReEvo & Shinka \\
\midrule
F1 & 0.983 (3) & \underline{1.000} (1) & 0.764 (7) & 0.963 (4) & 0.946 (6) & 0.957 (5) & 0.990 (2) \\
F2 & 0.852 (5) & 0.832 (6) & 0.995 (3) & 0.995 (3) & 0.728 (7) & \underline{1.000} (1) & 0.999 (2) \\
F3 & 0.959 (3) & 0.959 (4) & 0.927 (6) & 0.967 (2) & 0.956 (5) & \underline{1.000} (1) & 0.885 (7) \\
F4 & 0.981 (2) & 0.979 (4) & 0.824 (7) & 0.976 (5) & 0.961 (6) & \underline{1.000} (1) & 0.981 (2) \\
F5 & 0.954 (2) & 0.933 (3) & 0.623 (7) & 0.900 (6) & 0.900 (5) & \underline{1.000} (1) & 0.919 (4) \\
F6 & 0.965 (3) & 0.966 (2) & 0.763 (7) & 0.956 (4) & 0.937 (5) & \underline{1.000} (1) & 0.913 (6) \\
F7 & 0.983 (3) & 0.926 (6) & - (7) & 0.995 (2) & 0.976 (5) & \underline{1.000} (1) & 0.979 (4) \\
F8 & 0.970 (3) & \underline{1.000} (1) & 0.771 (7) & 0.963 (4) & 0.941 (5) & 0.996 (2) & 0.930 (6) \\
F9 & \underline{1.000} (1) & 0.966 (2) & 0.941 (5) & 0.950 (3) & 0.949 (4) & 0.934 (6) & 0.872 (7) \\
F10 & 0.989 (4) & \underline{1.000} (1) & 0.643 (7) & 0.989 (3) & 0.974 (6) & 0.984 (5) & 0.996 (2) \\
F11 & 0.991 (3) & 0.911 (4) & 0.996 (2) & 0.831 (6) & 0.863 (5) & \underline{1.000} (1) & 0.812 (7) \\
F12 & \underline{1.000} (1) & 0.838 (4) & 0.706 (7) & 0.846 (3) & 0.729 (6) & 0.834 (5) & 0.976 (2) \\
F13 & 0.901 (4) & 0.922 (3) & 0.708 (6) & 0.894 (5) & 0.922 (2) & \underline{1.000} (1) & 0.637 (7) \\
F14 & 0.892 (2) & 0.737 (4) & 0.316 (7) & \underline{1.000} (1) & 0.674 (5) & 0.801 (3) & 0.534 (6) \\
F15 & \underline{1.000} (1) & 0.758 (4) & 0.217 (7) & 0.818 (3) & 0.723 (5) & 0.823 (2) & 0.646 (6) \\
F16 & \underline{1.000} (1) & 0.947 (4) & 0.673 (7) & 0.954 (2) & 0.864 (5) & 0.954 (3) & 0.740 (6) \\
F17 & 0.901 (2) & \underline{1.000} (1) & 0.130 (7) & 0.817 (3) & 0.784 (4) & 0.430 (6) & 0.615 (5) \\
F18 & \underline{1.000} (1) & 0.969 (3) & 0.832 (7) & 0.975 (2) & 0.874 (5) & 0.929 (4) & 0.840 (6) \\
F19 & 0.920 (2) & 0.874 (4) & 0.642 (7) & \underline{1.000} (1) & 0.888 (3) & 0.869 (5) & 0.851 (6) \\
F20 & \underline{1.000} (1) & 0.990 (2) & 0.668 (7) & 0.902 (5) & 0.891 (6) & 0.982 (4) & 0.986 (3) \\
F21 & \underline{1.000} (1) & 0.932 (2) & 0.779 (5) & 0.765 (6) & 0.850 (3) & 0.783 (4) & 0.611 (7) \\
F22 & \underline{1.000} (1) & 0.832 (7) & 0.884 (5) & 0.942 (3) & 0.854 (6) & 0.885 (4) & 0.950 (2) \\
F23 & \underline{1.000} (1) & 0.999 (2) & 0.834 (6) & 0.835 (5) & 0.814 (7) & 0.989 (3) & 0.933 (4) \\
\midrule \
Mean & 0.967 & 0.925 & 0.711 & 0.923 & 0.869 & 0.920 & 0.852 \\
Std & 0.042 & 0.076 & 0.224 & 0.070 & 0.087 & 0.126 & 0.144 \\
\midrule \
Average Rank & \underline{2.12} & 3.17 & 6.25 & 3.50 & 5.04 & 3.04 & 4.79 \\
\bottomrule
\end{tabular}
}
\end{table}
\begin{table}
\centering
\caption{
\textcolor{black}{
The best normalized (higher is better) AUC achieved by four LLM-driven approaches on 23 pbo problems.
Results are obtained using GPT 5 Nano. 
Each normalized AUC value is followed by its corresponding rank in brackets. 
The best entries are underlined. }}
\label{tab:results_PBO_gpt-5-nano}
\resizebox{0.48\textwidth}{!}{
\begin{tabular}{c|ccccccc}
\toprule
Function ID & BAG & EoH & LHNS & LLaMEA & MCTS-AHD & ReEvo & Shinka\\
\midrule
F1 & 0.988 (4) & 0.992 (2) & 0.534 (7) & \underline{1.000} (1) & 0.939 (6) & 0.992 (3) & 0.946 (5) \\
F2 & 0.860 (2) & 0.674 (6) & 0.704 (4) & 0.725 (3) & 0.620 (7) & 0.702 (5) & \underline{1.000} (1) \\
F3 & 0.995 (2) & 0.971 (4) & 0.959 (6) & \underline{1.000} (1) & 0.967 (5) & 0.980 (3) & 0.934 (7) \\
F4 & 0.937 (4) & 0.922 (6) & 0.939 (3) & \underline{1.000} (1) & 0.922 (5) & 0.956 (2) & 0.911 (7) \\
F5 & 0.928 (5) & 0.978 (2) & \underline{1.000} (1) & 0.900 (7) & 0.939 (4) & 0.962 (3) & 0.927 (6) \\
F6 & 0.970 (3) & 0.993 (2) & 0.820 (7) & 0.968 (4) & 0.967 (5) & 0.951 (6) & \underline{1.000} (1) \\
F7 & 0.922 (5) & 0.988 (2) & 0.877 (6) & \underline{1.000} (1) & 0.979 (3) & 0.929 (4) & 0.855 (7) \\
F8 & 0.993 (2) & 0.953 (4) & 0.936 (7) & \underline{1.000} (1) & 0.952 (5) & 0.944 (6) & 0.961 (3) \\
F9 & 0.946 (3) & 0.936 (5) & 0.832 (7) & 0.941 (4) & \underline{1.000} (1) & 0.954 (2) & 0.930 (6) \\
F10 & 0.927 (4) & 0.916 (5) & 0.723 (7) & \underline{1.000} (1) & 0.982 (2) & 0.880 (6) & 0.956 (3) \\
F11 & 0.658 (4) & 0.704 (2) & \underline{1.000} (1) & 0.586 (7) & 0.634 (6) & 0.658 (5) & 0.699 (3) \\
F12 & 0.997 (2) & \underline{1.000} (1) & 0.690 (7) & 0.727 (6) & 0.972 (3) & 0.768 (5) & 0.912 (4) \\
F13 & 0.925 (3) & 0.922 (4) & \underline{1.000} (1) & 0.917 (5) & 0.949 (2) & 0.865 (7) & 0.873 (6) \\
F14 & 0.695 (4) & \underline{1.000} (1) & 0.252 (7) & 0.833 (2) & 0.804 (3) & 0.539 (6) & 0.606 (5) \\
F15 & \underline{1.000} (1) & 0.908 (2) & 0.590 (7) & 0.760 (5) & 0.743 (6) & 0.841 (4) & 0.855 (3) \\
F16 & 0.982 (2) & 0.858 (5) & 0.646 (7) & 0.938 (4) & \underline{1.000} (1) & 0.945 (3) & 0.774 (6) \\
F17 & \underline{1.000} (1) & 0.974 (3) & 0.333 (7) & 0.931 (5) & 0.924 (6) & 0.970 (4) & 0.977 (2) \\
F18 & 0.916 (4) & 0.957 (3) & - (7) & \underline{1.000} (1) & 0.869 (5) & 0.972 (2) & 0.855 (6) \\
F19 & 0.812 (7) & 0.823 (6) & 0.968 (2) & 0.826 (5) & 0.858 (3) & 0.856 (4) & \underline{1.000} (1) \\
F20 & 0.943 (5) & 0.975 (2) & \underline{1.000} (1) & 0.933 (6) & 0.902 (7) & 0.952 (4) & 0.962 (3) \\
F21 & 0.904 (4) & 0.934 (2) & 0.827 (6) & 0.875 (5) & \underline{1.000} (1) & 0.908 (3) & 0.684 (7) \\
F22 & 0.923 (4) & 0.994 (2) & 0.789 (5) & 0.989 (3) & 0.030 (7) & \underline{1.000} (1) & 0.329 (6) \\
F23 & 0.901 (2) & 0.885 (3) & 0.737 (7) & 0.783 (6) & 0.855 (4) & \underline{1.000} (1) & 0.840 (5) \\
\midrule \
Mean & 0.918 & 0.924 & 0.780 & 0.897 & 0.861 & 0.892 & 0.860 \\
Std & 0.088 & 0.086 & 0.206 & 0.112 & 0.206 & 0.117 & 0.153 \\
\midrule \
Average Rank & 3.29 & \underline{3.12} & 5.29 & 3.62 & 4.25 & 3.88 & 4.54 \\
\bottomrule
\end{tabular}
}
\end{table}
\begin{table}
\centering
\caption{
\textcolor{black}{The best normalized (higher is better) AUC achieved by four LLM-driven approaches on 23 pbo problems.
Results are obtained using Qwen3 Coder Flash. 
Each normalized AUC value is followed by its corresponding rank in brackets. 
The best entries are underlined.}
}
\label{tab:results_PBO_qwen3-coder-flash}
\resizebox{0.48\textwidth}{!}{
\begin{tabular}{c|ccccccc}
\toprule
Function ID & BAG & EoH & LHNS & LLaMEA & MCTS-AHD & ReEvo & Shinka\\
\midrule
F1 & 0.973 (3) & \underline{1.000} (1) & 0.877 (7) & 0.970 (4) & 0.977 (2) & 0.944 (5) & 0.926 (6) \\
F2 & 0.773 (5) & 0.830 (2) & 0.580 (7) & 0.811 (4) & 0.824 (3) & 0.591 (6) & \underline{1.000} (1) \\
F3 & 0.970 (4) & 0.977 (2) & 0.860 (7) & 0.977 (2) & 0.964 (5) & 0.940 (6) & \underline{1.000} (1) \\
F4 & 0.980 (3) & 0.986 (2) & 0.911 (7) & 0.932 (6) & \underline{1.000} (1) & 0.938 (5) & 0.965 (4) \\
F5 & 0.990 (2) & 0.971 (4) & 0.904 (7) & 0.943 (5) & 0.989 (3) & 0.922 (6) & \underline{1.000} (1) \\
F6 & 0.968 (5) & 0.973 (4) & 0.924 (7) & 0.957 (6) & 0.982 (3) & 0.986 (2) & \underline{1.000} (1) \\
F7 & \underline{1.000} (1) & 0.958 (4) & 0.864 (7) & 0.978 (3) & 0.993 (2) & 0.931 (6) & 0.948 (5) \\
F8 & \underline{1.000} (1) & 0.955 (3) & 0.888 (7) & 0.916 (5) & 0.909 (6) & 0.969 (2) & 0.941 (4) \\
F9 & \underline{1.000} (2) & 0.979 (5) & 0.944 (7) & 0.962 (6) & 0.997 (3) & 0.996 (4) & \underline{1.000} (1) \\
F10 & 0.991 (2) & 0.987 (3) & 0.785 (7) & 0.966 (6) & 0.978 (4) & \underline{1.000} (1) & 0.972 (5) \\
F11 & 0.951 (2) & \underline{1.000} (1) & 0.942 (3) & 0.931 (4) & 0.927 (5) & 0.835 (7) & 0.920 (6) \\
F12 & 0.946 (5) & \underline{1.000} (2) & 0.960 (4) & 0.960 (3) & 0.829 (6) & 0.828 (7) & \underline{1.000} (1) \\
F13 & \underline{1.000} (1) & 0.998 (2) & 0.913 (7) & 0.984 (4) & 0.981 (5) & 0.968 (6) & 0.987 (3) \\
F14 & 0.649 (4) & 0.762 (2) & 0.447 (7) & 0.732 (3) & \underline{1.000} (1) & 0.621 (5) & 0.556 (6) \\
F15 & 0.887 (2) & 0.802 (5) & 0.734 (7) & 0.802 (4) & 0.774 (6) & 0.806 (3) & \underline{1.000} (1) \\
F16 & \underline{1.000} (1) & 0.865 (4) & 0.903 (3) & 0.768 (6) & 0.984 (2) & 0.856 (5) & 0.754 (7) \\
F17 & \underline{1.000} (1) & 0.929 (4) & 0.364 (7) & 0.891 (5) & 0.961 (3) & 0.867 (6) & 0.967 (2) \\
F18 & 0.987 (4) & 0.977 (5) & \underline{1.000} (1) & 0.992 (2) & 0.898 (6) & 0.842 (7) & 0.990 (3) \\
F19 & \underline{1.000} (1) & 0.922 (4) & 0.941 (3) & 0.947 (2) & 0.907 (5) & 0.884 (7) & 0.892 (6) \\
F20 & 0.885 (2) & 0.871 (3) & 0.747 (7) & \underline{1.000} (1) & 0.772 (6) & 0.818 (5) & 0.844 (4) \\
F21 & \underline{1.000} (1) & 0.984 (3) & 0.937 (4) & 0.823 (7) & 0.920 (5) & 0.868 (6) & 0.993 (2) \\
F22 & 0.945 (4) & \underline{1.000} (1) & 0.860 (5) & 0.976 (2) & 0.713 (6) & 0.946 (3) & 0.331 (7) \\
F23 & 0.962 (4) & 0.969 (3) & 0.632 (7) & 0.972 (2) & 0.695 (6) & 0.917 (5) & \underline{1.000} (1) \\
\midrule \
Mean & 0.950 & 0.943 & 0.822 & 0.921 & 0.912 & 0.882 & 0.912 \\
Std & 0.083 & 0.068 & 0.164 & 0.076 & 0.094 & 0.103 & 0.160 \\
\midrule \
Average Rank & \underline{2.54} & 2.96 & 5.92 & 3.96 & 4.12 & 5.04 & 3.42 \\
\bottomrule
\end{tabular}
}
\end{table}

\begin{table}
\centering
\caption{
\textcolor{black}{
The best normalized (higher is better) AUC achieved by four LLM-driven approaches on 24 bbob problems.
Results are obtained using Gemini 2.5 Flash Lite. 
Each normalized AUC value is followed by its corresponding rank in brackets. 
The best entries are underlined.}
}
\label{tab:results_BBOB_gemini-2.5-flash-lite}
\resizebox{0.48\textwidth}{!}{
\begin{tabular}{c|ccccccc}
\toprule
Function ID & BAG & EoH & LHNS & LLaMEA & MCTS-AHD & ReEvo & Shinka\\
\midrule
F1 & 0.994 (2) & 0.991 (3) & 0.976 (5) & 0.904 (6) & 0.990 (4) & \underline{1.000} (1) & 0.603 (7) \\
F2 & 0.876 (3) & 0.881 (2) & 0.025 (7) & \underline{1.000} (1) & 0.787 (5) & 0.805 (4) & 0.042 (6) \\
F3 & 0.608 (2) & 0.559 (3) & 0.151 (6) & \underline{1.000} (1) & 0.452 (5) & 0.493 (4) & 0.135 (7) \\
F4 & 0.240 (5) & 0.404 (3) & 0.118 (7) & \underline{1.000} (1) & 0.670 (2) & 0.352 (4) & 0.187 (6) \\
F5 & 0.996 (5) & \underline{1.000} (1) & 0.936 (7) & 0.998 (4) & 0.995 (6) & 0.999 (2) & 0.998 (3) \\
F6 & 0.960 (5) & \underline{1.000} (1) & 0.757 (6) & 0.972 (4) & 0.972 (3) & 0.990 (2) & 0.520 (7) \\
F7 & 0.990 (2) & 0.922 (3) & 0.724 (6) & \underline{1.000} (1) & 0.884 (4) & 0.800 (5) & 0.200 (7) \\
F8 & 0.944 (2) & 0.937 (3) & 0.352 (7) & \underline{1.000} (1) & 0.365 (6) & 0.753 (4) & 0.377 (5) \\
F9 & 0.862 (3) & 0.504 (6) & 0.411 (7) & 0.950 (2) & \underline{1.000} (1) & 0.653 (4) & 0.558 (5) \\
F10 & 0.672 (3) & 0.766 (2) & 0.054 (6) & \underline{1.000} (1) & 0.050 (7) & 0.662 (4) & 0.056 (5) \\
F11 & 0.762 (3) & 0.699 (4) & 0.519 (6) & 0.662 (5) & \underline{1.000} (1) & 0.904 (2) & 0.119 (7) \\
F12 & \underline{1.000} (1) & 0.408 (5) & 0.350 (6) & 0.777 (2) & - (7) & 0.550 (4) & 0.560 (3) \\
F13 & 0.966 (3) & 0.999 (2) & 0.371 (7) & \underline{1.000} (1) & 0.547 (5) & 0.842 (4) & 0.373 (6) \\
F14 & 0.802 (4) & 0.870 (3) & 0.566 (7) & 0.881 (2) & 0.791 (5) & \underline{1.000} (1) & 0.647 (6) \\
F15 & 0.960 (2) & 0.826 (4) & 0.866 (3) & 0.770 (6) & \underline{1.000} (1) & 0.782 (5) & 0.732 (7) \\
F16 & 0.775 (3) & 0.632 (6) & 0.741 (4) & 0.737 (5) & 0.834 (2) & \underline{1.000} (1) & 0.602 (7) \\
F17 & 0.932 (2) & 0.925 (3) & 0.561 (6) & \underline{1.000} (1) & 0.720 (5) & 0.778 (4) & 0.346 (7) \\
F18 & \underline{1.000} (1) & 0.726 (2) & 0.241 (7) & 0.670 (4) & 0.687 (3) & 0.548 (5) & 0.307 (6) \\
F19 & \underline{1.000} (2) & \underline{1.000} (1) & 0.531 (7) & 0.640 (5) & 0.736 (3) & 0.668 (4) & 0.616 (6) \\
F20 & \underline{1.000} (1) & 0.834 (5) & 0.372 (7) & 0.855 (4) & 0.917 (2) & 0.858 (3) & 0.383 (6) \\
F21 & 0.981 (2) & 0.734 (4) & 0.382 (7) & \underline{1.000} (1) & 0.834 (3) & 0.733 (5) & 0.656 (6) \\
F22 & 0.831 (4) & 0.838 (3) & 0.456 (7) & 0.754 (5) & 0.920 (2) & \underline{1.000} (1) & 0.689 (6) \\
F23 & 0.765 (2) & 0.762 (3) & 0.742 (4) & 0.636 (7) & 0.681 (6) & \underline{1.000} (1) & 0.702 (5) \\
F24 & \underline{1.000} (1) & 0.875 (3) & 0.638 (7) & 0.912 (2) & 0.825 (5) & 0.861 (4) & 0.724 (6) \\
\midrule \
Mean & 0.871 & 0.796 & 0.493 & 0.880 & 0.768 & 0.793 & 0.464 \\
Std & 0.173 & 0.179 & 0.264 & 0.133 & 0.231 & 0.180 & 0.247 \\
\midrule \
Average Rank & \underline{2.60} & 3.12 & 6.20 & 2.92 & 3.92 & 3.28 & 5.96 \\
\bottomrule
\end{tabular}
}
\end{table}

\begin{table}
\centering
\caption{
\textcolor{black}{The best normalized (higher is better) AUC achieved by four LLM-driven approaches on 24 bbob problems.
Results are obtained using GPT 5 Nano. 
Each normalized AUC value is followed by its corresponding rank in brackets. 
The best entries are underlined.}
}
\label{tab:results_BBOB_gpt-5-nano}
\resizebox{0.48\textwidth}{!}{
\begin{tabular}{c|ccccccc}
\toprule
Function ID & BAG & EoH & LHNS & LLaMEA & MCTS-AHD & ReEvo & Shinka\\
\midrule
F1 & 0.985 (4) & \underline{1.000} (1) & 0.987 (2) & 0.986 (3) & 0.985 (4) & 0.977 (6) & 0.930 (7) \\
F2 & \underline{1.000} (1) & 0.955 (4) & 0.000 (7) & 0.944 (5) & 0.771 (6) & 0.988 (2) & 0.975 (3) \\
F3 & 0.760 (3) & 0.505 (4) & 0.117 (7) & 0.413 (5) & 0.790 (2) & 0.330 (6) & \underline{1.000} (1) \\
F4 & 0.862 (2) & 0.704 (3) & 0.180 (7) & \underline{1.000} (1) & 0.552 (4) & 0.182 (6) & 0.250 (5) \\
F5 & 0.999 (4) & \underline{1.000} (3) & 0.987 (7) & 0.999 (5) & 0.995 (6) & \underline{1.000} (1) & \underline{1.000} (1) \\
F6 & 0.987 (4) & 0.996 (3) & 0.689 (7) & 0.999 (2) & 0.839 (5) & \underline{1.000} (1) & 0.832 (6) \\
F7 & \underline{1.000} (1) & 0.564 (4) & 0.319 (7) & 0.967 (2) & 0.536 (5) & 0.418 (6) & 0.569 (3) \\
F8 & 0.839 (2) & 0.276 (5) & 0.290 (4) & \underline{1.000} (1) & 0.230 (7) & 0.272 (6) & 0.298 (3) \\
F9 & 0.800 (3) & 0.297 (6) & 0.120 (7) & 0.968 (2) & \underline{1.000} (1) & 0.563 (4) & 0.496 (5) \\
F10 & \underline{1.000} (1) & 0.080 (7) & 0.090 (6) & 0.723 (2) & 0.633 (3) & 0.090 (5) & 0.186 (4) \\
F11 & \underline{1.000} (1) & 0.144 (3) & 0.134 (4) & 0.266 (2) & 0.130 (6) & 0.120 (7) & 0.134 (4) \\
F12 & \underline{1.000} (1) & 0.295 (7) & 0.320 (6) & 0.370 (5) & 0.414 (2) & 0.380 (4) & 0.382 (3) \\
F13 & \underline{1.000} (1) & 0.330 (4) & 0.186 (7) & 0.314 (6) & 0.318 (5) & 0.350 (3) & 0.454 (2) \\
F14 & \underline{1.000} (1) & 0.720 (3) & 0.619 (7) & 0.822 (2) & 0.710 (4) & 0.654 (5) & 0.652 (6) \\
F15 & 0.860 (3) & 0.437 (7) & 0.581 (5) & 0.958 (2) & 0.716 (4) & 0.445 (6) & \underline{1.000} (1) \\
F16 & \underline{1.000} (1) & 0.842 (2) & 0.340 (7) & 0.582 (4) & 0.382 (6) & 0.742 (3) & 0.479 (5) \\
F17 & 0.959 (2) & 0.610 (5) & 0.444 (6) & \underline{1.000} (1) & 0.672 (4) & 0.688 (3) & 0.392 (7) \\
F18 & \underline{1.000} (1) & 0.575 (5) & 0.585 (4) & 0.438 (6) & 0.387 (7) & 0.691 (2) & 0.600 (3) \\
F19 & 0.256 (4) & 0.247 (6) & \underline{1.000} (1) & 0.264 (3) & 0.250 (5) & 0.247 (7) & 0.276 (2) \\
F20 & \underline{1.000} (1) & 0.644 (6) & 0.660 (5) & 0.669 (4) & 0.816 (2) & 0.488 (7) & 0.710 (3) \\
F21 & 0.918 (3) & 0.853 (4) & 0.320 (7) & \underline{1.000} (1) & 0.793 (6) & 0.924 (2) & 0.832 (5) \\
F22 & \underline{1.000} (1) & 0.818 (4) & 0.610 (7) & 0.958 (2) & 0.903 (3) & 0.692 (6) & 0.746 (5) \\
F23 & \underline{1.000} (1) & 0.748 (2) & 0.591 (6) & 0.670 (3) & 0.602 (5) & 0.660 (4) & 0.547 (7) \\
F24 & 0.165 (2) & \underline{1.000} (1) & 0.119 (7) & 0.140 (4) & 0.132 (5) & 0.128 (6) & 0.157 (3) \\
\midrule \
Mean & 0.891 & 0.610 & 0.429 & 0.727 & 0.606 & 0.543 & 0.579 \\
Std & 0.218 & 0.287 & 0.294 & 0.294 & 0.267 & 0.293 & 0.282 \\
\midrule \
Average Rank & \underline{1.96} & 4.08 & 5.88 & 3.00 & 4.44 & 4.56 & 3.96 \\
\bottomrule
\end{tabular}
}
\end{table}

\begin{table}
\centering
\caption{
\textcolor{black}{
The best normalized (higher is better) AUC achieved by four LLM-driven approaches on 24 bbob problems.
Results are obtained using Qwen3 Coder Flash. 
Each normalized AUC value is followed by its corresponding rank in brackets. 
The best entries are underlined.}
}
\label{tab:results_BBOB_qwen3-coder-flash}
\resizebox{0.48\textwidth}{!}{
\begin{tabular}{c|ccccccc}
\toprule
Function ID & BAG & EoH & LHNS & LLaMEA & MCTS-AHD & ReEvo & Shinka\\
\midrule
F1 & 0.986 (3) & 0.994 (2) & 0.956 (6) & \underline{1.000} (1) & 0.475 (7) & 0.968 (5) & 0.979 (4) \\
F2 & \underline{1.000} (1) & 0.936 (5) & 0.937 (4) & 0.965 (3) & 0.270 (7) & 0.992 (2) & 0.310 (6) \\
F3 & \underline{1.000} (1) & 0.769 (3) & 0.518 (4) & 0.970 (2) & 0.182 (5) & 0.109 (7) & 0.146 (6) \\
F4 & 0.946 (3) & 0.566 (5) & 0.425 (7) & 0.883 (4) & 0.553 (6) & \underline{1.000} (1) & 0.994 (2) \\
F5 & 0.999 (2) & 0.999 (3) & 0.988 (7) & 0.999 (4) & 0.997 (5) & \underline{1.000} (1) & 0.996 (6) \\
F6 & \underline{1.000} (1) & 0.820 (3) & 0.866 (2) & 0.746 (5) & 0.540 (7) & 0.698 (6) & 0.749 (4) \\
F7 & 0.993 (2) & 0.882 (4) & 0.904 (3) & \underline{1.000} (1) & 0.564 (7) & 0.637 (6) & 0.832 (5) \\
F8 & \underline{1.000} (1) & 0.738 (3) & 0.733 (4) & 0.989 (2) & 0.565 (5) & 0.545 (6) & 0.472 (7) \\
F9 & 0.986 (2) & 0.867 (3) & 0.514 (6) & 0.572 (5) & 0.793 (4) & \underline{1.000} (1) & 0.184 (7) \\
F10 & 0.985 (2) & 0.377 (4) & 0.273 (5) & 0.768 (3) & 0.012 (7) & \underline{1.000} (1) & 0.064 (6) \\
F11 & \underline{1.000} (1) & 0.350 (5) & 0.366 (4) & 0.376 (3) & 0.179 (6) & 0.540 (2) & 0.115 (7) \\
F12 & \underline{1.000} (1) & 0.749 (3) & 0.393 (5) & 0.409 (4) & 0.185 (7) & 0.921 (2) & 0.286 (6) \\
F13 & \underline{1.000} (1) & 0.472 (4) & 0.579 (3) & 0.675 (2) & 0.358 (5) & 0.306 (7) & 0.327 (6) \\
F14 & \underline{1.000} (1) & 0.867 (2) & 0.775 (5) & 0.861 (3) & 0.839 (4) & 0.652 (6) & 0.598 (7) \\
F15 & \underline{1.000} (1) & 0.561 (3) & 0.305 (7) & 0.783 (2) & 0.502 (4) & 0.458 (5) & 0.392 (6) \\
F16 & 0.747 (3) & 0.871 (2) & 0.345 (7) & 0.746 (4) & 0.557 (5) & \underline{1.000} (1) & 0.479 (6) \\
F17 & 0.819 (4) & 0.854 (3) & 0.871 (2) & 0.645 (7) & 0.816 (5) & \underline{1.000} (1) & 0.684 (6) \\
F18 & \underline{1.000} (1) & 0.540 (4) & 0.616 (3) & 0.422 (7) & 0.452 (6) & 0.475 (5) & 0.667 (2) \\
F19 & \underline{1.000} (1) & \underline{1.000} (2) & 0.708 (7) & 0.934 (4) & 0.974 (3) & 0.848 (6) & 0.921 (5) \\
F20 & 0.772 (4) & 0.983 (2) & \underline{1.000} (1) & 0.622 (6) & 0.791 (3) & 0.700 (5) & 0.441 (7) \\
F21 & 0.944 (3) & 0.954 (2) & 0.590 (6) & 0.833 (5) & 0.918 (4) & 0.442 (7) & \underline{1.000} (1) \\
F22 & 0.991 (2) & \underline{1.000} (1) & 0.557 (6) & 0.896 (3) & 0.760 (4) & 0.557 (5) & 0.531 (7) \\
F23 & 0.856 (5) & 0.869 (4) & 0.626 (7) & 0.717 (6) & 0.876 (3) & \underline{1.000} (1) & 0.970 (2) \\
F24 & \underline{1.000} (1) & 0.864 (3) & 0.745 (7) & 0.884 (2) & 0.835 (4) & 0.810 (6) & 0.820 (5) \\
\midrule \
Mean & 0.959 & 0.787 & 0.650 & 0.779 & 0.583 & 0.736 & 0.582 \\
Std & 0.076 & 0.198 & 0.225 & 0.189 & 0.275 & 0.258 & 0.306 \\
\midrule \
Average Rank & \underline{1.92} & 3.08 & 4.92 & 3.64 & 5.16 & 3.96 & 5.32 \\
\bottomrule
\end{tabular}
}
\end{table}


\clearpage
\begin{figure*}[!htb]
    \centering
    \includegraphics[width=0.9\textwidth]{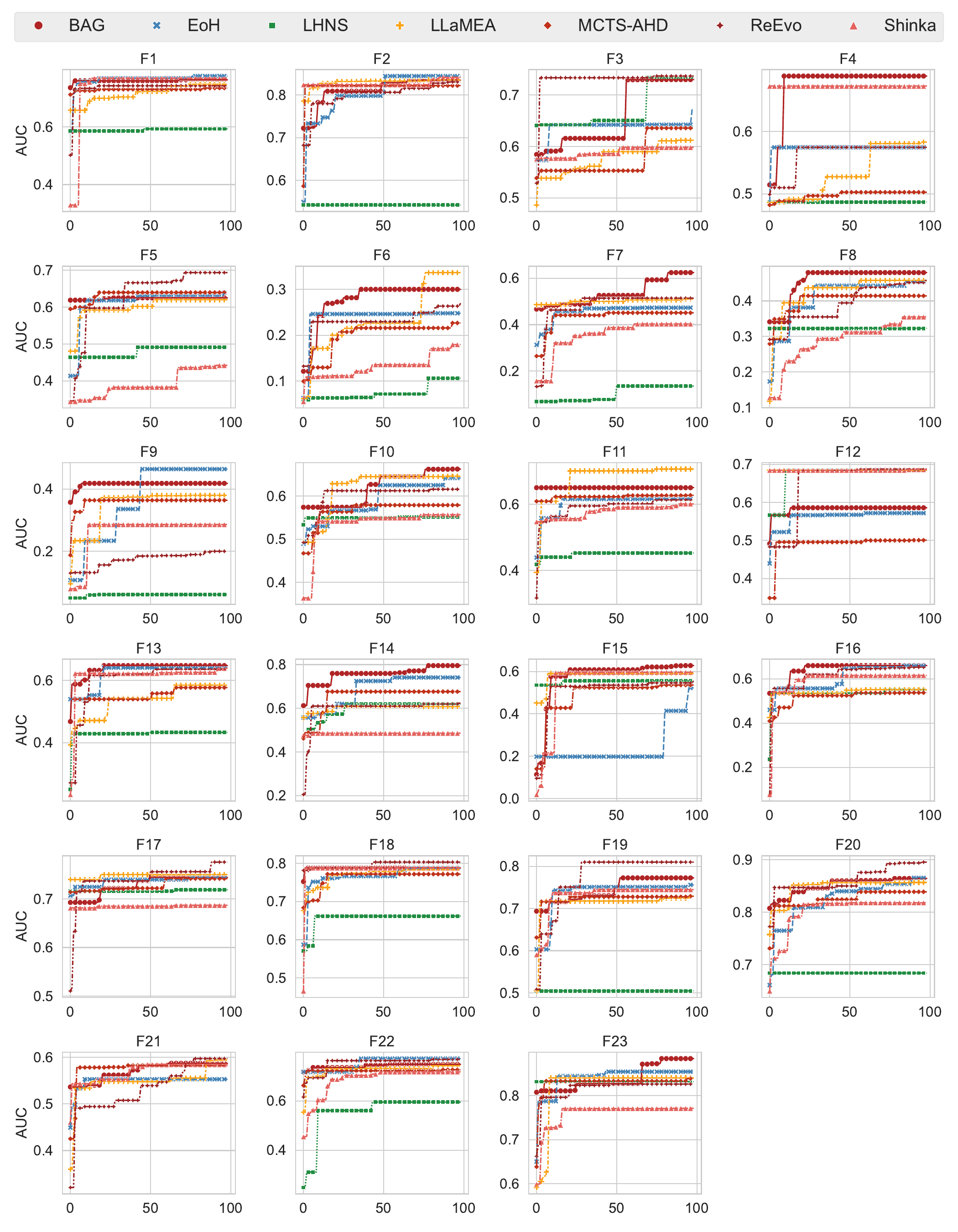}
    \caption{
    \textcolor{black}{
    AUC values of the algorithms obtained by LLM-driven approaches on all pbo problems. The $x$-axis represents the cumulative number of algorithms generated by the LLM, and the $y$-axis indicates the best-so-far AUC value. The results are obtained using Gemini 2.5 Flash-Lite. 
    }}
    \label{fig:convergence_pbo_gemini}
\end{figure*}

\begin{figure*}[!htb]
    \centering
    \includegraphics[width=0.9\textwidth]{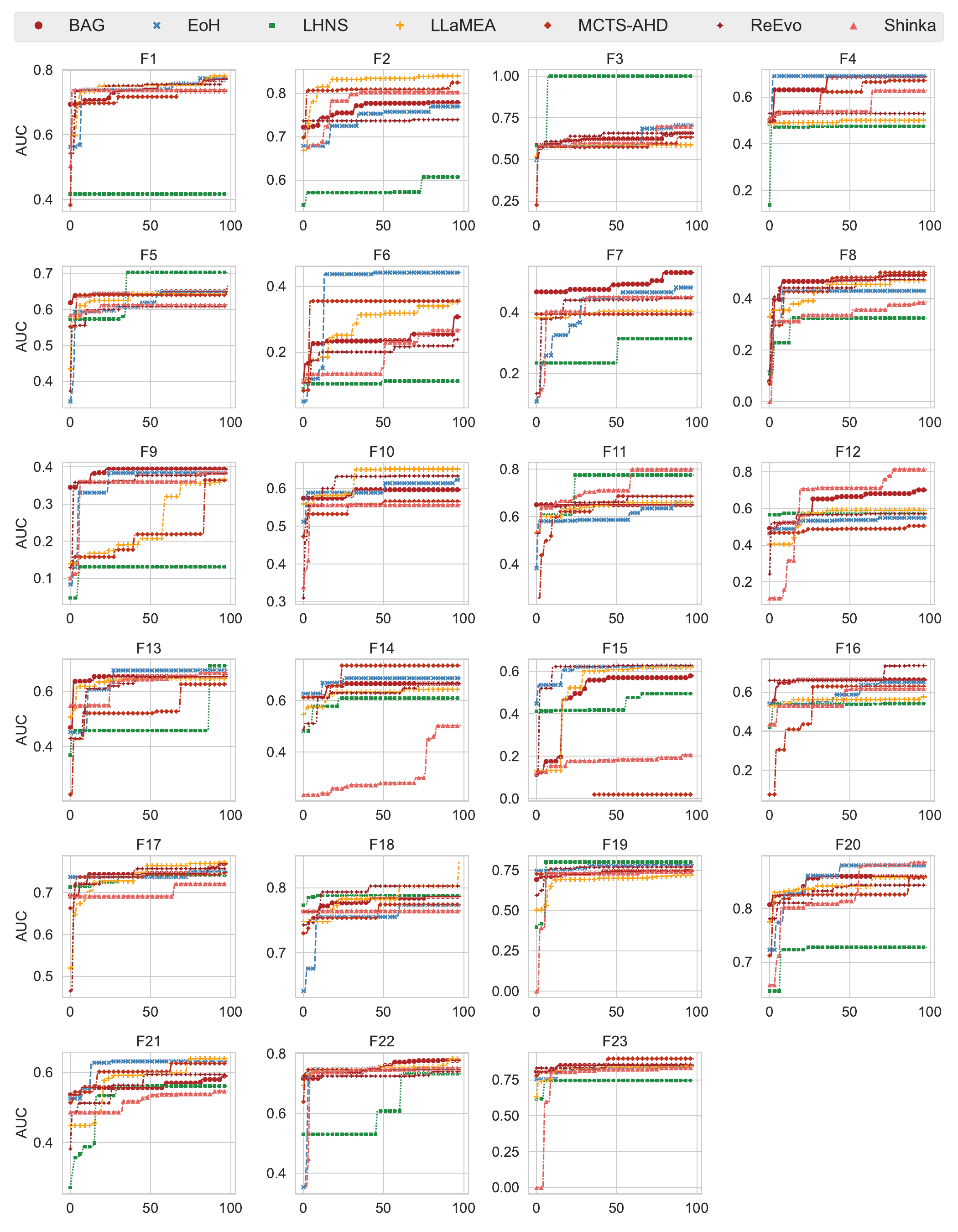}
    \caption{\textcolor{black}{
    AUC values of the algorithms obtained by LLM-driven approaches on all pbo problems. The $x$-axis represents the cumulative number of algorithms generated by the LLM, and the $y$-axis indicates the best-so-far AUC value. The results are obtained using GPT 5 Nano. 
    }}
    \label{fig:convergence_pbo_gpt}
\end{figure*}

\begin{figure*}[!htb]
    \centering
    \includegraphics[width=0.9\textwidth]{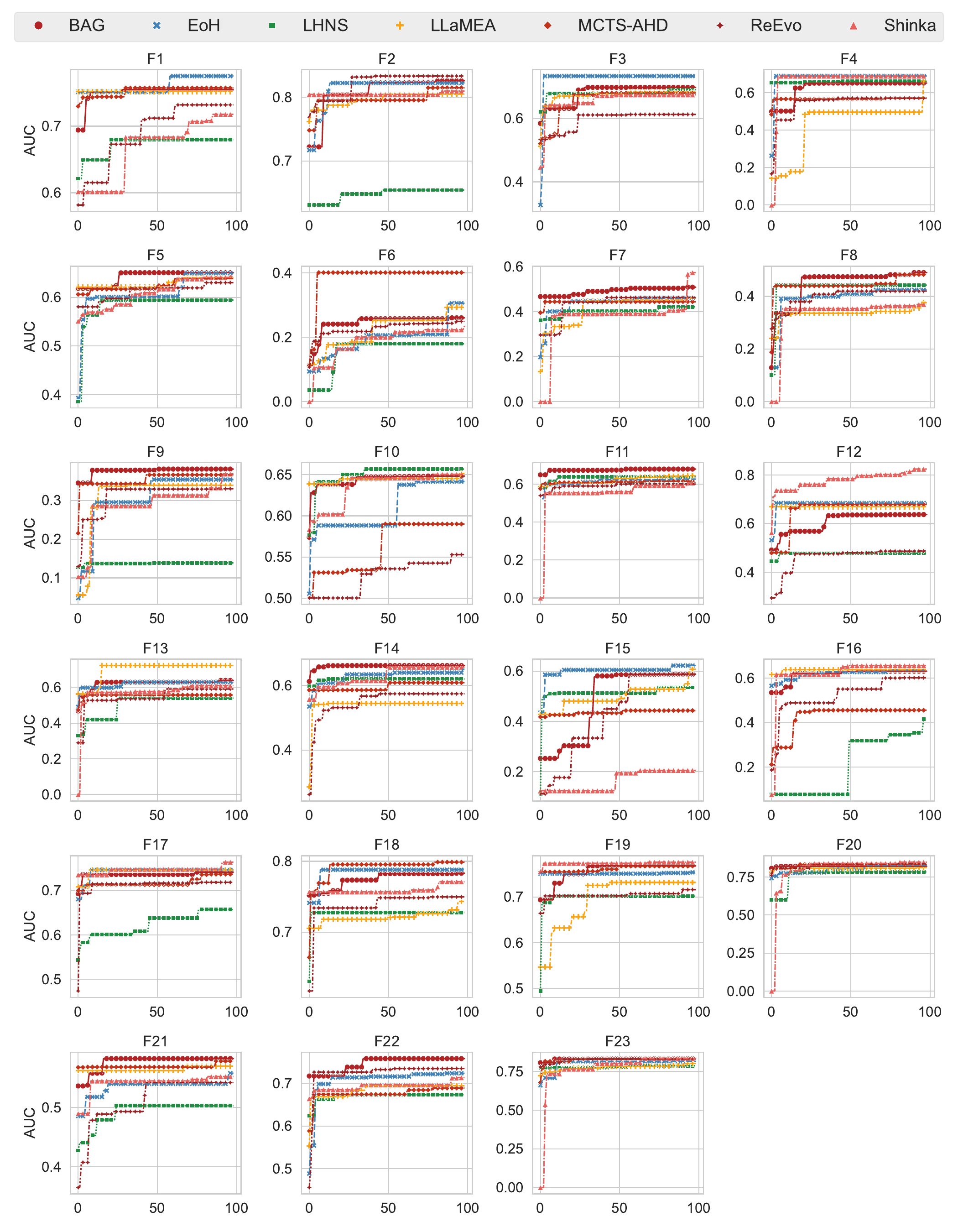}
    \caption{\textcolor{black}{
    AUC values of the algorithms obtained by LLM-driven approaches on all pbo problems. The $x$-axis represents the cumulative number of algorithms generated by the LLM, and the $y$-axis indicates the best-so-far AUC value. The results are obtained using Qwen3 Coder Flash.}}
    \label{fig:convergence_pbo_qwen}
\end{figure*}

\begin{figure*}[!htb]
    \centering
    \includegraphics[width=0.9\textwidth]{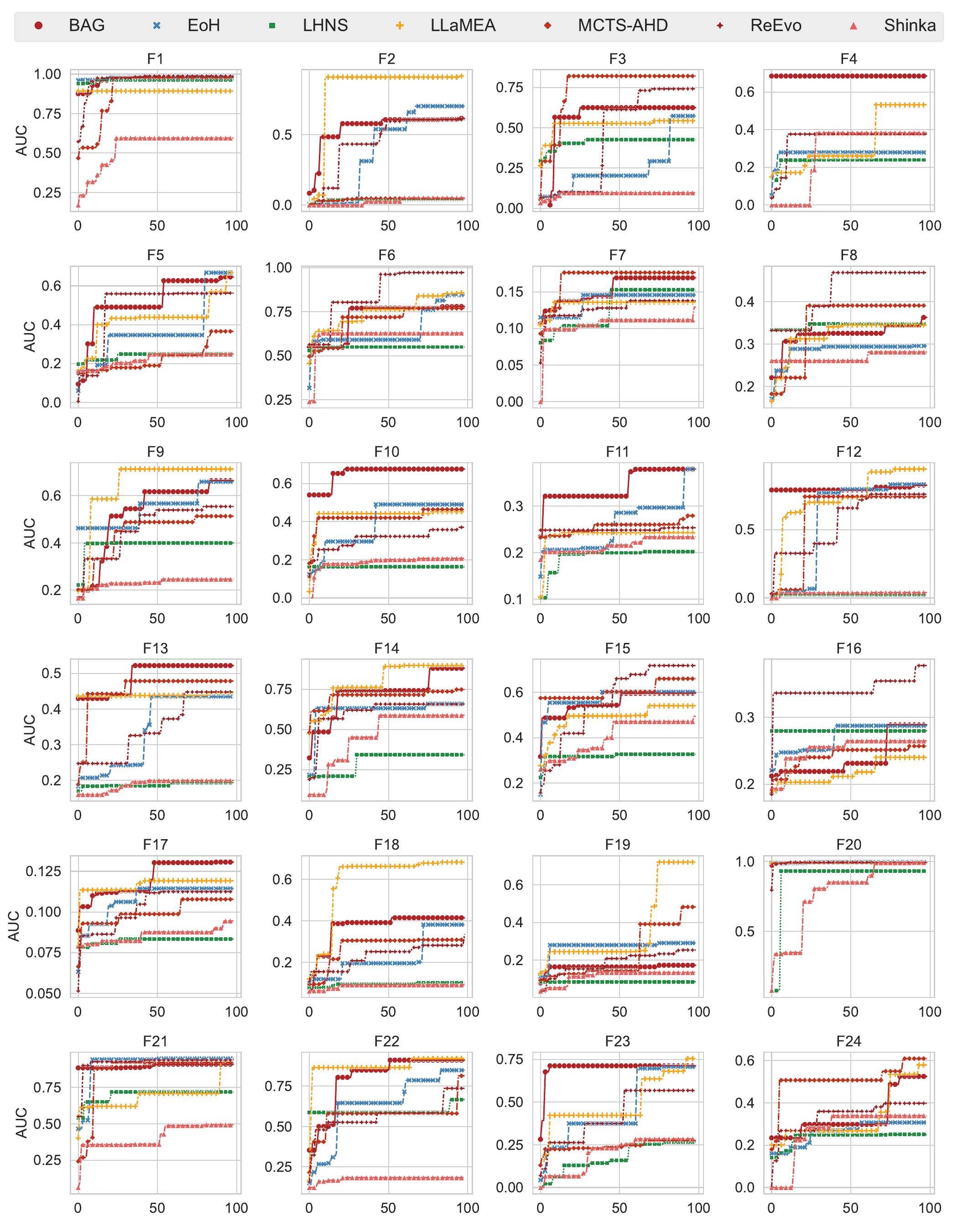}
    \caption{\textcolor{black}{AUC values of the algorithms obtained by LLM-driven approaches on all bbob problems. The $x$-axis represents the cumulative number of algorithms generated by the LLM, and the $y$-axis indicates the best-so-far AUC value. The results are obtained using Gemini 2.5 Flash-Lite. }}
    \label{fig:convergence_bbob_gemini}
\end{figure*}

\begin{figure*}[!htb]
    \centering
    \includegraphics[width=0.9\textwidth]{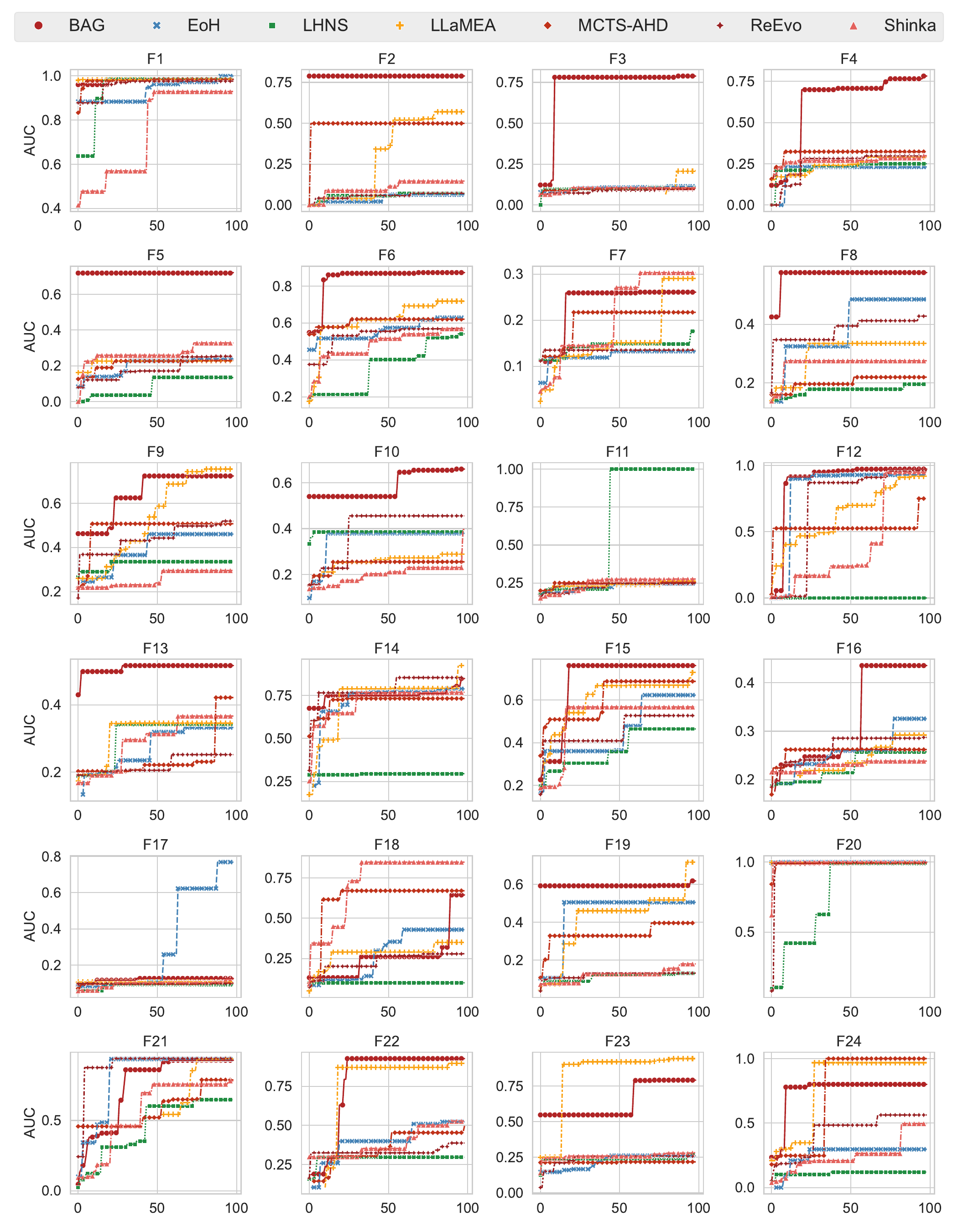}
    \caption{\textcolor{black}{AUC values of the algorithms obtained by LLM-driven approaches on all bbob problems. The $x$-axis represents the cumulative number of algorithms generated by the LLM, and the $y$-axis indicates the best-so-far AUC value. The results are obtained using GPT 5 Nano. }}
    \label{fig:convergence_bbob_gpt}
\end{figure*}
\begin{figure*}[!htb]
    \centering
    \includegraphics[width=0.9\textwidth]{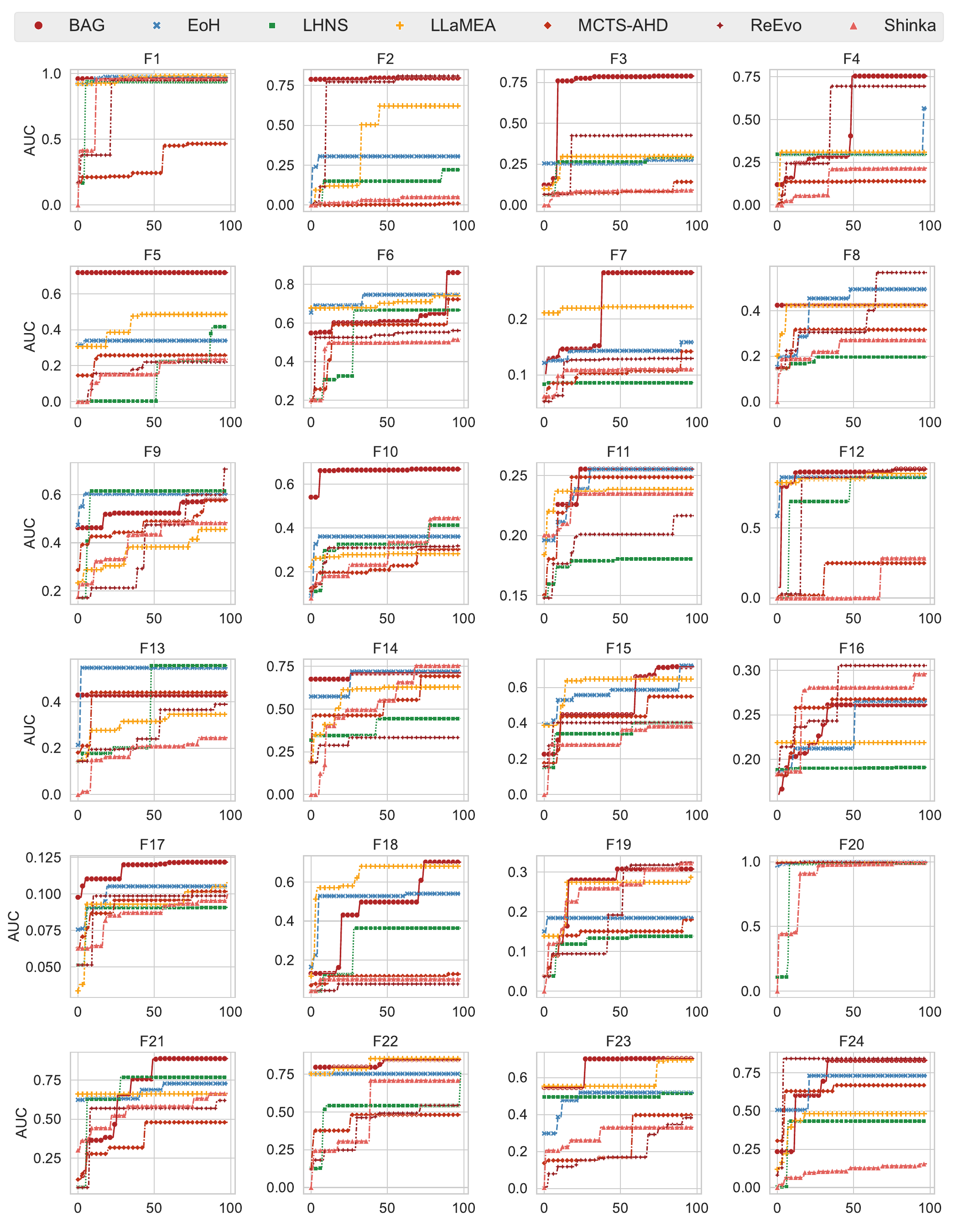}
    \caption{\textcolor{black}{AUC values of the algorithms obtained by LLM-driven approaches on all bbob problems. The $x$-axis represents the cumulative number of algorithms generated by the LLM, and the $y$-axis indicates the best-so-far AUC value. The results are obtained using Qwen3 Coder Flash. }}
    \label{fig:convergence_bbob_qwen}
\end{figure*}

\clearpage



\clearpage


\vfill

\clearpage

\bibliographystyle{IEEEtranN}
\bibliography{arxiv}

@article{romera2024mathematical,
  title={Mathematical discoveries from program search with large language models},
  author={Romera-Paredes, Bernardino and Barekatain, Mohammadamin and Novikov, Alexander and Balog, Matej and Kumar, M Pawan and Dupont, Emilien and Ruiz, Francisco JR and Ellenberg, Jordan S and Wang, Pengming and Fawzi, Omar and others},
  journal={Nature},
  volume={625},
  number={7995},
  pages={468--475},
  year={2024},
  publisher={Nature Publishing Group UK London}
}

@article{grochow2019new,
  title={New applications of the polynomial method: the cap set conjecture and beyond},
  author={Grochow, Joshua},
  journal={Bulletin of the American Mathematical Society},
  volume={56},
  number={1},
  pages={29--64},
  year={2019}
}

@incollection{coffman1984approximation,
  title={Approximation algorithms for bin-packing—an updated survey},
  author={Coffman Jr, Edward G and Garey, Michael R and Johnson, David S},
  booktitle={Algorithm Design for Computer System Design},
  pages={49--106},
  year={1984},
  publisher={Springer}
}

@article{schede2022survey,
  title={A survey of methods for automated algorithm configuration},
  author={Schede, Elias and Brandt, Jasmin and Tornede, Alexander and Wever, Marcel and Bengs, Viktor and H{\"u}llermeier, Eyke and Tierney, Kevin},
  journal={Journal of Artificial Intelligence Research},
  volume={75},
  pages={425--487},
  year={2022}
}

@inproceedings{liuevolution,
  title={{Evolution of Heuristics}: Towards Efficient Automatic Algorithm Design Using Large Language Model},
  author={Liu, Fei and Xialiang, Tong and Yuan, Mingxuan and Lin, Xi and Luo, Fu and Wang, Zhenkun and Lu, Zhichao and Zhang, Qingfu},
  booktitle={International Conference on Machine Learning},
  year = {2024}
}

@inproceedings{yao2025multi,
  title={Multi-objective evolution of heuristic using large language model},
  author={Yao, Shunyu and Liu, Fei and Lin, Xi and Lu, Zhichao and Wang, Zhenkun and Zhang, Qingfu},
  booktitle={AAAI Conference on Artificial Intelligence},
  volume={39},
  pages={27144--27152},
  year={2025}
}

@article{van2024llamea,
  title={{LLaMEA}: A large language model evolutionary algorithm for automatically generating metaheuristics},
  author={van Stein, Niki and B{\"a}ck, Thomas},
  journal={IEEE Transactions on Evolutionary Computation},
  year={2024},
  publisher={IEEE}
}

@article{li2025llameabolargelanguagemodel,
      title={{LLaMEA-BO}: A Large Language Model Evolutionary Algorithm for Automatically Generating {Bayesian} Optimization Algorithms}, 
      author={Wenhu Li and Niki van Stein and Thomas Bäck and Elena Raponi},
      year={2025},
      eprint={2505.21034},
      journal={arXiv},
      primaryClass={cs.LG} 
}

@article{llameahpo,
author = {van Stein, Niki and Vermetten, Diederick and B\"{a}ck, Thomas},
title = {In-the-loop Hyper-Parameter Optimization for {LLM}-Based Automated Design of Heuristics},
year = {2025},
publisher = {Association for Computing Machinery},
address = {New York, NY, USA},
journal = {ACM Transactions Evolutionary Learning and Optimization},
month = apr
}

@article{sun2025automatically,
  title={Automatically discovering heuristics in a complex {SAT} solver with large language models},
  author={Sun, Yiwen and Ye, Furong and Chen, Zhihan and Wei, Ke and Cai, Shaowei},
  journal={arXiv preprint arXiv:2507.22876},
  year={2025}
}

@inproceedings{liularge,
  title={Large Language Models to Enhance {Bayesian} Optimization},
  author={Liu, Tennison and Astorga, Nicol{\'a}s and Seedat, Nabeel and van der Schaar, Mihaela},
  booktitle={International Conference on Learning Representations},
  year = {2024}
}

@article{li2025autopbo,
  title={{AutoPBO}: {LLM}-powered Optimization for Local Search PBO Solvers},
  author={Li, Jinyuan and Chu, Yi and Sun, Yiwen and Zou, Mengchuan and Cai, Shaowei},
  journal={arXiv preprint arXiv:2509.04007},
  year={2025}
}

@inproceedings{zhang2024understanding,
  title={Understanding the importance of evolutionary search in automated heuristic design with large language models},
  author={Zhang, Rui and Liu, Fei and Lin, Xi and Wang, Zhenkun and Lu, Zhichao and Zhang, Qingfu},
  booktitle={International Conference on Parallel Problem Solving from Nature},
  pages={185--202},
  year={2024},
  organization={Springer}
}

@inproceedings{achtibat2024attnlrp,
  title={{AttnLRP}: Attention-Aware Layer-Wise Relevance Propagation for Transformers},
  author={Achtibat, Reduan and Vakilzadeh Hatefi, Sayed Mohammad and Dreyer, Maximilian and Jain, Aakriti and Wiegand, Thomas and Lapuschkin, Sebastian Roland and Samek, Wojciech},
  booktitle={International Conference on Machine Learning},
  year={2024}
}

@article{kerschke2019automated,
  title={Automated algorithm selection: Survey and perspectives},
  author={Kerschke, Pascal and Hoos, Holger H and Neumann, Frank and Trautmann, Heike},
  journal={Evolutionary Computation},
  volume={27},
  number={1},
  pages={3--45},
  year={2019},
  publisher={mIT Press}
}

@article{bartz2020benchmarking,
  title={Benchmarking in optimization: Best practice and open issues},
  author={Bartz-Beielstein, Thomas and Doerr, Carola and Berg, Daan van den and Bossek, Jakob and Chandrasekaran, Sowmya and Eftimov, Tome and Fischbach, Andreas and Kerschke, Pascal and La Cava, William and Lopez-Ibanez, Manuel and others},
  journal={arXiv preprint arXiv:2007.03488},
  year={2020}
}

@article{bennet2021nevergrad,
  title={Nevergrad: black-box optimization platform},
  author={Bennet, Pauline and Doerr, Carola and Moreau, Antoine and Rapin, Jeremy and Teytaud, Fabien and Teytaud, Olivier},
  journal={ACM SIGEVOlution},
  volume={14},
  number={1},
  pages={8--15},
  year={2021},
  publisher={ACM New York, NY, USA}
}

@article{doerr2020benchmarking,
  title={Benchmarking discrete optimization heuristics with {IOH}profiler},
  author={Doerr, Carola and Ye, Furong and Horesh, Naama and Wang, Hao and Shir, Ofer and Back, Thomas},
  journal={Applied Soft Computing},
  volume={88},
  pages={106027},
  year={2020}
}

@article{hansen2021coco,
  title={{COCO}: A platform for comparing continuous optimizers in a black-box setting},
  author={Hansen, Nikolaus and Auger, Anne and Ros, Raymond and Mersmann, Olaf and Tu{\v{s}}ar, Tea and Brockhoff, Dimo},
  journal={Optimization Methods and Software},
  volume={36},
  number={1},
  pages={114--144},
  year={2021},
  publisher={Taylor \& Francis}
}

@article{yu2025autonomous,
  title={Autonomous Code Evolution Meets {NP}-Completeness},
  author={Yu, Cunxi and Liang, Rongjian and Ho, Chia-Tung and Ren, Haoxing},
  journal={arXiv preprint arXiv:2509.07367},
  year={2025}
}

@article{back2023evolutionary,
  title={Evolutionary algorithms for parameter optimization—thirty years later},
  author={B{\"a}ck, Thomas HW and Kononova, Anna V and van Stein, Bas and Wang, Hao and Antonov, Kirill A and Kalkreuth, Roman T and de Nobel, Jacob and Vermetten, Diederick and de Winter, Roy and Ye, Furong},
  journal={Evolutionary Computation},
  volume={31},
  number={2},
  pages={81--122},
  year={2023},
  publisher={MIT Press}
}

@article{papenmeier2025bencher,
  title={Bencher: Simple and Reproducible Benchmarking for Black-Box Optimization},
  author={Papenmeier, Leonard and Nardi, Luigi},
  journal={International Conference on Machine Learning},
  year={2025}
}

@article{montavon2019layer,
  title={Layer-wise relevance propagation: an overview},
  author={Montavon, Gr{\'e}goire and Binder, Alexander and Lapuschkin, Sebastian and Samek, Wojciech and M{\"u}ller, Klaus-Robert},
  journal={Explainable AI: interpreting, explaining and visualizing deep learning},
  pages={193--209},
  year={2019},
  publisher={Springer}
}

@article{droste2002analysis,
  title={On the analysis of the (1 + 1) evolutionary algorithm},
  author={Droste, Stefan and Jansen, Thomas and Wegener, Ingo},
  journal={Theoretical Computer Science},
  volume={276},
  number={1-2},
  pages={51--81},
  year={2002},
  publisher={Elsevier}
}

@article{witt2006runtime,
  title={Runtime analysis of the ($\mu$ + 1) {EA} on simple pseudo-Boolean functions},
  author={Witt, Carsten},
  journal={Evolutionary Computation},
  volume={14},
  number={1},
  pages={65--86},
  year={2006},
  publisher={MIT Press One Rogers Street, Cambridge, MA 02142-1209, USA journals-info~…}
}

@article{zhao2024explainability,
  title={Explainability for large language models: A survey},
  author={Zhao, Haiyan and Chen, Hanjie and Yang, Fan and Liu, Ninghao and Deng, Huiqi and Cai, Hengyi and Wang, Shuaiqiang and Yin, Dawei and Du, Mengnan},
  journal={ACM Transactions on Intelligent Systems and Technology},
  volume={15},
  number={2},
  pages={1--38},
  year={2024},
  publisher={ACM New York, NY}
}

@inproceedings{ye2024reevo,
  title={Re{E}vo: Large language models as hyper-heuristics with reflective evolution},
  author={Ye, Haoran and Wang, Jiarui and Cao, Zhiguang and Berto, Federico and Hua, Chuanbo and Kim, Haeyeon and Park, Jinkyoo and Song, Guojie},
  booktitle={Advances in Neural Information Processing Systems},
  volume={37},
  pages={43571--43608},
  year={2024}
}

@article{wu2020perturbed,
  title={Perturbed masking: Parameter-free probing for analyzing and interpreting {BERT}},
  author={Wu, Zhiyong and Chen, Yun and Kao, Ben and Liu, Qun},
  journal={arXiv preprint arXiv:2004.14786},
  year={2020}
}

@incollection{hoos2018stochastic,
  title={Stochastic local search},
  author={Hoos, Holger H and St\"utzle, Thomas},
  booktitle={Handbook of Approximation Algorithms and Metaheuristics},
  pages={297--307},
  year={2018},
  publisher={Chapman and Hall/CRC}
}

@Inproceedings{kariyappa2024progressive,
  title = 	 {Progressive Inference: Explaining Decoder-Only Sequence Classification Models Using Intermediate Predictions},
  author =       {Kariyappa, Sanjay and Lecue, Freddy and Mishra, Saumitra and Pond, Christopher and Magazzeni, Daniele and Veloso, Manuela},
  booktitle = 	 {International Conference on Machine Learning},
  pages = 	 {23238--23255},
  year = 	 {2024},
  editor = 	 {Salakhutdinov, Ruslan and Kolter, Zico and Heller, Katherine and Weller, Adrian and Oliver, Nuria and Scarlett, Jonathan and Berkenkamp, Felix},
  volume = 	 {235},
  series = 	 {Proceedings of Machine Learning Research},
  month = 	 {21--27 Jul},
  publisher =    {PMLR},
}

@inproceedings{li2016visualizing,
  title={Visualizing and understanding neural models in {NLP}},
  author={Li, Jiwei and Chen, Xinlei and Hovy, Eduard and Jurafsky, Dan},
  booktitle={Proceedings of the 2016 Conference of the North American Chapter of the Association for Computational Linguistics: Human Language Technologies},
  pages={681--691},
  year={2016}
}

@article{schneider2024explainable,
  title={Explainable {Generative AI} ({GenXAI}): a survey, conceptualization, and research agenda},
  author={Schneider, Johannes},
  journal={Artificial Intelligence Review},
  volume={57},
  number={11},
  pages={289},
  year={2024},
  publisher={Springer}
}

@inproceedings{kokalj2021bert,
    title = "{BERT} meets Shapley: Extending {SHAP} Explanations to Transformer-based Classifiers",
    author = "Kokalj, Enja  and
      {\v{S}}krlj, Bla{\v{z}}  and
      Lavra{\v{c}}, Nada  and
      Pollak, Senja  and
      Robnik-{\v{S}}ikonja, Marko",
    booktitle = "EACL Hackashop on News Media Content Analysis and Automated Report Generation",
    year = "2021",
    pages = "16--21",
}

@inproceedings{ribeiro2016should,
  title={Why should {I} trust you? Explaining the predictions of any classifier},
  author={Ribeiro, Marco Tulio and Singh, Sameer and Guestrin, Carlos},
  booktitle={ACM SIGKDD International Conference on Knowledge Discovery and Data Mining},
  pages={1135--1144},
  year={2016}
}

@article{hansen2022anytime,
  title={Anytime performance assessment in blackbox optimization benchmarking},
  author={Hansen, Nikolaus and Auger, Anne and Brockhoff, Dimo and Tu{\v{s}}ar, Tea},
  journal={IEEE Transactions on Evolutionary Computation},
  volume={26},
  number={6},
  pages={1293--1305},
  year={2022},
  publisher={IEEE}
}

@article{wang2022iohanalyzer,
  title={{IOH}analyzer: Detailed performance analyses for iterative optimization heuristics},
  author={Wang, Hao and Vermetten, Diederick and Ye, Furong and Doerr, Carola and B{\"a}ck, Thomas},
  journal={ACM Transactions on Evolutionary Learning and Optimization},
  volume={2},
  number={1},
  pages={1--29},
  year={2022},
  publisher={ACM New York, NY}
}

@inproceedings{sundararajan2017axiomatic,
  title={Axiomatic attribution for deep networks},
  author={Sundararajan, Mukund and Taly, Ankur and Yan, Qiqi},
  booktitle={International Conference on Machine Learning},
  pages={3319--3328},
  year={2017},
  organization={PMLR}
}

@article{enguehard2023sequential,
  title={Sequential integrated gradients: a simple but effective method for explaining language models},
  author={Enguehard, Joseph},
  journal={arXiv preprint arXiv:2305.15853},
  year={2023}
}

@inproceedings{chang2025jopa,
    title = "{J}o{PA}: Explaining Large Language Model{'}s Generation via Joint Prompt Attribution",
    author = "Chang, Yurui  and
      Cao, Bochuan  and
      Wang, Yujia  and
      Chen, Jinghui  and
      Lin, Lu",
    booktitle = "Annual Meeting of the Association for Computational Linguistics",
    year = "2025",
    pages = "22106--22122"
}

@article{hansen2016cma,
  title={The {CMA} evolution strategy: A tutorial},
  author={Hansen, Nikolaus},
  journal={arXiv preprint arXiv:1604.00772},
  year={2016}
}

@inproceedings{krause2016cma,
  title={{CMA-ES} with optimal covariance update and storage complexity},
  author={Krause, Oswin and Arbon{\`e}s, D{\'\i}dac Rodr{\'\i}guez and Igel, Christian},
  booktitle={Advances in Neural Information Processing Systems},
  volume={29},
  year={2016}
}

@inproceedings{chotard2012cumulative,
  title={Cumulative step-size adaptation on linear functions},
  author={Chotard, Alexandre and Auger, Anne and Hansen, Nikolaus},
  booktitle={International Conference on Parallel Problem Solving from Nature},
  pages={72--81},
  year={2012},
  organization={Springer}
}

@article{wang2018particle,
  title={Particle swarm optimization algorithm: an overview},
  author={Wang, Dongshu and Tan, Dapei and Liu, Lei},
  journal={Soft computing},
  volume={22},
  number={2},
  pages={387--408},
  year={2018},
  publisher={Springer}
}

@article{das2010differential,
  title={Differential evolution: A survey of the state-of-the-art},
  author={Das, Swagatam and Suganthan, Ponnuthurai Nagaratnam},
  journal={IEEE Transactions on Evolutionary Computation},
  volume={15},
  number={1},
  pages={4--31},
  year={2010},
  publisher={IEEE}
}

@article{ye2022automated,
  title={Automated configuration of genetic algorithms by tuning for anytime performance},
  author={Ye, Furong and Doerr, Carola and Wang, Hao and B{\"a}ck, Thomas},
  journal={IEEE Transactions on Evolutionary Computation},
  volume={26},
  number={6},
  pages={1526--1538},
  year={2022},
  publisher={IEEE}
}

@InProceedings{mameta,
  title = 	 {Meta-Black-Box-Optimization through Offline Q-function Learning},
  author =       {Ma, Zeyuan and Cao, Zhiguang and Jiang, Zhou and Guo, Hongshu and Gong, Yue-Jiao},
  booktitle = 	 {International Conference on Machine Learning},
  pages = 	 {41807--41826},
  year = 	 {2025},
  volume = 	 {267},
  month = 	 {13--19 Jul},
  publisher =    {PMLR},
}

@inproceedings{li2024pretrained,
  title={Pretrained optimization model for zero-shot black box optimization},
  author={Li, Xiaobin and Wu, Kai and Zhang, Xiaoyu and Wang, Handing and Liu, Jing and others},
  booktitle={Advances in Neural Information Processing Systems},
  volume={37},
  pages={14283--14324},
  year={2024}
}

@inproceedings{
song2025reinforced,
title={Reinforced In-Context Black-Box Optimization},
author={Lei Song and Chen-Xiao Gao and Ke Xue and Chenyang Wu and Dong Li and Jianye HAO and Zongzhang Zhang and Chao Qian},
year={2025},
booktitle={International Conference on Learning Representations}
}

@inproceedings{hansen2010comparing,
  title={Comparing results of 31 algorithms from the black-box optimization benchmarking BBOB-2009},
  author={Hansen, Nikolaus and Auger, Anne and Ros, Raymond and Finck, Steffen and Po{\v{s}}{\'\i}k, Petr},
  booktitle={Conference Companion on Genetic and Evolutionary Computation},
  pages={1689--1696},
  year={2010}
}

@inproceedings{wolf2020transformers,
    title = "Transformers: State-of-the-Art Natural Language Processing",
    author = "Wolf, Thomas  and
      Debut, Lysandre  and
      Sanh, Victor  and
      Chaumond, Julien  and
      Delangue, Clement  and
      Moi, Anthony  and
      Cistac, Pierric  and
      Rault, Tim  and
      Louf, Remi  and
      Funtowicz, Morgan  and
      Davison, Joe  and
      Shleifer, Sam  and
      von Platen, Patrick  and
      Ma, Clara  and
      Jernite, Yacine  and
      Plu, Julien  and
      Xu, Canwen  and
      Le Scao, Teven  and
      Gugger, Sylvain  and
      Drame, Mariama  and
      Lhoest, Quentin  and
      Rush, Alexander",
    booktitle = "Conference on Empirical Methods in Natural Language Processing: System Demonstrations",
    year = "2020",
    pages = "38--45",
}

@article{huang2025autonomous,
  title={Autonomous multi-objective optimization using large language model},
  author={Huang, Yuxiao and Wu, Shenghao and Zhang, Wenjie and Wu, Jibin and Feng, Liang and Tan, Kay Chen},
  journal={IEEE Transactions on Evolutionary Computation},
  year={2025},
  publisher={IEEE}
}

@inproceedings{dong2024survey,
  title={A survey on in-context learning},
  author={Dong, Qingxiu and Li, Lei and Dai, Damai and Zheng, Ce and Ma, Jingyuan and Li, Rui and Xia, Heming and Xu, Jingjing and Wu, Zhiyong and Chang, Baobao and others},
  booktitle={Proceedings of the 2024 conference on empirical methods in natural language processing},
  pages={1107--1128},
  year={2024}
}

@article{ren2020codebleu,
  title={{CodeBLEU}: a method for automatic evaluation of code synthesis},
  author={Ren, Shuo and Guo, Daya and Lu, Shuai and Zhou, Long and Liu, Shujie and Tang, Duyu and Sundaresan, Neel and Zhou, Ming and Blanco, Ambrosio and Ma, Shuai},
  journal={arXiv preprint arXiv:2009.10297},
  year={2020}
}

@incollection{stutzle2018automated,
  title={Automated design of metaheuristic algorithms},
  author={St{\"u}tzle, Thomas and L{\'o}pez-Ib{\'a}{\~n}ez, Manuel},
  booktitle={Handbook of metaheuristics},
  pages={541--579},
  year={2018},
  publisher={Springer}
}

@inproceedings{xie2025llm,
  title={{LLM}-Driven Neighborhood Search for Efficient Heuristic Design},
  author={Xie, Zhuoliang and Liu, Fei and Wang, Zhenkun and Zhang, Qingfu},
  booktitle={2025 IEEE Congress on Evolutionary Computation (CEC)},
  pages={1--8},
  year={2025},
  organization={IEEE}
}

@inproceedings{zheng2025monte,
  title={{Monte} {Carlo} Tree Search for Comprehensive Exploration in {LLM}-Based Automatic Heuristic Design},
  author={Zheng, Zhi and Xie, Zhuoliang and Wang, Zhenkun and Hooi, Bryan},
  booktitle={International Conference on Machine Learning},
  pages={78338--78373},
  year={2025},
  organization={PMLR}
}

@inproceedings{papineni2002bleu,
  title={{Bleu}: a method for automatic evaluation of machine translation},
  author={Papineni, Kishore and Roukos, Salim and Ward, Todd and Zhu, Wei-Jing},
  booktitle={Proceedings of the 40th annual meeting of the Association for Computational Linguistics},
  pages={311--318},
  year={2002}
}

@inproceedings{
lange2026shinkaevolve,
title={{ShinkaEvolve}: Towards Open-Ended and Sample-Efficient Program Evolution},
author={Robert Tjarko Lange and Yuki Imajuku and Edoardo Cetin},
booktitle={The Fourteenth International Conference on Learning Representations},
year={2026},
url={https://openreview.net/forum?id=lKEdGCoDNC}
}

@article{liu2023algorithm,
  title={Algorithm evolution using large language model},
  author={Liu, Fei and Tong, Xialiang and Yuan, Mingxuan and Zhang, Qingfu},
  journal={arXiv preprint arXiv:2311.15249},
  year={2023}
}

@article{novikov2025alphaevolve,
  title={{AlphaEvolve}: A coding agent for scientific and algorithmic discovery},
  author={Novikov, Alexander and V{\~u}, Ng{\^a}n and Eisenberger, Marvin and Dupont, Emilien and Huang, Po-Sen and Wagner, Adam Zsolt and Shirobokov, Sergey and Kozlovskii, Borislav and Ruiz, Francisco JR and Mehrabian, Abbas and others},
  journal={arXiv preprint arXiv:2506.13131},
  year={2025}
}

@inproceedings{
gideoni2025random,
title={Random Baselines for Simple Code Problems are Competitive with Code Evolution},
author={Yonatan Gideoni and Yujin Tang and Sebastian Risi and Yarin Gal},
booktitle={NeurIPS 2025 Fourth Workshop on Deep Learning for Code},
year={2025},
url={https://openreview.net/forum?id=rbVIpbmbTc}
}

\end{document}